\documentclass[10pt,twocolumn,letterpaper]{article}

\usepackage[pagenumbers]{cvpr} 

\usepackage{siunitx}
\usepackage{pifont}
\usepackage{multirow}
\usepackage[symbol]{footmisc}

\usepackage[dvipsnames]{xcolor}

\usepackage{verbatim}
\usepackage{fancyvrb}
\usepackage{framed}
\usepackage{xcolor}
\usepackage{tabularx}
\usepackage{amssymb}
\usepackage[accsupp]{axessibility}

\definecolor{shadecolor}{gray}{0.95}  

\definecolor{cvprblue}{rgb}{0.21,0.49,0.74}
\usepackage[pagebackref,breaklinks,colorlinks,allcolors=cvprblue]{hyperref}

\def\paperID{} 
\def\confName{CVPR}
\def\confYear{2025}

\usepackage{tabularx}

\title{MObI: Multimodal Object Inpainting Using Diffusion Models}

\author{
Alexandru Buburuzan\textsuperscript{1,2,}\footnotemark ~
Anuj Sharma\textsuperscript{1} ~
John Redford\textsuperscript{1} ~
Puneet K. Dokania\textsuperscript{1,3} ~
Romain Mueller\textsuperscript{1} \\
\textsuperscript{1}FiveAI \quad
\textsuperscript{2}The University of Manchester \quad
\textsuperscript{3}University of Oxford \\
}

\begin{document}
\twocolumn[{%
\renewcommand\twocolumn[1][]{#1}%
\maketitle
\vspace{-0.2cm}
\centering
\captionsetup{type=figure}
\includegraphics[width=0.96\textwidth]{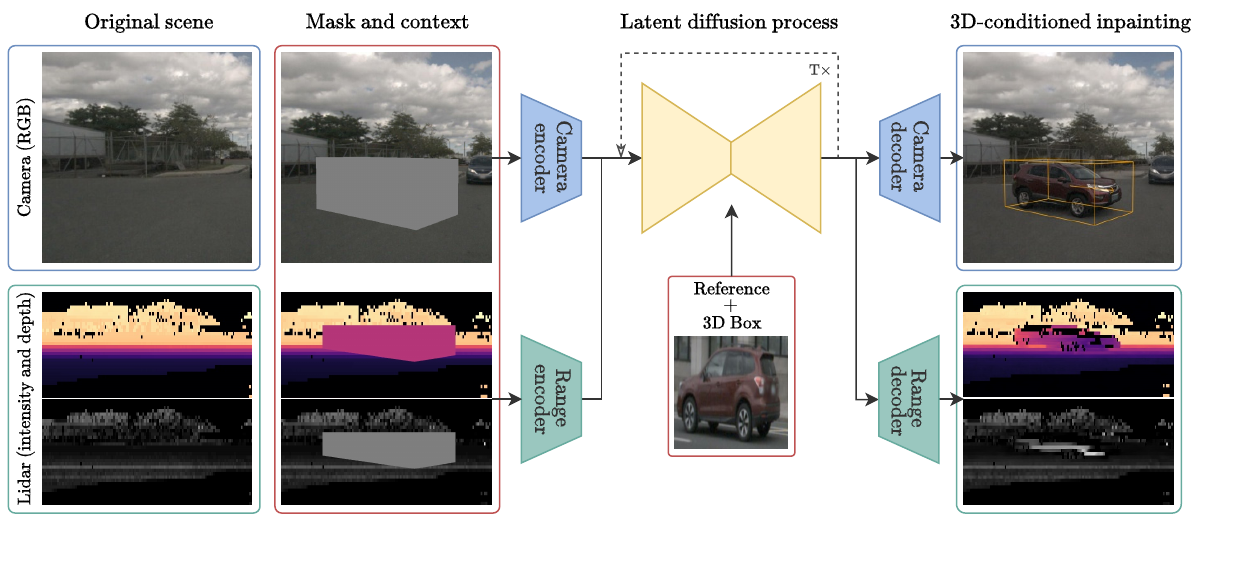}
\vspace{-0.4cm}
}]
\maketitle
\footnotetext[1]{Work done during an internship at FiveAI. Corresponding author {\tt\scriptsize alexandru-stefan.buburuzan@student.manchester.ac.uk}.}
\begin{abstract}
Safety-critical applications, such as autonomous driving, require extensive multimodal data for rigorous testing. Methods based on synthetic data are gaining prominence due to the cost and complexity of gathering real-world data but require a high degree of realism and controllability in order to be useful. This paper introduces MObI, a novel framework for \textbf{M}ultimodal \textbf{Ob}ject \textbf{I}npainting that leverages a diffusion model to create realistic and controllable object inpaintings across perceptual modalities, demonstrated for both camera and lidar simultaneously. Using a single reference RGB image, MObI enables objects to be seamlessly inserted into existing multimodal scenes at a 3D location specified by a bounding box, while maintaining semantic consistency and multimodal coherence. Unlike traditional inpainting methods that rely solely on edit masks, our 3D bounding box conditioning gives objects accurate spatial positioning and realistic scaling. As a result, our approach can be used to insert novel objects flexibly into multimodal scenes, providing significant advantages for testing perception models.
Project page: \href{https://alexbubu.com/mobi}{https://alexbubu.com/mobi}
\end{abstract}    
\begin{figure*}[htbp]
    \centering
    \small
    \begin{minipage}{0.57\textwidth}
    \begin{tabularx}{0.96\columnwidth}{@{}>{\centering\arraybackslash}X>{\centering\arraybackslash}X>{\centering\arraybackslash}X>{\centering\arraybackslash}X@{}}
        Reference & Edit mask & PbE~\cite{yang2023paint} & Ours \\
    \end{tabularx}
    \includegraphics[width=\columnwidth]{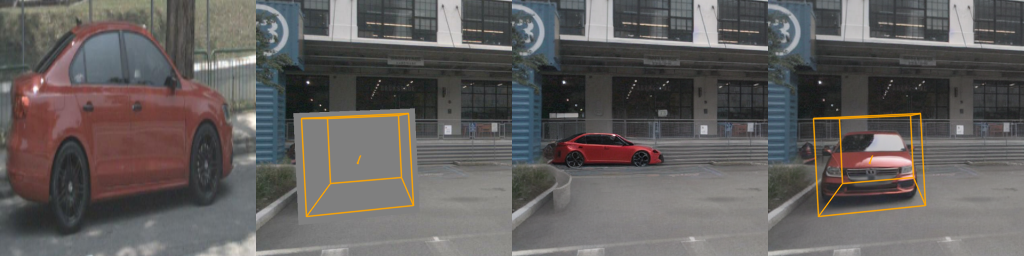}
    \includegraphics[width=\columnwidth]{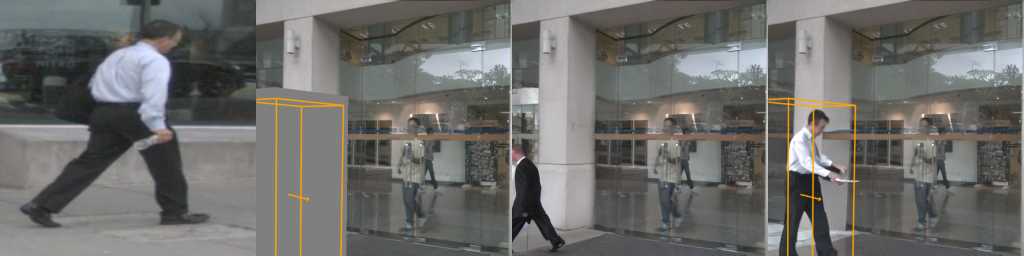} \\
    \centering Object inpainting
    \end{minipage}
    \small
    \hspace{10pt}
    \begin{minipage}{0.4\textwidth}
    \centering
    \vspace{11pt}
    \begin{tabular}{r@{\hspace{4pt}}c@{\hspace{2pt}}c@{\hspace{4pt}}l}
        \rotatebox{90}{\qquad{Original}} & 
        \includegraphics[width=0.35\columnwidth]{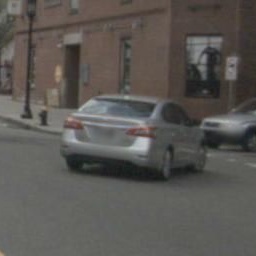} &
        \includegraphics[width=0.35\columnwidth]{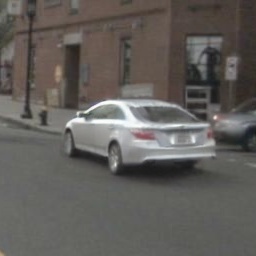} & 
        \rotatebox{90}{\qquad{PbE~\cite{yang2023paint}}} \\
        \rotatebox{90}{{\quad NeuRAD~\cite{tonderski2024neurad}}} & 
        \includegraphics[width=0.35\columnwidth]{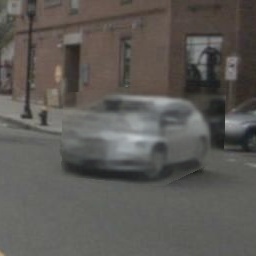} &
        \includegraphics[width=0.35\columnwidth]{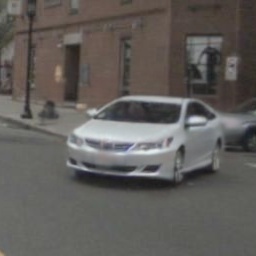} & 
        \rotatebox{90}{\qquad\quad{Ours}} \\
    \end{tabular}
    \centering Object 180$^\circ$ flip\\
    \end{minipage}
    \caption{Our method can inpaint objects with a high degree of realism and controllability. Left: object inpainting methods based on edit masks alone such as Paint-by-Example~\cite{yang2023paint} (PbE) achieve high realism but can lead to surprising results because there are often multiple semantically consistent ways to inpaint an object within a scene.
    Right: methods based on 3D reconstruction such as NeuRAD~\cite{tonderski2024neurad} have strong controllability but sometimes lead to low realism, especially for object viewpoints that have not been observed.
    Our method achieves both high semantic consistency and controllability of the generation.
    }
    \label{fig:failure modes of pbe}
\end{figure*}
\section{Introduction}
\label{sec:intro}
Extensive multimodal data, including camera and lidar, is crucial for the safe testing and deployment of autonomous driving systems. 
However, collecting large amounts of multimodal data in the real world can be prohibitively expensive because rare but high-severity failures have an outstripped impact on the overall safety of such systems~\cite{koopman2016challenges}.
Synthetic data offers a way to address this problem by allowing the generation of diverse safety-critical situations before deployment, but existing methods often fall short either by lacking controllability or realism.

For example, reference-based image inpainting methods~\cite{yang2023paint, chen2023anydoor, ruiz2024magicinsertstyleawaredraganddrop,kulal2023puttingpeopleplaceaffordanceaware} can produce realistic samples that seamlessly blend into the scene using a single reference, but they often lack precise control over the 3D positioning and orientation of the inserted objects.
In contrast, methods based on actor insertion using 3D assets~\cite{wang2023cadsim, chang2024just, zhou2023scene, wei2024editable, chen2021geosim, lin2024drive, multitest, li2023lift3d} provide a high degree of control---enabling precise object placement in the scene---but often struggle to achieve realistic blending and require high-quality 3D assets, which can be challenging to produce. Similarly, reconstruction methods~\cite{prism1bywayve,tonderski2024neurad, yang2023unisim} are also highly controllable but require almost full coverage of the inserted actor.
We illustrate some of these shortcomings in \cref{fig:failure modes of pbe}.
More recent methods have explored 3D geometric control for image editing~\cite{wang2025diffusion, wu2024neural, yenphraphai2024image, pandey2024diffusion, michel2024object, yuan2023customnet}, as well as object-level lidar generation~\cite{kirby2024logen}. However, none consider multimodal generation, which is crucial in autonomous driving.
We provide an extended overview of the prior art in \cref{sec:supple:extended related work}.

Recent advancements in controllable full-scene generation in autonomous driving for multiple cameras~\cite{gao2023magicdrive, li2023drivingdiffusion, wen2023panacea, su2024text2street, huang2024subjectdrivescalinggenerativedata, drivescape}, and lidar~\cite{lidardiffusion, lidargen, hu2024rangeldmfastrealisticlidar, xiong2023ultralidarlearningcompactrepresentations, bian2024dynamiccitylargescalelidargeneration,xie2024x} have led to impressive results. 
However, generating full scenes can create a large domain gap, especially for downstream tasks such as object detection, making it difficult to generate realistic counterfactual examples.
For this reason, works such as GenMM~\cite{singh2024genmm} have focused instead on camera-lidar object inpainting using a multi-stage pipeline.
We take a similar approach, but propose an end-to-end method that generates camera and lidar jointly.

The contributions of this work are threefold: (i) we propose a multimodal inpainting method for joint camera-lidar editing from a single reference image, (ii) we condition an object image inpainting method on a 3D bounding box to enforce precise spatial placement, and (iii) we demonstrate the effectiveness of our approach in generating realistic and controllable multimodal counterfactuals of driving scenes.

\begin{figure*}[t]
  \centering
  \includegraphics[width=\linewidth]{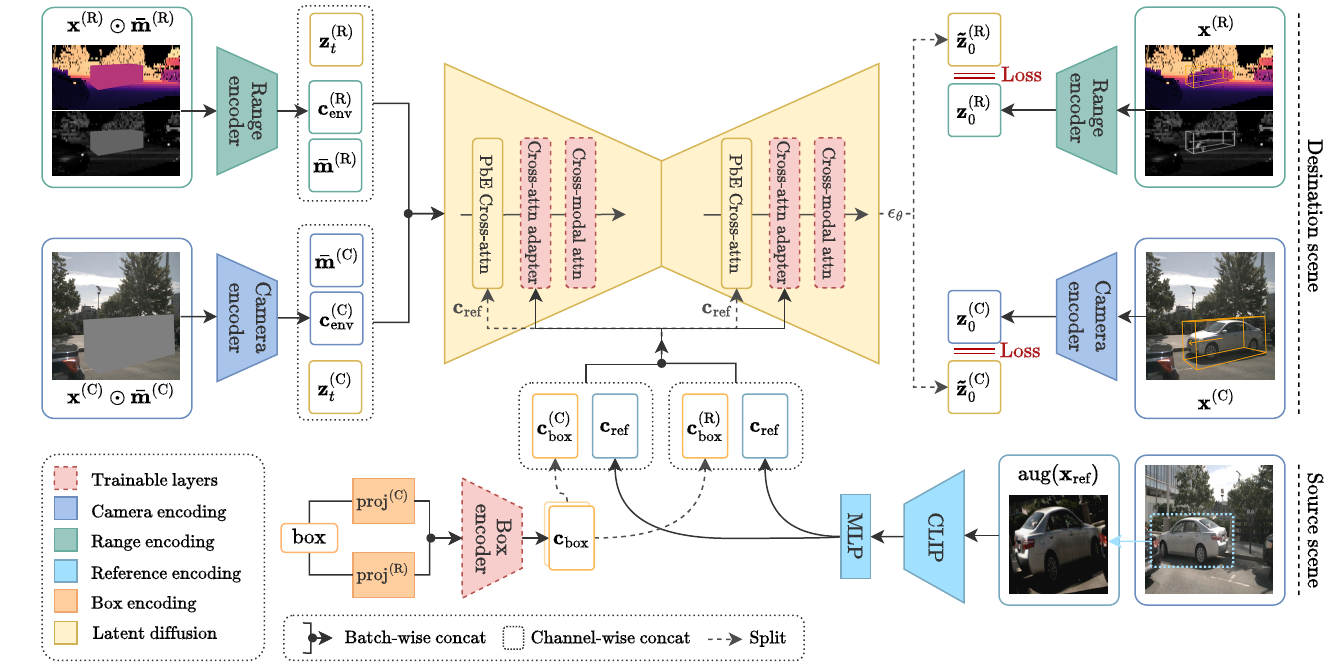}
  \caption{MObI architecture and training procedure.}
  \label{fig:architecture}
\end{figure*}

\section{Method}
\label{sec:method}

We extend Paint-by-Example~\cite{yang2023paint} (PbE), a reference-based image inpainting method, to include bounding box conditioning and to jointly generate camera and lidar perception inputs. We train a diffusion model~\cite{rombach2022high, ho2020denoising, sohl2015deep} using the architecture illustrated on~\cref{fig:architecture}, where the denoising process is conditioned on the latent representations of the camera and lidar range view contexts ($\mathbf{c}^{\text{(R)}}_{\text{env}}$ and $\mathbf{c}^{\text{(C)}}_{\text{env}}$), the RGB object reference $\mathbf{c}_{\text{ref}}$, a per-modality projected 3D bounding box conditioning ($\mathbf{c}_{\text{box}}^{\text{(R)}}$ and $\mathbf{c}_{\text{box}}^{\text{(C)}}$) and the complement of the edit mask targets ($\mathbf{\bar{m}}^{\text{(C)}}$ and $\mathbf{\bar{m}}^{\text{(R)}}$). 
The diffusion model \( \epsilon_\theta \) is trained in a self-supervised manner as in~\cite{yang2023paint} to predict the full scene based on the masked-out inputs.
More formally, the model predicts the total noise added to the latent representation of the scene \(\{ \mathbf{z}_0^{\text{(R)}}, \mathbf{z}_0^{\text{(C)}} \} \) using the loss
\begin{align*}
  \mathcal{L} = \mathbb{E}_{\mathbf{z}^{\text{(R)}}_0, \mathbf{z}^{\text{(C)}}_0, t, \mathbf{c}, \epsilon \sim \mathcal{N}(0, 1)} 
  \left[ \left\| \epsilon - \epsilon_{\theta}(\mathbf{z}^{\text{(R)}}_t, \mathbf{z}^{\text{(C)}}_t, \mathbf{c}, t) \right\|^2 \right],
\end{align*}
where $\mathbf{c} = \{ \mathbf{c}^{\text{(R)}}_{\text{env}}, \mathbf{c}^{\text{(C)}}_{\text{env}}, \mathbf{c}_{\text{ref}}, \mathbf{c}_{\text{box}}^{\text{(R)}}, \mathbf{c}_{\text{box}}^{\text{(C)}}, \mathbf{\bar{m}}^{\text{(R)}}, \mathbf{\bar{m}}^{\text{(C)}}\}$.
The input of the UNet-style network~\cite{ronneberger2015u} is the noised sample ($\mathbf{z}_t^{\text{(R)}}$ and $\mathbf{z}_t^{\text{(C)}}$) at step \( t \), concatenated channel-wise with the latent representation of the scene context and its corresponding edit mask, resized to the latent dimension.

\subsection{Multimodal encoding}
\label{sec:method:multimodal encoding}

\paragraph{Image encoding}
The model is trained to insert an object from a source scene with image $ I_s \in \mathbb{R}^{H \times W \times 3}$ and bonding box $ \text{box}_s \in \mathbb{R}^{8 \times 3}$, into a destination scene with corresponding camera image $ I_d \in \mathbb{R}^{H \times W \times 3}$ and annotation bounding box $ \text{box}_d \in \mathbb{R}^{8 \times 3}$. During training, these bounding boxes correspond to the same object at different timestamps, while at inference, they can be chosen arbitrarily.
We project the bounding boxes onto the image space, obtaining \( \text{box}_s^{\text{(C)}}, \text{box}_d^{\text{(C)}} \in \mathbb{R}^{8 \times 2} \). Following the zoom-in strategy of AnyDoor~\cite{chen2023anydoor}, we crop and resize \( I_d \) to \( \mathbf{x}^{\text{(C)}} \in \mathbb{R}^{D \times D \times 3} \), centering it around \( \text{box}_d^{\text{(C)}} \), and apply the same viewport transformation to \( \text{box}_d^{\text{(C)}} \).
Following PbE~\cite{yang2023paint}, we encode the image \( \mathbf{x}^{\text{(C)}} \) using the pre-trained VAE~\cite{kingma2013auto} from StableDiffusion~\cite{rombach2022high}, obtaining the latent \( \mathbf{z}_0^{\text{(C)}} = \mathcal{E}^{\text{(C)}}(\mathbf{x}^{\text{(C)}}) \). Similarly, we obtain the latent representation of the camera context \( \mathbf{c}^{\text{(C)}}_{\text{env}} = \mathcal{E}^{\text{(C)}}(\mathbf{x}^{\text{(C)}} \odot \mathbf{\bar{m}}^{\text{(C)}}) \), where \( \odot \) denotes element-wise multiplication and the edit mask \( \mathbf{m}^{\text{(C)}} \in \{0, 1\}^{D \times D} \) is obtained by filling the projected bounding box $ \text{box}_d $ region with ones and \( \mathbf{\bar{m}}^{\text{(C)}} = 1 - \mathbf{m}^{\text{(C)}} \) is its complement.

\paragraph{Reference encoding and extraction}
We extract the reference image \( \mathbf{x}_{\text{ref}} \) from the source image \( I_s \) by cropping the minimal 2D bounding box that encompasses \( \text{box}_s^{\text{(C)}} \), capturing the object's features. During inference, the reference image can be obtained from external sources. 
Following PbE~\cite{yang2023paint}, we encode \( \mathbf{x}_{\text{ref}} \) using CLIP~\cite{radford2021learning}, selecting the classification token and passing it through linear adaptation layers, which are kept frozen during training. While CLIP effectively preserves high-level details such as gestures or car models, it lacks fine-detail preservation. For applications requiring finer details, other encoders like DINOv2~\cite{oquab2023dinov2} may be preferable, as demonstrated in~\cite{chen2023anydoor}.

\paragraph{Lidar encoding}
We process the destination scene's lidar point cloud \( P_d \in \mathbb{R}^{N \times 4} \) as follows, where each point includes \( x, y, z \) coordinates and intensity values. Using a lossless transformation (details in \cref{sec:suppl:method:lidar processing}), we project these points onto a range view \( R_d \in \mathbb{R}^{32 \times 1096 \times 2} \). The bounding box \( \text{box}_d \) is projected onto this range view, resulting in \( \text{box}_d^{\text{(R)}} \in \mathbb{R}^{8 \times 3} \), preserving depth information. To focus on the object of interest while retaining sufficient context, we employ a width-wise zoom-in strategy around \( \text{box}_d^{\text{(R)}} \), obtaining an object-centric range view, which we resize into the range image \( \mathbf{x}^{\text{(R)}} \in \mathbb{R}^{D \times D \times 2} \). The same viewport transformation is applied to \( \text{box}_d^{\text{(R)}} \). We define the edit mask \( \mathbf{m}^{\text{(R)}} \in \{0, 1\}^{D \times D} \) by filling the projected bounding box region with ones, and its complement is \( \mathbf{\bar{m}}^{\text{(R)}} = 1 - \mathbf{m}^{\text{(R)}} \).

We adapt the pre-trained image VAE~\cite{kingma2013auto} of StableDiffusion~\cite{rombach2022high} to the lidar modality through a series of adaptations—improved downsampling, intensity and depth normalisation, and fine-tuning of input and output adaptation layers—to achieve better object reconstruction. We demonstrate these in \cref{tab:lidar reconstruction} and provide more detail in \cref{sec:suppl:method:lidar processing}. We encode the range image into $ \mathbf{z}_0^{\text{(R)}} = \mathcal{E}^{\text{(R)}}(\text{norm}({\mathbf{x}^{\text{(R)}}})) $
and the range context into $ \mathbf{c}^{\text{(R)}}_{\text{env}} = \mathcal{E}^{\text{(R)}}(\text{norm}(\mathbf{x}^{\text{(R)}} \odot \mathbf{\bar{m}}^{\text{(R)}})) $.

\paragraph*{Bounding box encoding}
We consider the projected bounding boxes $ \text{box}_d^{\text{(C)}} \in \mathbb{R}^{8 \times 2} $ and $ \text{box}_d^{\text{(R)}} \in \mathbb{R}^{8 \times 3} $. The box $\text{box}_d^{\text{(C)}} $ captures the $ (x, y) $ coordinates in the camera view, scaled by the image dimensions; note some points may lie outside the image. The depth dimension from $ \text{box}_d^{\text{(R)}} $ is incorporated into $ \text{box}_d^{\text{(C)}} $ to aid with spatial consistency across modalities, resulting in $ \widetilde{\text{box}}_d^{\text{(C)}} \in \mathbb{R}^{8 \times 3} $.
We encode these bounding boxes into conditioning tokens $ \mathbf{c}_{\text{box}}^{\text{(C)}} $ and $ \mathbf{c}_{\text{box}}^{\text{(R)}} $ using Fourier embeddings, similar to MagicDrive~\cite{gao2023magicdrive}, and modality-agnostic trainable linear layers:
\begin{align*}
  \mathbf{c}_{\text{box}}^{\text{(M)}} = \text{MLP}_{\text{box}}(\text{Fourier}(\widetilde{\text{box}}_d^{\text{(M)}})), \quad \text{for~} \text{M} \in \{\text{C}, \text{R}\}.
\end{align*}

\subsection{Multimodal generation}
\label{sec:method:multimodal generation}
We finetune a single latent diffusion model for both modalities, leveraging the pre-trained weights of PbE~\cite{yang2023paint}.
Similar to the adaptation strategy of Flamingo~\cite{alayrac2022flamingo}, we interleave separate gated cross-attention layers: a modality-agnostic bounding box adapter and modality-dependent cross-modal attention. The use of such layers is a commonly used strategy for methods in scene generation~\cite{gao2023magicdrive, xie2024x} and we use a zero-initialised gating as in ControlNet~\cite{zhang2023controlnet}.

\paragraph*{Cross-modal attention}
We introduce a modality-dependent cross-modal attention which looks at the tokens of the other modality from the same scene in the batch.
We derive the query, key, and value representations from the input camera and lidar features with layer normalisation applied for cross-attention from camera to lidar. Using learnable transformations \( W_Q^{\text{(C)}}, W_K^{\text{(R)}}, W_V^{\text{(R)}} \), we compute the cross-attention as:
\(
\text{Attn}^{\text{(C)}} = \text{softmax}\left({Q^{\text{(C)}} (K^{\text{(R)}})^T}/{\sqrt{d_{\text{head}}}}\right) V^{\text{(R)}},
\)
where \( Q^{\text{(C)}} = W_Q^{\text{(C)}} \mathbf{h}^{\text{(C)}} \), \( K^{\text{(R)}} = W_K^{\text{(R)}} \mathbf{h}^{\text{(R)}} \), and \( V^{\text{(R)}} = W_V^{\text{(R)}} \mathbf{h}^{\text{(R)}} \). We then update the camera features by adding a residual connection through a zero-initialised gating module:
\(
\mathbf{h}^{\text{(C)}} \leftarrow \mathbf{h}^{\text{(C)}} + \text{Gate}^{\text{(C)}}(\text{Attn}^{\text{(C)}}).
\)
The computation for lidar-to-camera cross-attention is analogous, with lidar features attending to the camera modality. We do not restrict the cross-modal attention and let the network learn an implicit correspondence which is facilitated by the respective projected bounding boxes. Lastly, we concatenate the camera and lidar tokens within the batch.

\paragraph*{Bounding box adapter}
The bounding box adapter is a modality-agnostic layer designed to provide bounding box conditioning while preserving reference features encoded in $\mathbf{c}_{\text{ref}}$. This adapter employs the same gating mechanism as the cross-attention module but instead is conditioned on one of the bounding box tokens \( \mathbf{c}_{\text{box}}^{\text{(R)}} \) or \( \mathbf{c}_{\text{box}}^{\text{(C)}} \), depending on the modality, and the reference token \( \mathbf{c}_{\text{ref}} \). This enables flexible conditioning across modalities, ensuring that spatial information from the bounding box is effectively integrated alongside the reference features. We employ classifier-free guidance~\cite{ho2022classifier} with a scale of 5  as in PbE~\cite{yang2023paint}, extending it to both reference and bounding box conditioning.

\subsection{Inference and compositing}
\label{sec:method:spatial compositing}
\paragraph*{Inference process}
At inference, we start from random noise \( \mathbf{\epsilon} \sim \mathcal{N}(0, \mathbf{I}) \) combined with the latent scene context and resized edit mask, and iteratively denoise this input for \( T = 50 \) steps using the PLMS scheduler~\cite{liu2022pseudo}, conditioned on the reference \( \mathbf{c}_{\text{ref}} \) and 3D bounding box token \( \mathbf{c}_{\text{box}} \), to yield the final latent representations \( \{ \tilde{\mathbf{z}}_0^{(\text{C})}, \tilde{\mathbf{z}}_0^{(\text{R})}\} \). These latent representations are then decoded by the image and range decoders to produce the edited camera and range images \( \tilde{\mathbf{x}}^{(\text{C})} = \mathcal{D}^{(\text{C})}(\tilde{\mathbf{z}}_0^{(\text{C})}) \) and \( \tilde{\mathbf{x}}^{(\text{R})} = \mathcal{D}^{(\text{R})}(\tilde{\mathbf{z}}_0^{(\text{R})}) \).

\paragraph*{Spatial compositing}
Final results are obtained by compositing the edited camera and range images back into the original scene.
For images, we extract the region within the projected bounding box from the edited image $\tilde{\mathbf{x}}^{(\text{C})}$ and insert it back into the destination image \( I_d \). Following the approach of POC~\cite{de2024placing}, a Gaussian kernel is applied to improve blending, resulting in the final composited image.
For lidar, we create a 2D mask \( \mathbf{m}_{\text{points}} \) by selecting points from the original lidar point cloud \( P_d \) that fall within the destination 3D bounding box. The edited range image \( \mathbf{\tilde{x}}^{\text{(R)}} \) is resized to an object-centric range view using average pooling and denormalised before computing coordinate and intensity values (see \cref{sec:suppl:method:lidar processing}). We replace pixels in the original range view \( R_d \) with the corresponding pixels from the edited range image if either (i) they fall within \( \mathbf{m}_{\text{points}} \) or (ii) its corresponding 3D point in the edited range image is contained by the bounding box of the object, as seen in \cref{fig:compositing}.

\subsection{Training details}
\label{sec:method:training details}

\paragraph{Sample selection}
We consider objects from the nuScenes dataset~\cite{caesar2020nuscenes} train split with at least 64 lidar points, whose 2D bounding box is at least $100 \times 100$ pixels, with a 2D IoU overlap not exceeding 50\% with other objects, and current camera visibility of at least 70\%.
Unless stated otherwise, our model is trained on ``car'' and ``pedestrian'' categories, dynamically sampling 4096 new actors per class each epoch.
During training, once an object is selected, its current scene serves as the destination, from which we extract the 3D bounding box, environmental context, and ground truth insertion.

\paragraph{Reference selection}
Object references are taken from the same object at a different timestamp, picked randomly as follows.
We collect references for the current object across all frames that meet the previous criteria to ensure good visibility and arrange them by normalised temporal distance $\Delta t$, where $1$ represents the furthest reference in time and $0$ represents the current one. 
We then sample references randomly using a beta distribution $\Delta t \sim \text{Beta}(4, 1)$ which ensures a preference for instances of the object that are far away from the current timestamp, see \cref{sec:suppl:method} for details.
\paragraph{Augmentation}
During training, the reference image undergoes augmentations similar to those described in PbE~\cite{yang2023paint}, such as random flip, rotation, blurring and brightness and contrast transformations.
Additionally, we randomly sample empty bounding boxes (i.e., containing no objects) overriding both the reference image and bounding box with zero values.
This encourages the model to infer and reconstruct missing details based on surrounding context alone. Further details are provided in \cref{sec:suppl:method}.

\paragraph{Fine-tuning procedure}
During fine-tuning, the autoencoders and all other layers from the PbE~\cite{yang2023paint} framework remain frozen, while only the bounding box encoder, bounding box adaptation layer, and cross-modal attention layers are trained. We use an input dimension of \( D = 512 \) and a latent dimension of \( D_h = 64 \), training for 30 epochs and retain the top five models with the lowest loss. The final model is selected based on the best Fréchet Inception Distance (FID)~\cite{heusel2017gans} achieved on a test set of 200 pre-selected images, where objects are reinserted into scenes using the previously-described filters. See \cref{sec:suppl:method:training details} for details.

\section{Experiments}
\label{sec:experiments}

\subsection{Object insertion and replacement}
\label{sec:experiments:implementation}

\paragraph{Setup}
To avoid situations where inpainted objects are placed at locations incompatible with the scene (e.g. a car on pavement), we use the position of existing objects and perform either object reinsertion or replacement, which differ by the choice of reference.
This tests the model's ability to generate realistic objects conditioned on a 3D bounding box while being semantically consistent with the scene.
We sample 200 high-quality objects from the nuScenes val set as in~\cref{sec:method:training details}, balanced across ``car'' and ``pedestrian''.

\paragraph{Reinsertion}
We define two types of references: \textit{same reference}, where the source and destination images and bounding boxes are identical, meaning the object is reinserted in the same scene and position; and \textit{tracked reference}, where the object is reinserted given its reference from a different timestamp, using the sampling strategy described in \cref{sec:method:training details}. This setting tests if the model can preserve the object’s appearance, and realistically perform novel view synthesis (for \textit{tracked reference}).

\paragraph{Replacement}
We define two different domains based on the weather conditions ($\text{rainy}(I_s), \text{rainy}(I_d) \in \{0, 1\}$) and time of day ($\text{night}(I_s), \text{night}(I_d) \in \{0, 1\}$), and consider the following reference types: \textit{in-domain reference}, where the source and destination bounding boxes correspond to different objects that are of the same class and same domain $(\text{rainy}(I_d)=\text{rainy}(I'_d) ~\&~ \text{night}(I_s)=\text{night}(I'_d))$, and \textit{cross-domain reference}, where the bounding boxes correspond to different objects of the same class, yet draw from at least a different domain $(\text{rainy}(I_d) \neq \text{rainy}(I_d) \text{ or } \text{night}(I_s) \neq \text{night}(I_d))$. 
We select replacements within the same class only to make sure that object placement and dimensions are meaningful.

\begin{figure*}[ht!]
\centering
\setlength{\tabcolsep}{2pt}
\begin{tabular}{ccccccc}
& \multicolumn{3}{c}{\textbf{Original}} & \multicolumn{3}{c}{\textbf{Edited}}\\
\cmidrule(r){2-4} \cmidrule(r){5-7}
Reference & Camera & Depth & Intensity & Camera & Depth & Intensity \\
\cmidrule(r){1-1} \cmidrule(r){2-7}\\[-10pt]
\includegraphics[width=0.135\linewidth]{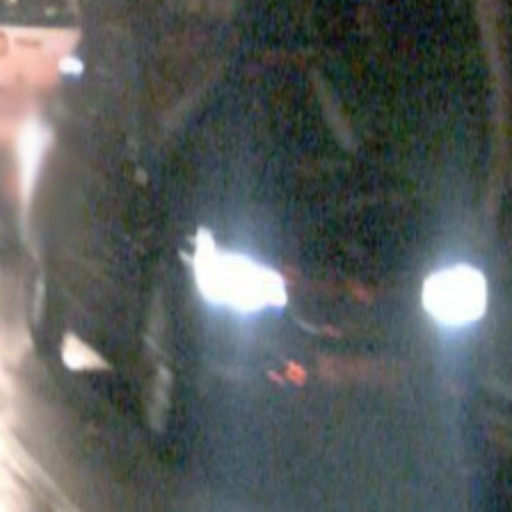} &
\includegraphics[width=0.135\linewidth]{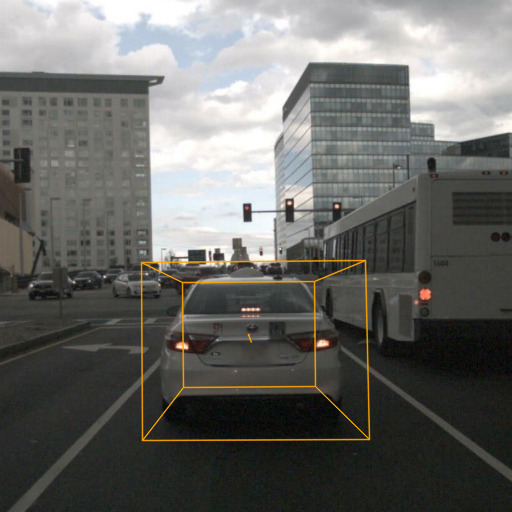} &
\includegraphics[width=0.135\linewidth]{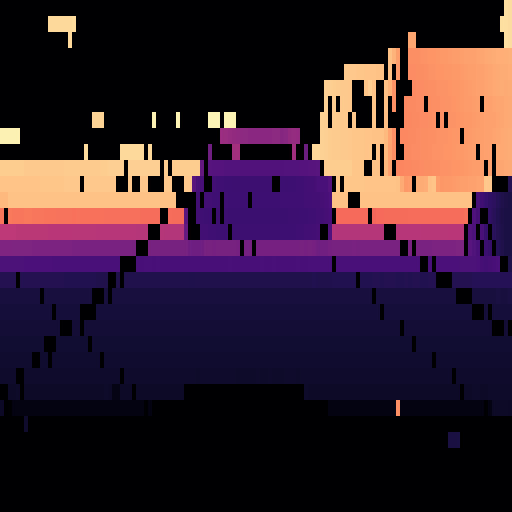} &
\includegraphics[width=0.135\linewidth]{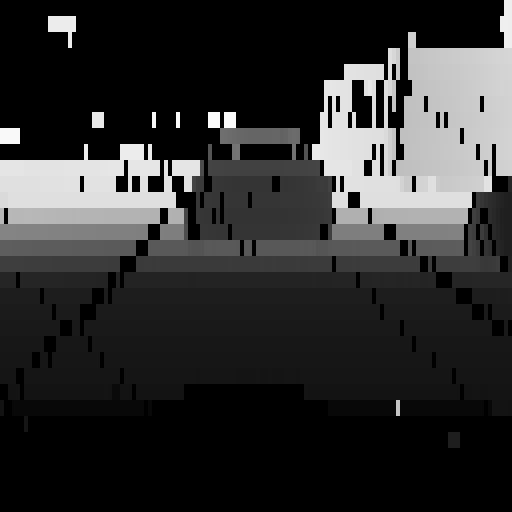} &
\includegraphics[width=0.135\linewidth]{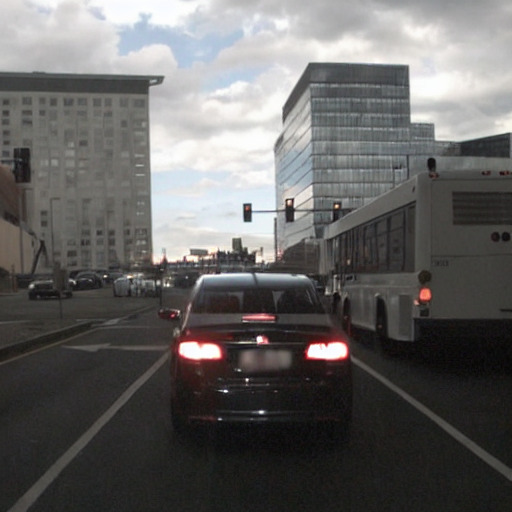} &
\includegraphics[width=0.135\linewidth]{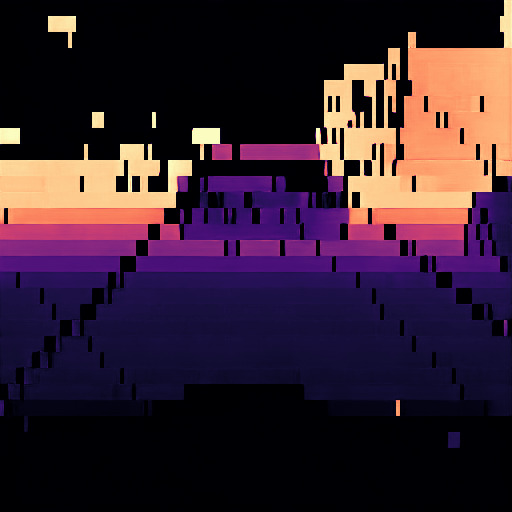} &
\includegraphics[width=0.135\linewidth]{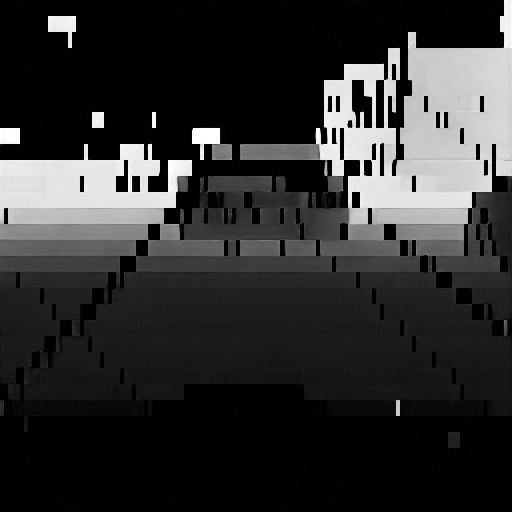} \\
\includegraphics[width=0.135\linewidth]{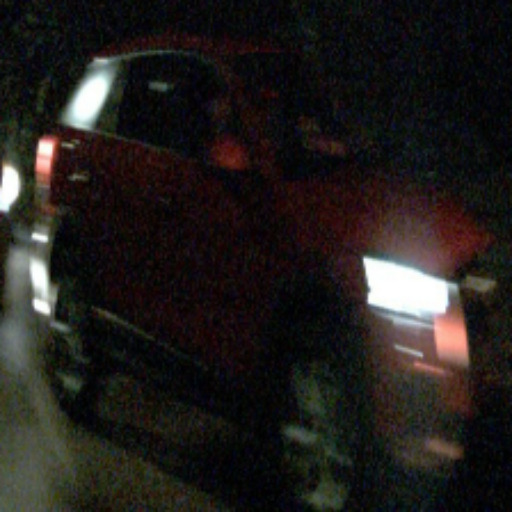} &
\includegraphics[width=0.135\linewidth]{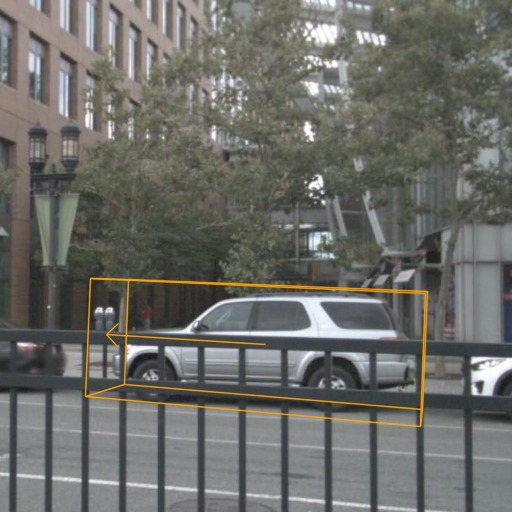} &
\includegraphics[width=0.135\linewidth]{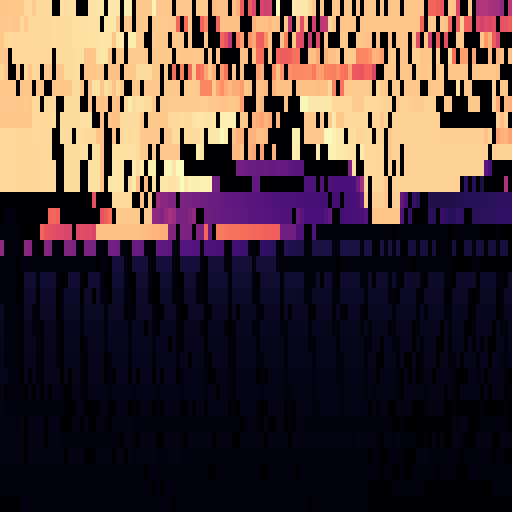} &
\includegraphics[width=0.135\linewidth]{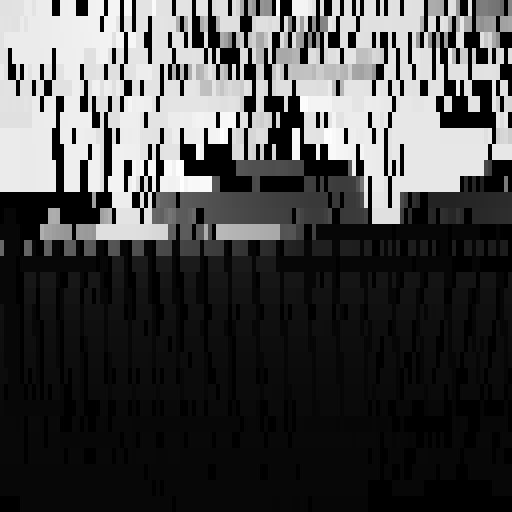} &
\includegraphics[width=0.135\linewidth]{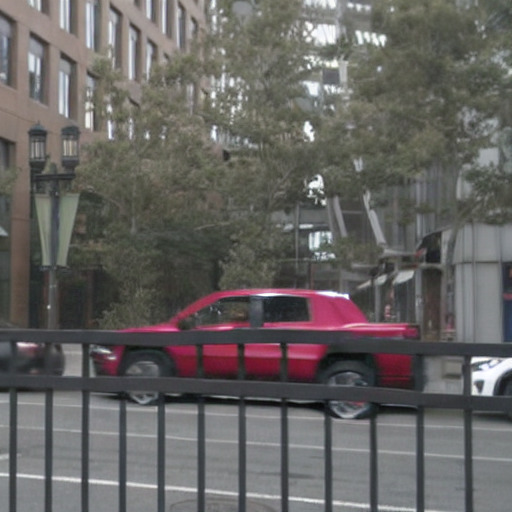} &
\includegraphics[width=0.135\linewidth]{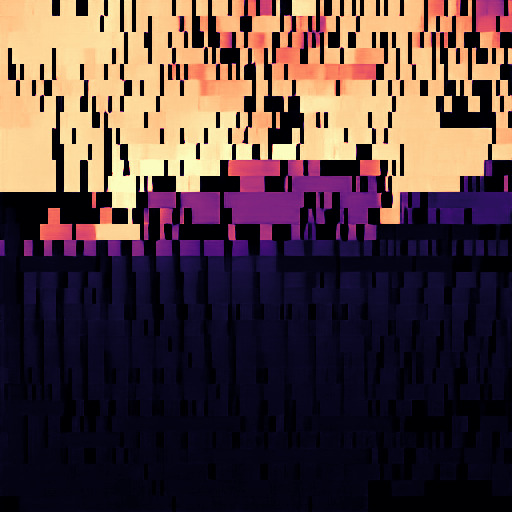} &
\includegraphics[width=0.135\linewidth]{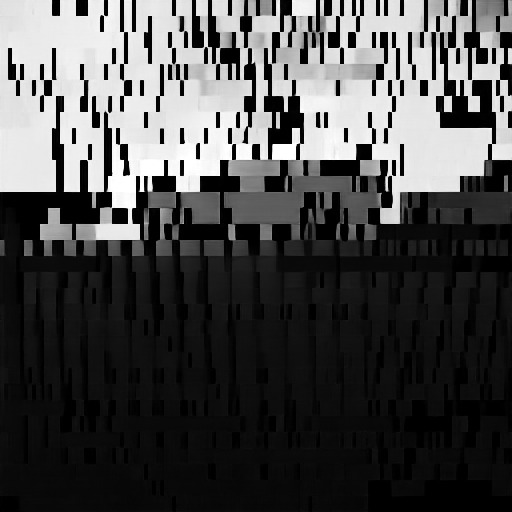} \\
\includegraphics[width=0.135\linewidth]{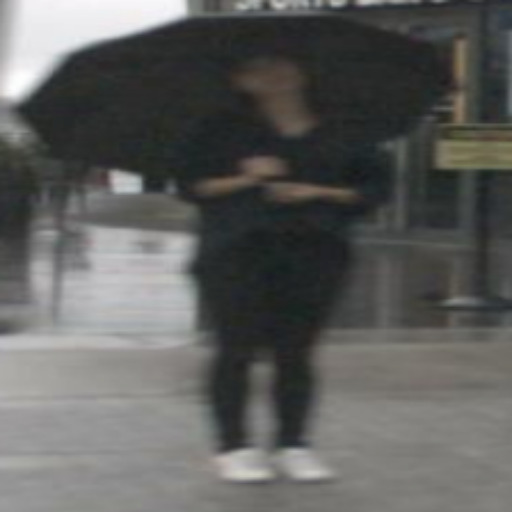} &
\includegraphics[width=0.135\linewidth]{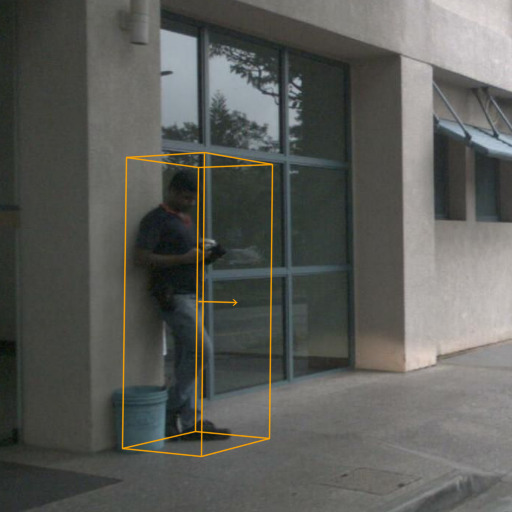} &
\includegraphics[width=0.135\linewidth]{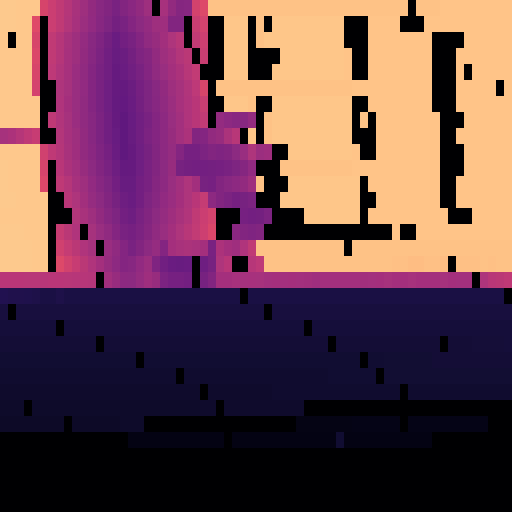} &
\includegraphics[width=0.135\linewidth]{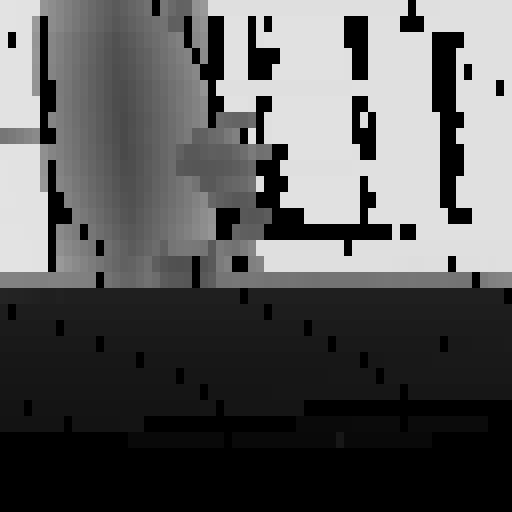} &
\includegraphics[width=0.135\linewidth]{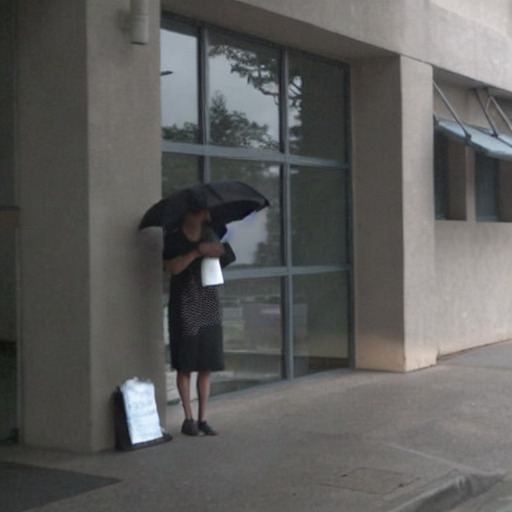} &
\includegraphics[width=0.135\linewidth]{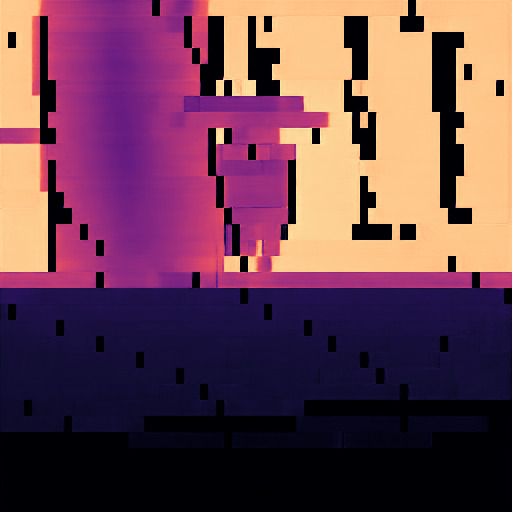} &
\includegraphics[width=0.135\linewidth]{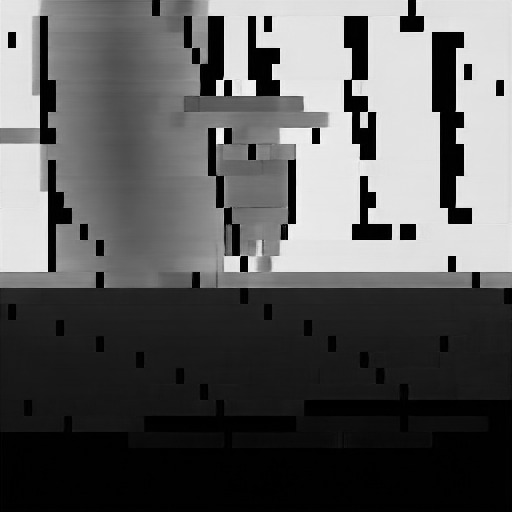} \\
\includegraphics[width=0.135\linewidth]{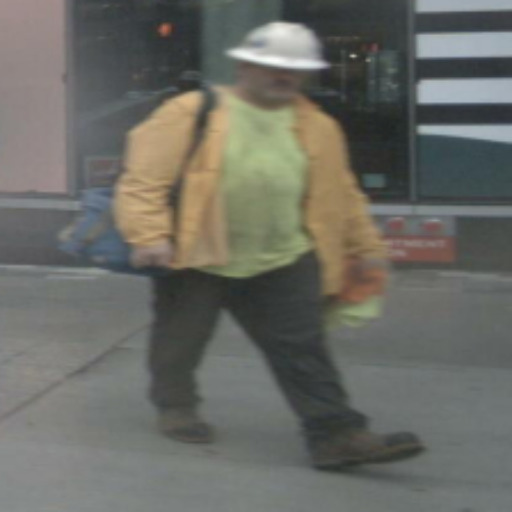} &
\includegraphics[width=0.135\linewidth]{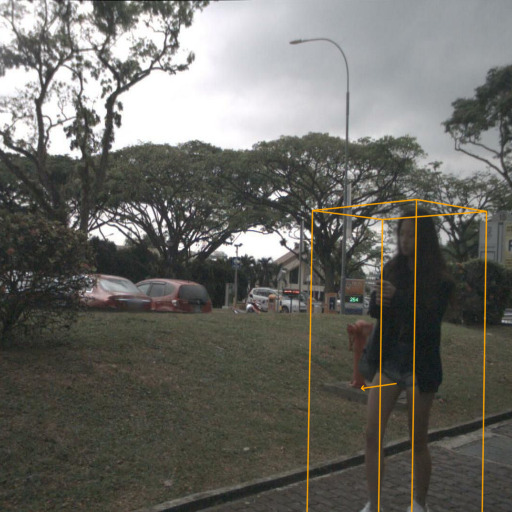} &
\includegraphics[width=0.135\linewidth]{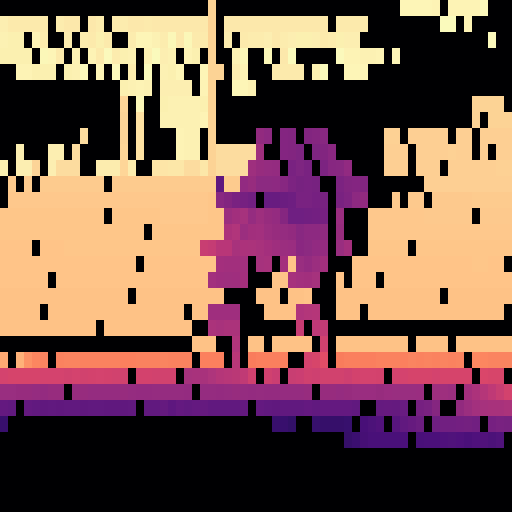} &
\includegraphics[width=0.135\linewidth]{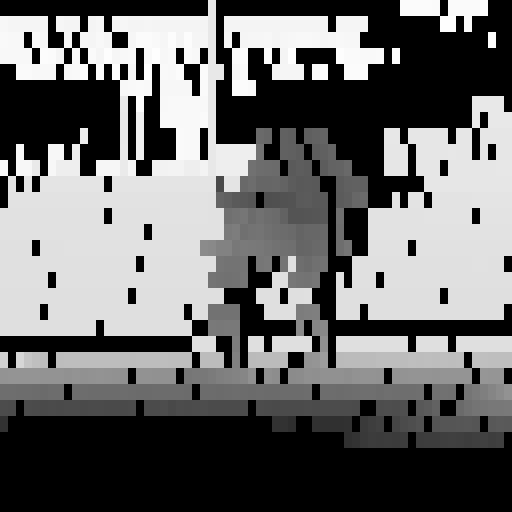} &
\includegraphics[width=0.135\linewidth]{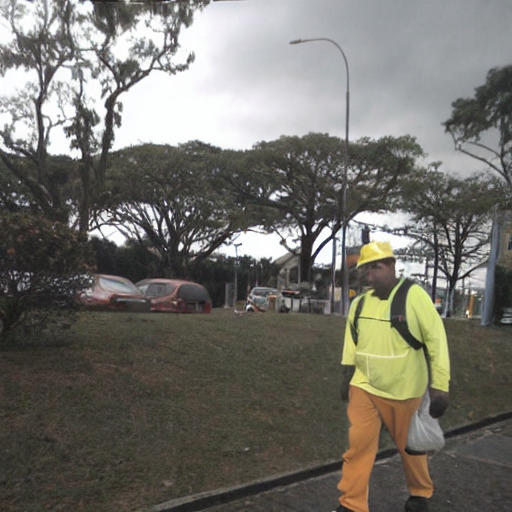} &
\includegraphics[width=0.135\linewidth]{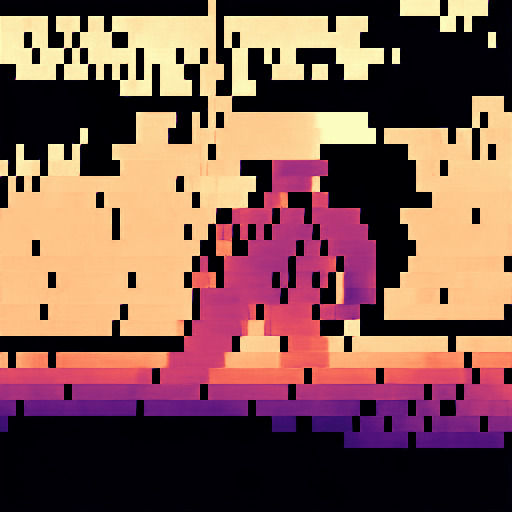} &
\includegraphics[width=0.135\linewidth]{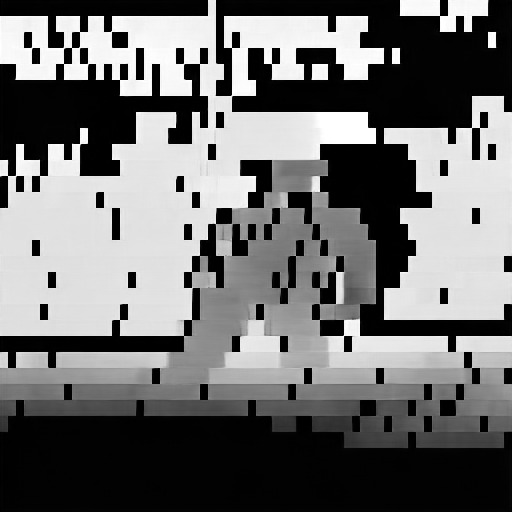} \\
\includegraphics[width=0.135\linewidth]{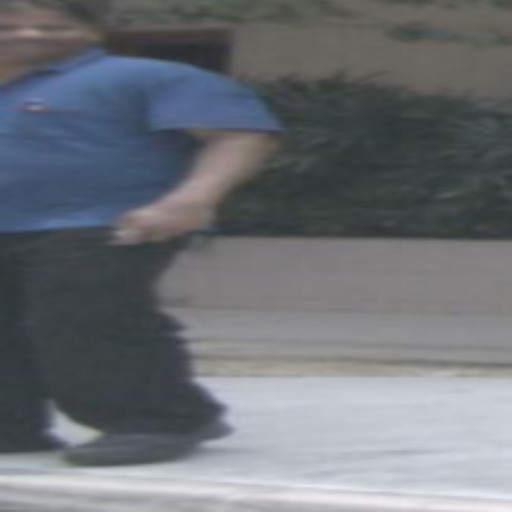} &
\includegraphics[width=0.135\linewidth]{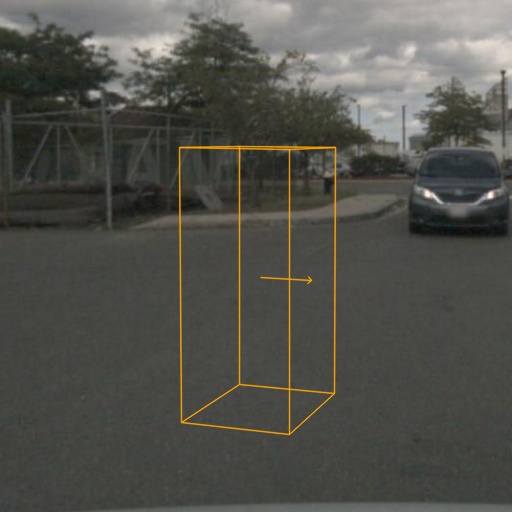} &
\includegraphics[width=0.135\linewidth]{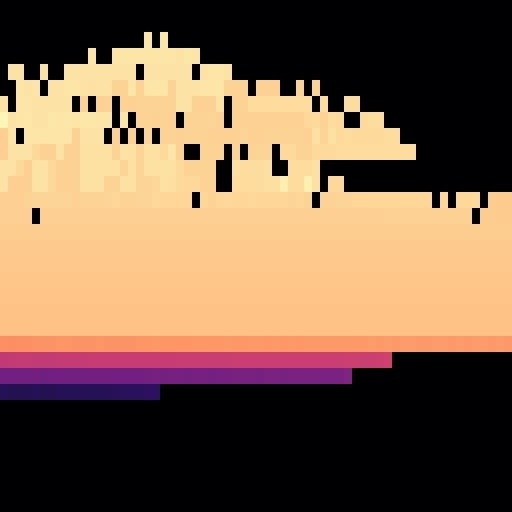} &
\includegraphics[width=0.135\linewidth]{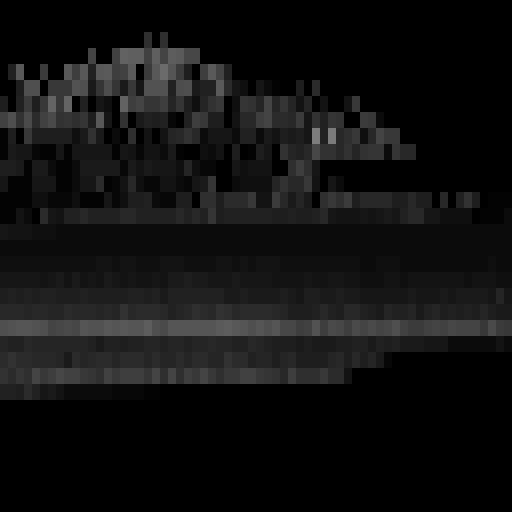} &
\includegraphics[width=0.135\linewidth]{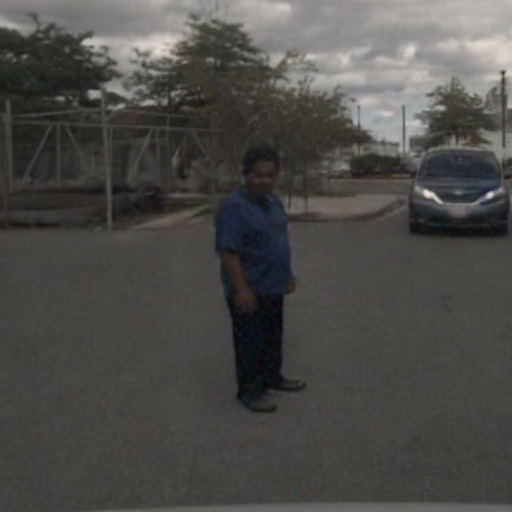} &
\includegraphics[width=0.135\linewidth]{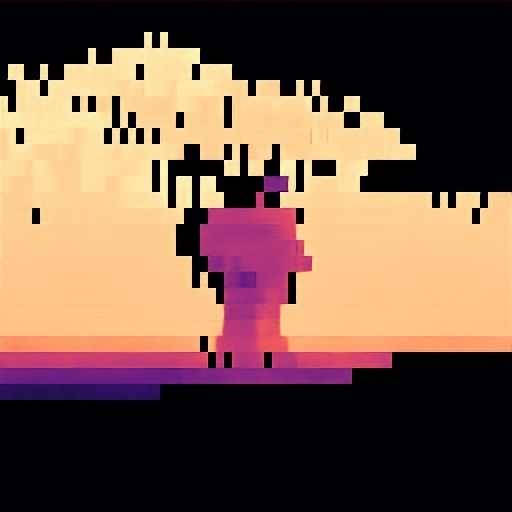} &
\includegraphics[width=0.135\linewidth]{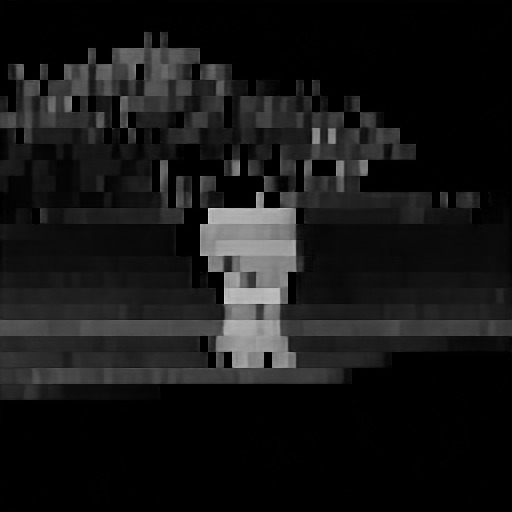} \\
\includegraphics[width=0.135\linewidth]{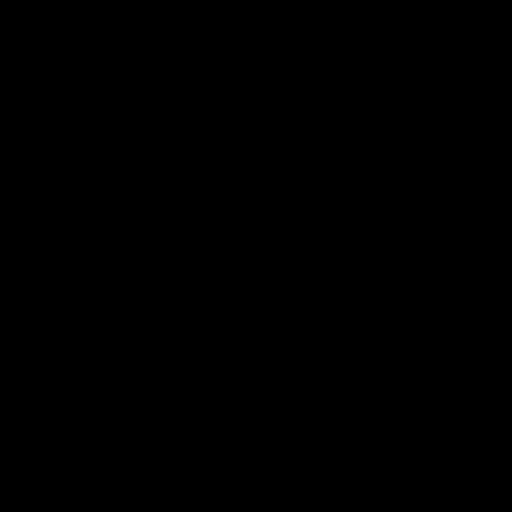} &
\includegraphics[width=0.135\linewidth]{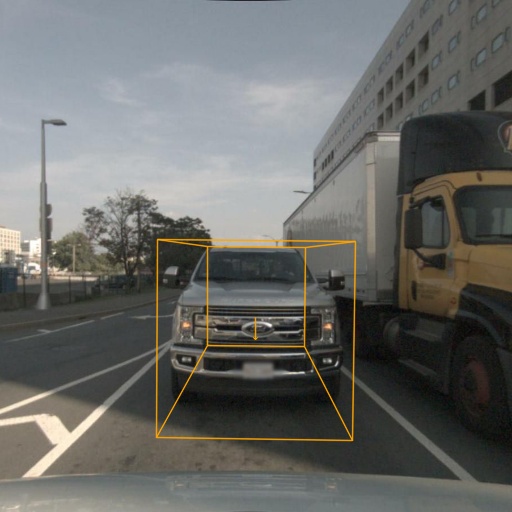} &
\includegraphics[width=0.135\linewidth]{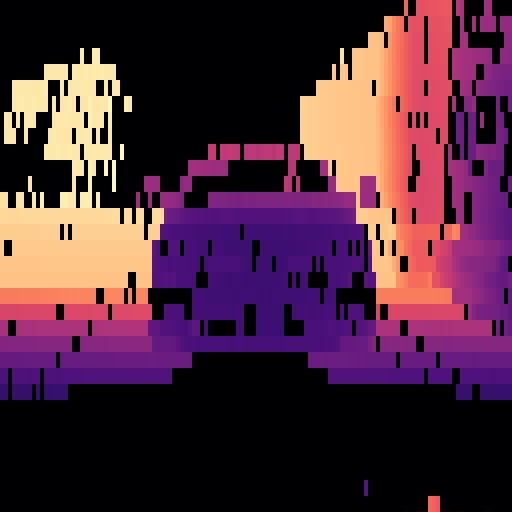} &
\includegraphics[width=0.135\linewidth]{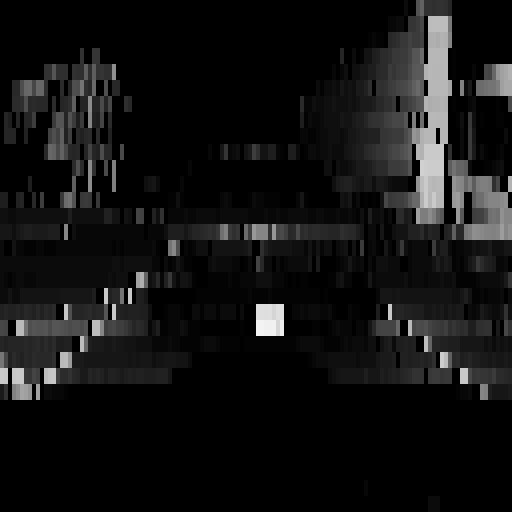} &
\includegraphics[width=0.135\linewidth]{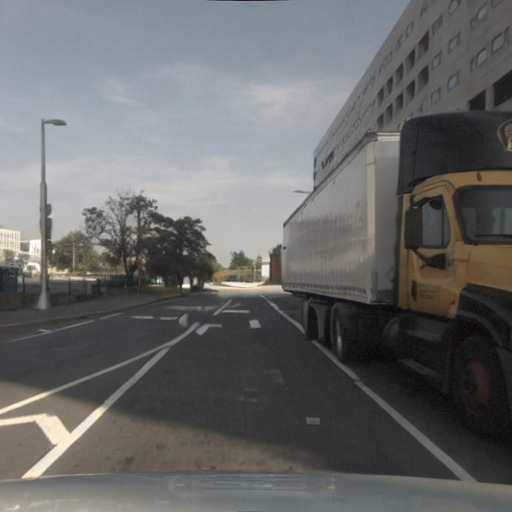} &
\includegraphics[width=0.135\linewidth]{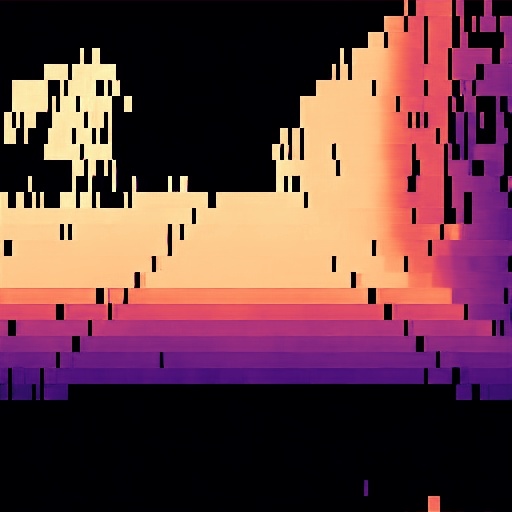} &
\includegraphics[width=0.135\linewidth]{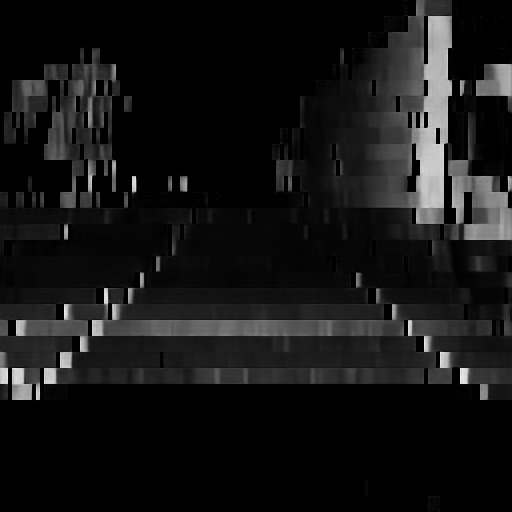} \\
\end{tabular}
\caption{Examples of object inpainting using MObI in the following settings: replacement (rows 1--4), insertion (row 5), and deletion (row 6, using a black reference). Our model can inpaint objects corresponding to a 3D bounding box with a high degree of realism while preserving coherence with the rest of the scene.
Note that even though some references are from a different domain (time of day, weather condition), the model is able to preserve coherence of the resulting insertion.
}
\label{fig:results}
\end{figure*}

\begin{figure}[t!]
  \centering
  \footnotesize
  \renewcommand{\arraystretch}{0.3} 
  \hspace*{-11pt}\mbox{
  \begin{tabular}{c@{\hspace{1pt}}c@{\hspace{1pt}}c@{\hspace{1pt}}c@{\hspace{1pt}}c}
      {\textbf{Ref. image}} &
      {\textbf{Original}} &
      \multicolumn{3}{c}{\textbf{Edited scenes}}
      \\[4pt]
      
      \includegraphics[width=0.2\linewidth]{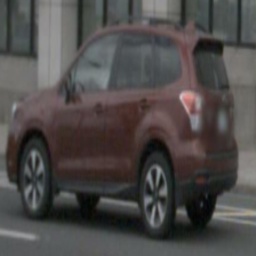} &
      \includegraphics[width=0.2\linewidth]{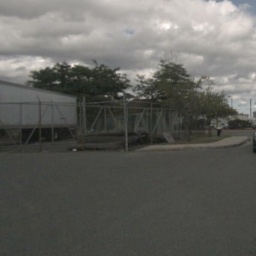} &
      \includegraphics[width=0.2\linewidth]{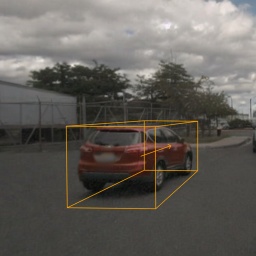} &
      \includegraphics[width=0.2\linewidth]{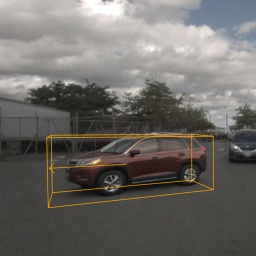} &
      \includegraphics[width=0.2\linewidth]{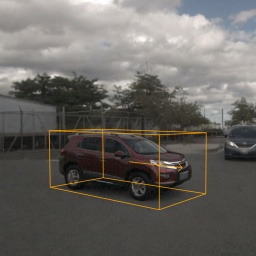} \\[1pt]
      
      &
      \includegraphics[width=0.2\linewidth]{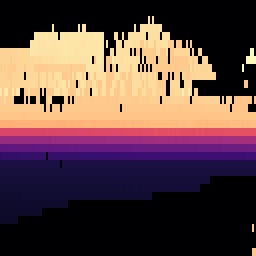} &
      \includegraphics[width=0.2\linewidth]{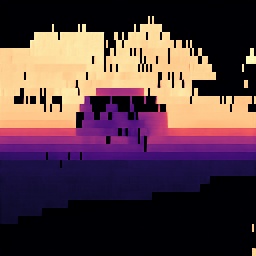} &
      \includegraphics[width=0.2\linewidth]{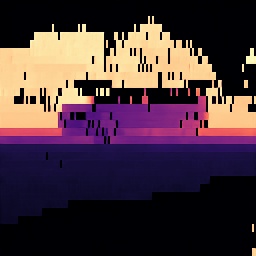} &
      \includegraphics[width=0.2\linewidth]{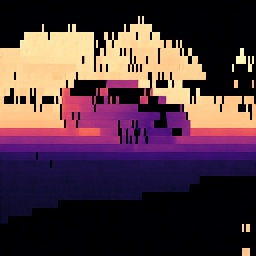} \\[3pt]

      \includegraphics[width=0.2\linewidth]{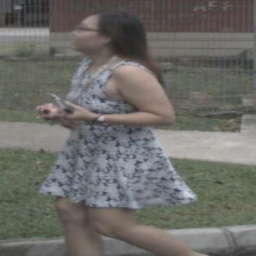} &
      \includegraphics[width=0.2\linewidth]{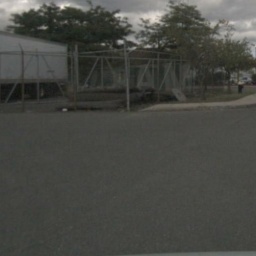} &
      \includegraphics[width=0.2\linewidth]{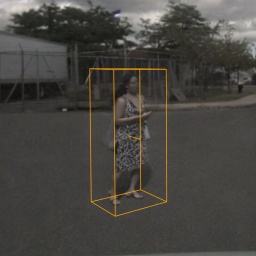} &
      \includegraphics[width=0.2\linewidth]{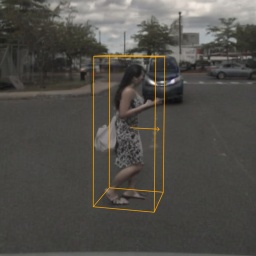} &
      \includegraphics[width=0.2\linewidth]{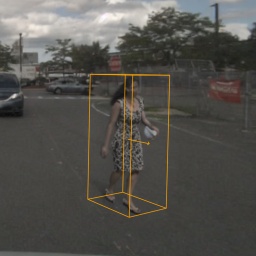} \\[1pt]
      
      &
      \includegraphics[width=0.2\linewidth]{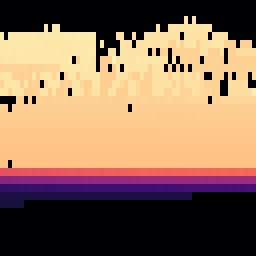} &
      \includegraphics[width=0.2\linewidth]{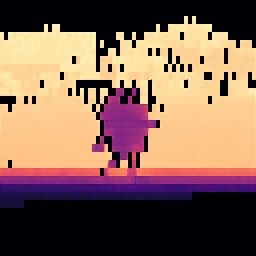} &
      \includegraphics[width=0.2\linewidth]{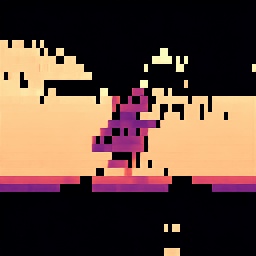} &
      \includegraphics[width=0.2\linewidth]{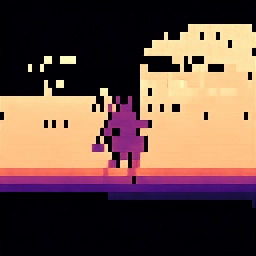} \\[1pt]
  \end{tabular}
  }
  \caption{
  Our method can generate multiple novel views from a single reference image while maintaining multimodal consistency. 
  From left to right: reference image $ \mathbf{x}_{\text{ref}} $, extracted from a separate scene; original destination scene with the RGB image $ \mathbf{x}^{\text{(C)}} $ and lidar range depth $ \mathbf{x}_0^{\text{(R)}} $; and edited scenes. Note, the inpainted pedestrian moves to the right between frames, shifting the background to the left. Check \cref{fig:suppl:rotation_results} for extended results, including intensity.
  }
  \label{fig:controllability}
\end{figure}

\begin{figure}[t!]
    \begin{center}
        \includegraphics[width=\linewidth, trim={0 24pt 0 66pt}, clip]{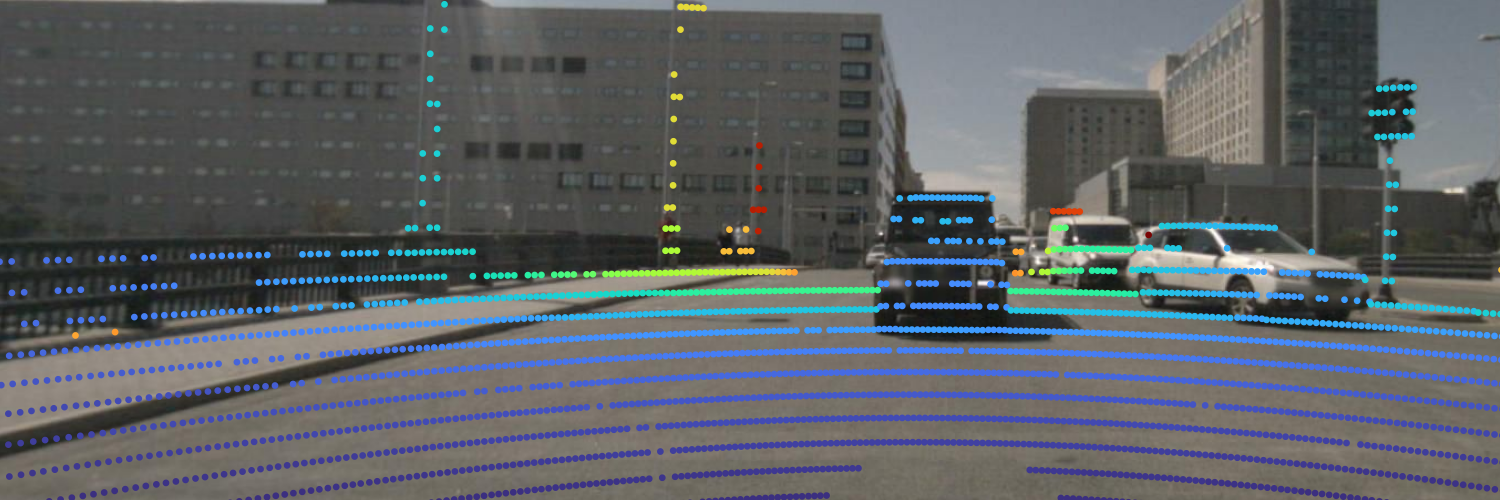}
        \includegraphics[width=\linewidth, trim={0 24pt 0 68pt}, clip]{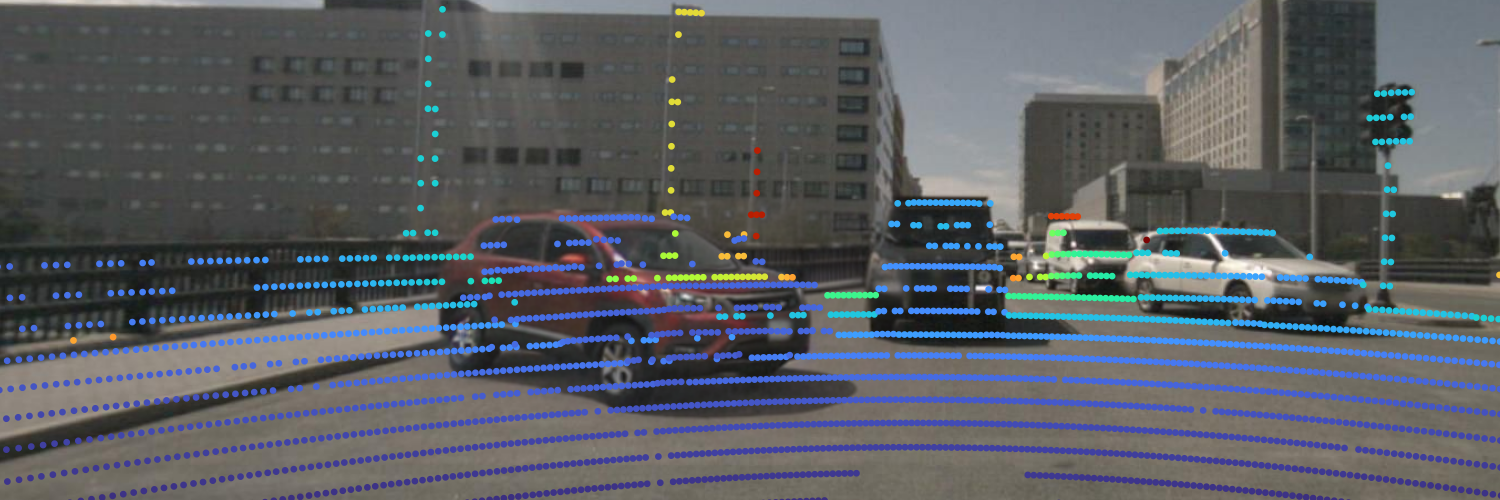}
    \end{center}
    \caption{Spatial compositing of camera-lidar object inpainting. Note that some background points are not overridden due to lidar reflections on the hood of the inserted car (bottom).}
    \label{fig:compositing}
\end{figure}

\paragraph{Qualitative results}
Results are presented in \cref{fig:results} both for replacement (rows 1--4) and insertion (row 5).
We see that inpainted objects correspond tightly to their conditioning 3D bounding boxes while having a high degree of realism, both for camera (RGB) and lidar (depth and intensity), and show a strong coherence (lightning, weather conditions, occlusions, etc.) with the rest of the scene.
The last row showcases object deletion, which can be achieved by using an empty reference image (note that we use empty references during training, as described in~\cref{sec:method:training details}).
Even though references in the replacement setting are from a different domain (time of day/weather), the model is able to inpaint such objects realistically. See \cref{fig:inpainting-hard-suppl} for more examples, including failure cases. We show an example of the composited camera and lidar scene in \cref{fig:compositing}.
We illustrate the flexibility of our bounding box conditioning and show that it is able to generate multiple views with a high degree of consistency, as illustrated in \cref{fig:controllability}, \cref{fig:suppl:rotation_results} and \cref{fig:suppl:controllability_main_full}.

\begin{table}[h]
    \centering
    \scriptsize
    \begin{tabular}{l
                S[table-format=1.3] 
                S[table-format=1.3] 
                S[table-format=1.3] 
                S[table-format=1.3]}
        & \multicolumn{2}{c}{\textbf{Median depth error}} & \multicolumn{2}{c}{\textbf{MSE intensity}} \\
        \cmidrule(r){2-3} \cmidrule(r){4-5}
        \textbf{Lidar encoder} & {object} & {mask} & {object} & {mask} \\
        \midrule
        {pretrained image VAE~\cite{kingma2013auto}} & 0.451 & 0.320 & 7.854 & 7.372 \\
        {+ average pooling} & 0.306 & 0.263 & 3.496 & 3.236 \\
        {+ object-aware norm.} & 0.04 & 0.315 & 3.792 & 2.941 \\
        {+ fine-tune lidar adapter} & \textbf{0.037} & \textbf{0.180}  & \textbf{2.397} & \textbf{2.009} \\
        \bottomrule
        \\
    \end{tabular}
    \caption{Adaptations of the pre-trained image VAE~\cite{kingma2013auto} from StableDiffusion~\cite{rombach2022high} improving lidar reconstruction for depth (meters) and intensity (on a scale of $ [0, 255] $).}
    \label{tab:lidar reconstruction}
\end{table}

\subsection{Realism of the inpainting} 
\label{sec:experiments:realism}

\paragraph{Camera realism metrics}
We evaluate the realism of the camera inpainting using the following metrics: Fréchet Inception Distance (FID)~\cite{heusel2017gans}, Learned Perceptual Image Patch Similarity (LPIPS)~\cite{zhang2018unreasonable}, and CLIP-I~\cite{ruiz2023dreambooth}.
By computing the cosine similarity, the CLIP-I score evaluates how well the inpainted object matches the reference in terms of semantics and high-level details which the CLIP image encoder~\cite{radford2021learning} captures.
FID measures the realism of inpainted object patches compared to the real ones by looking at the feature distribution.
LPIPS measures a learned similarity between the feature maps of the inpainted patch and the ground truth patch, capturing differences across multiple levels of a deep neural network.
For FID and LPIPS we consider extended patches around the object from the final composited images, compared to the real patches.
For CLIP-I, we only consider the region within the bounding box of the inpainted object and the reference image. 

\paragraph{Lidar realism metrics}
To the best of our knowledge, metrics specifically designed for lidar editing, particularly those capable of capturing fine perceptual differences, are not available. 
Existing metrics based on the Fréchet distance~\cite{lidardiffusion, lidargen, nakashima2024lidar} operate on the full lidar point clouds and lack the granularity necessary to detect fine object-level differences which are essential for actor insertion and detailed editing tasks.
We therefore assess the differences in depth and intensity between the original and inpainted range images by using LPIPS~\cite{zhang2018unreasonable} on rasterized patches, resulting in the following adapted distances: \( \text{D-LPIPS}( \mathbf{x}^{\text{(R)}}_0, \tilde{\mathbf{x}}^{\text{(R)}}_0) \) for depth and \( \text{I-LPIPS}( \mathbf{x}^{\text{(R)}}_1, \tilde{\mathbf{x}}^{\text{(R)}}_1) \) for intensity.
We use the output of the diffusion model (after the range decoder) which is normalised between 0 and 1, and tile both depth and intensity 3 times to create an RGB input for LPIPS. We report individual scores for depth and intensity by averaging the corresponding perceptual distances across all patch pairs.

\paragraph{Results}
We report all realism metrics for camera-lidar object inpainting in \cref{tab:full-realism} for the reinsertion and replacement settings. Compared to camera-only inpainting methods, MObI ($D=512$) achieves better results than PbE~\cite{yang2023paint} across almost all benchmarks. We note that this method achieves competitive results in terms of FID, producing samples which are close in distribution to the target ones, yet LPIPS is much worse. This perceptual misalignment, which is more severe than even MObI (256) with no bounding box conditioning, could indicate that the use of joint generation of camera and lidar improves semantic consistency within the scene.
We compare to simple copy\&paste and show that this produces unrealistic composited images when replacing objects, even though it is sometimes used for training object detectors \cite{georgakis2017synthesizing, dwibedi2017cut, wang2021pointaugmenting, zhang2020exploring}. Note that object reinsertion results for copy\&paste, as well as CLIP-I scores, are not computed as the comparison would be unfair in this setting. We provide additional realism metrics in \cref{tab:supp:camera-realism-comparison}

We provide ablations for the 3D bounding box and the gated cross-attention adapter for $D=256$. Without the adapter, the box token is concatenated with the reference token in the PbE~\cite{yang2023paint} cross-attention layer, followed by direct fine-tuning. Due to the lack of a baseline for lidar object inpainting realism, we provide comparative results for all experiments and ablation in the hope that this could constitute a baseline for future work. Comparing MObI (256) with both bbox conditioning and adapter to the one without bbox conditioning we notice significant improvements in perceptual alignment. Gated cross-attention leads to more realistic samples in the camera space, yet it does not improve lidar, hinting towards differences in training regimes for the two modalities.
Finally, we note that realism scales strongly with resolution, leading us to believe that models operating at larger resolution would improve realism even more.

\begin{table*}
    \centering
    \scriptsize
    \begin{tabular}{cccccccccccccc}
        \multicolumn{3}{c}{} & \multicolumn{5}{c}{\textbf{Reinsertion}} & \multicolumn{5}{c}{\textbf{Replacement}} \\
        \cmidrule(r){4-8}  \cmidrule(r){9-13}
        \multicolumn{3}{c}{} & \multicolumn{3}{c}{\textbf{Camera Realism}} & \multicolumn{2}{c}{\textbf{Lidar Realism}} & \multicolumn{3}{c}{\textbf{Camera Realism}} & \multicolumn{2}{c}{\textbf{Lidar Realism}} \\
        \cmidrule(r){4-6} \cmidrule(r){7-8} \cmidrule(r){9-11} \cmidrule(r){12-13}
        \textbf{Model} & 3D Box & Adapter & FID\textdownarrow & LPIPS\textdownarrow & CLIP-I\textuparrow & D-LPIPS\textdownarrow & I-LPIPS\textdownarrow & FID\textdownarrow & LPIPS\textdownarrow & CLIP-I\textuparrow & D-LPIPS\textdownarrow & I-LPIPS\textdownarrow \\
        \midrule
        copy\&paste & \multicolumn{2}{c}{n/a} & \multicolumn{3}{c}{n/a} & \multicolumn{2}{c}{n/a} & 15.29 & 0.205 & n/a & \multicolumn{2}{c}{n/a} \\
        PbE~\cite{yang2023paint} & \multicolumn{2}{c}{n/a} & \underline{7.46} & 0.133 & \underline{83.91} & \multicolumn{2}{c}{n/a} & 10.08 & 0.149 & \textbf{77.25} & \multicolumn{2}{c}{n/a}\\
        \midrule
        & \ding{55} & \checkmark & 8.18 & 0.123 & 82.56 & 0.195 & 0.231 & 10.31 & 0.140 & \underline{77.22} & 0.198 & \underline{0.236} \\
        MObI (256)& \checkmark & \ding{55} & 8.31 & 0.120 & 82.88 & \underline{0.188} & 0.231 & 10.43 & 0.134 & 76.03 & \underline{0.191} & 0.237 \\
        & \checkmark & \checkmark & {7.74} & \underline{0.119} & 83.03 & 0.192 & \underline{0.230} & \underline{9.87} & \underline{0.133} & {76.75} & 0.195 & \underline{0.236} \\
        \midrule
        MObI (512) & \checkmark & \checkmark & \textbf{6.60} & \textbf{0.115} & \textbf{84.22} & \textbf{0.129} & \textbf{0.148} & \textbf{9.00} & \textbf{0.129} & {76.75} & \textbf{0.132} & \textbf{0.153}  \\
        \bottomrule
    \end{tabular}
    \caption{Camera and lidar realism metrics for reinsertion and replacement tasks, with values averaged over \textit{tracked} and \textit{same reference} settings for reinsertion, and \textit{in-domain} and \textit{cross-domain reference} settings for replacement. We compare with camera-only methods and provide separate ablations on the use of the 3D bounding box and the gated cross-attention adapter.}
    \label{tab:full-realism}
\end{table*}

\begin{figure*}
        \begin{minipage}[b]{.65\linewidth}
        \scriptsize
        \raisebox{1.5cm}{
        \begin{tabular}{lcccccccc}
            & \multicolumn{2}{c}{\textbf{Scene-level}} & \multicolumn{6}{c}{\textbf{Restricted to reinserted objects}}\\
            \cmidrule(r){2-3} \cmidrule(r){4-9}
            & \multicolumn{2}{c}{\textbf{mAP}} & \multicolumn{2}{c}{\textbf{ATE}} & \multicolumn{2}{c}{\textbf{ASE}} & \multicolumn{2}{c}{\textbf{AOE}} \\
            \cmidrule(r){2-3} \cmidrule(r){4-5} \cmidrule(r){6-7} \cmidrule(r){8-9}
            & car & ped. & car & ped. & car & ped. & car & ped. \\
            \midrule
            Original & 0.885 & 0.873 & 0.145 & 0.103 & 0.138 & 0.278 & 0.024 & 0.462 \\
            Reinsertions & 0.878 & 0.863 & 0.299 & 0.140 & 0.145 & 0.303 & 0.161 & 0.754  \\
            \bottomrule
        \end{tabular}
        }
        \end{minipage}
        %
        %
        \begin{minipage}[b]{.3\linewidth}
            \includegraphics[width=\linewidth]{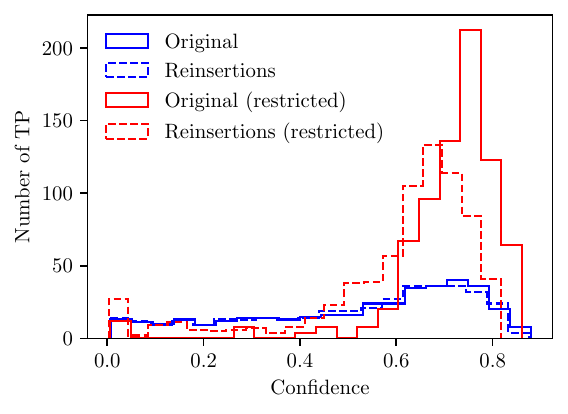}
            \vspace{-0.7cm}
        \end{minipage}
        \caption{Camera-lidar detection performance of an off-the-shelf BEVFusion~\cite{liu2023bevfusion} object detector on objects reinserted using our method. Left: we compute mAP at the scene-level, and TP errors (translation, scale, and orientation) on the reinserted objects only.
        Left: distribution of the scores of the true-positives show a modest shift towards lower scores for edited objects.
        }
        \label{fig:detection-results}
\end{figure*}

\subsection{Object detection on reinserted objects}
\label{sec:domain-gap}

\paragraph{Setup}
The inpainted objects must correspond tightly to the 3D box used during generation in order to be useful for various downstream tasks.
We analyse the quality of the 3D-box conditioning using an off-the-shelf object detector and compare detections to the boxes used for conditioning.
We use the nuScenes val split and select objects to reinsert based on the same filters as in \cref{sec:method:training details}.
If there are multiple such objects per frame, we pick one randomly and obtain 372 objects in total.
We follow the \emph{tracked reference} procedure described in \cref{sec:experiments:implementation} and replace each selected object using MObI, given a reference of the same object, taken at a random timestamp that is far from the inpainting timestamp.
We restrict the evaluation to those scenes that have been edited.
We use the multimodal BEVFusion~\cite{liu2023bevfusion} object detector with a SwinT~\cite{liu2021swin} backbone trained on nuScenes and do not accumulate lidar points over successive sweeps.

\paragraph{Metrics}
We compute mAP and error metrics on the re-inserted objects.
Scene-level metrics such as mAP cannot be easily restricted to edited objects\footnote{Because such metrics require false-positives which cannot be easily defined here.} and will not be very sensitive to detection errors on these objects.
We thus complement mAP with true-positive error metrics restricted to the re-inserted objects, computed following the usual matching procedure of the nuScenes devkit~\cite{caesar2020nuscenes} but considering only ground-truth/detection pairs for inpainted objects.


\paragraph{Results}
Camera-lidar object detection results are presented in \cref{fig:detection-results} (left), and we see that reinsertion comes at a small cost in object detection performance but that errors remain small (e.g. 0.161 AOE corresponds to a $9^{\circ}$ average error) while scene-level mAP is very similar.
We also show the distribution of the scores of the true-positives in \cref{fig:detection-results} (right), where the only scores suffer a modest decrease when the detector is applied to the reinserted samples.
Overall, this shows that while a small domain gap exists, our method leads to samples that are both realistic and geometrically accurate, and that an off-the-shelf detector can successfully detect such objects even though it has not been trained on any synthetic data generated by our method.
We show a sample of detections in \cref{fig:detection-comparison} where the reinserted object is detected accurately and the bounding boxes of the untouched objects remain almost identical.

\section{Strengths, limitations, and future work}

Our model generates objects that are coherent across viewpoints, as was illustrated in \cref{fig:controllability}.
This is surprising and indicates that references encoded using CLIP are highly informative about the appearance of the object and stable under changes in conditioning.
It would be interesting, however, to see if consistency for different viewpoints or time steps can be explicitly enforced.
This could be done by extending the cross-modal attention presented in \cref{sec:method:multimodal generation} to multiple time steps~\cite{gao2023magicdrive, li2023drivingdiffusion, drivescape, wen2023panacea, lu2025wovogen}, focusing on the same object.

Because we leverage a diffusion model pre-trained on webscale data, combined with the CLIP image encoding to guide the appearance of the inpainted object, our model can generate objects for classes unseen during training.
In \cref{fig:suppl:ood}, we replace objects from nuScenes~\cite{caesar2020nuscenes} classes beyond ``car'' and ``pedestrian'', producing plausible, yet seemingly lower-quality results compared to seen classes.

Inpainting completely out-of-domain references produces unpredictable results. Since the model is finetuned on a narrow domain, the model reverts to known classes when faced with unfamiliar references, such as turning a horse into a brown car in \cref{fig:suppl:ood}
Extending our method to a true open-world setting by training on multiple 3D object detection datasets, as in~\cite{minderer2022simple}, is an exciting future direction.

Similarly, inpainting can fail if the location of the edited object is in strong tension with the rest of the scene (e.g., placing a truck on the pavement), which can limit the applicability of our method for generating test cases that are deeply out-of-domain. Another limitation of our model comes from the fact that we condition only on a single bounding box, which means that the editing can sometimes modify background objects if they have a sizeable overlap with the edit mask.
This could be solved by using more precise segmentation masks, which are unfortunately not available for the nuScenes~\cite{caesar2020nuscenes} dataset.
An interesting approach, however, is to inpaint an object by additionally conditioning on all boxes in the scene, using a similar mechanism as in \cite{gao2023magicdrive}.
We leave full-scene conditioning for future work.

Despite these limitations, our approach offers an interesting avenue to edit multimodal scenes in a realistic and controllable manner.
We have shown that it is possible to insert new objects across modalities at a specific location and demonstrated the robustness and flexibility of our approach.
Such a capability could prove crucial to developing and testing safety-critical systems by providing a way to thoroughly explore the full range of possibilities that can occur in the real world using synthetic data.

\paragraph{Acknowledgements}
We thank Tom Joy for his early advisory support, which guided the initial direction of this work.
{
    \small
    \bibliographystyle{ieeenat_fullname}
    \bibliography{main}
}
\clearpage
\setcounter{page}{1}
\maketitlesupplementary

\setcounter{section}{0}
\setcounter{table}{0}
\setcounter{figure}{0}

\renewcommand{\thesection}{\Alph{section}}
\renewcommand{\thetable}{S\arabic{table}}
\renewcommand{\thefigure}{S\arabic{figure}}

\section{Extended Related Work}
\label{sec:supple:extended related work}

Multimodal data is crucial for ensuring safety in autonomous driving, and most state-of-the-art perception systems employ a sensor fusion approach, particularly for tasks like 3D object detection~\cite{liang2022bevfusionsimplerobustlidarcamera, liu2023bevfusion, gunn2024liftattendsplatbirdseyeviewcameralidarfusion}. However, testing and developing such safety-critical systems requires vast amounts of data, which is costly and time-consuming to obtain in the real world. Consequently, there is a growing need for simulated data, enabling models to be tested efficiently without requiring on-road vehicle testing.

\paragraph{Copy-and-paste} Early efforts in synthetic data generation relied on copy-and-paste methods. For example, \cite{georgakis2017synthesizing} used depth maps for accurate scaling and positioning when inserting objects, while later approaches like~\cite{dwibedi2017cut} focused on achieving patch-level realism through blending, improving 2D object detection. A more straightforward approach, presented by~\cite{ghiasi2021simple}, naively pastes objects into images without blending and demonstrates its efficacy in improving image segmentation.
In autonomous driving, PointAugmenting~\cite{wang2021pointaugmenting} extends this copy-and-paste approach to both camera and lidar data to enhance 3D object detection. Building on the lidar GT-Paste method~\cite{yan2018second}, it incorporates ideas from CutMix augmentation~\cite{yun2019cutmix} while ensuring multimodal consistency. This method addresses scale mismatches and occlusions by utilising the lidar point cloud for guidance during the insertion process. Similarly, MoCa~\cite{zhang2020exploring} employs a segmentation network to extract source objects before insertion, instead of directly pasting entire patches. Geometric consistency in monocular 3D object detection has also been explored in~\cite{lian2022exploring}. While these methods improve object detection and mitigate class imbalance, their compositing strategy leads to unrealistic blending, especially in image space. Furthermore, they lack controllability, such as the ability to adjust the position and orientation of inserted objects, limiting their utility for testing.

\paragraph{Image compositing} In this work, we aim to improve upon these approaches by drawing inspiration from recent advancements in image inpainting. Early efforts like ST-GAN~\cite{lin2018st} tackled the challenge of unrealistic foreground blending by using GANs~\cite{goodfellow2014generative} with spatial transformer networks to recursively predict and apply corrections, achieving natural blending via warp composition. ObjectStitch~\cite{song2023objectstitch} leverages diffusion within an edit mask for smooth patch-level blending. Methods like Paint-by-Example~\cite{yang2023paint} and AnyDoor~\cite{chen2023anydoor} extend this capability by generating entire images conditioned on scene context and an edit mask, achieving greater semantic coherence. AnyDoor achieves fine-grained object inpainting by using SAM~\cite{kirillov2023segment} 
for reference segmentation and more advanced feature extraction techniques. Other notable works include Magic Insert~\cite{ruiz2024magicinsertstyleawaredraganddrop}, which enables drag-and-drop object insertion between images with differing styles, and~\cite{kulal2023puttingpeopleplaceaffordanceaware}, which adjusts object pose to respect scene affordances. ObjectDrop~\cite{winter2024objectdropbootstrappingcounterfactualsphotorealistic} trains on counterfactual examples to enhance object insertion. Although these methods improve seamless and context-aware image compositing, they do not control the 3D position and orientation of objects in the real world, a critical requirement for training and testing, nor do they consider multimodal extensions.

\paragraph{Full scene generation} Recent advancements in conditional full-scene generation have yielded impressive results. BEVControl~\cite{yang2023bevcontrol} uses a two-stage method (controller and coordinator) to generate scenes conditioned on sketches, ensuring accurate foreground and background content. Text2Street~\cite{su2024text2street} combines bounding box encoding with text conditions, employing a ControlNet-like~\cite{zhang2023controlnet} architecture for guidance. DrivingDiffusion~\cite{li2023drivingdiffusion} represents bounding boxes as layout images passed as an extra channel in the U-Net~\cite{ronneberger2015u}. MagicDrive~\cite{gao2023magicdrive} incorporates bounding boxes and camera parameters alongside text conditions for full-scene generation, with a cross-view attention module leveraging BEV layouts. SubjectDrive~\cite{huang2024subjectdrivescalinggenerativedata} generates camera videos conditioned on the appearance of foreground objects. LiDM~\cite{lidardiffusion} focuses on lidar scene generation conditioned on semantic maps, text, and bounding boxes. DriveScape~\cite{drivescape} introduces a method to generate multi-view camera videos conditioned on 3D bounding boxes and maps using a bi-directional modulated transformer for spatial and temporal consistency.

Synthetic lidar data generation has also advanced significantly. LidarGen~\cite{lidargen} and LiDM~\cite{lidardiffusion} employ diffusion for lidar generation, with the latter also incorporating semantic maps, bounding boxes, and text. UltraLidar~\cite{xiong2023ultralidarlearningcompactrepresentations} densifies sparse lidar point clouds, while RangeLDM~\cite{hu2024rangeldmfastrealisticlidar} accelerates lidar data generation by converting point clouds into range images using Hough sampling and enhancing reconstruction through a range-guided discriminator. DynamicCity~\cite{bian2024dynamiccitylargescalelidargeneration} generates lidar occupancy grid sequences conditioned on dynamic scene layouts. However, full-scene generation can result in a large domain gap, particularly for downstream tasks like object detection, making it challenging to create realistic counterfactuals.
\cite{XIANG2024105207, kirby2024logen} focus on object-level lidar generation, with LOGen~\cite{kirby2024logen} using a point-based conditional diffusion framework, enabling fine-grained control over object geometry, viewpoint, and distance.
None of the works mentioned above jointly generate camera and lidar data.

\paragraph{Multimodal object inpainting} GenMM~\cite{singh2024genmm} represents a new direction in multimodal object inpainting using a multi-stage pipeline that ensures temporal consistency. However, it remains limited in controllability, requiring the reference to closely align with the insertion angle. Furthermore, it does not generate lidar and camera modalities jointly, instead focusing on geometric alignment while excluding lidar intensity values. We take a similar approach, but propose an end-to-end method that jointly generates camera and lidar data for reference-guided multimodal object inpainting. Our method achieves realistic and consistent multimodal outputs across diverse object angles.

\section{Method Details}
\label{sec:suppl:method}

\subsection{Details on image processing}

\paragraph{Bounding Box Projection:} The bounding boxes from the source and destination scenes, $ \text{box}_s, \text{box}_d \in \mathbb{R}^{8 \times 3}$, are projected onto the image space using the respective camera transformations:
\[
\text{box}_s^{\text{(C)}} = \mathbf{T}_s^{\text{(C)}} \cdot \text{box}_s \in \mathbb{R}^{8 \times 2}, \quad 
\text{box}_d^{\text{(C)}} = \mathbf{T}_d^{\text{(C)}} \cdot \text{box}_d \in \mathbb{R}^{8 \times 2}.
\]
We randomly crop the source image around the corresponding bounding box in such a way that the projected bounding box covers at least 20\% of the area. We apply the corresponding viewport transformation to $ \text{box}_d $.

\paragraph{Edit Mask:} The edit region is defined by a binary mask $ \mathbf{m}^{\text{(C)}} \in \{0, 1\}^{D \times D} $, created by inpainting $ \text{box}_d^{\text{(C)}} $ onto an initially all-zero matrix, where the inpainted region is assigned values of 1. The complement of this mask is defined as:
\[
\mathbf{\bar{m}}^{\text{(C)}} = \mathbf{J} - \mathbf{m}^{\text{(C)}}, \quad \mathbf{J} \in \{1\}^{D \times D}.
\]

\subsection{Details on lidar processing and encoding}
\label{sec:suppl:method:lidar processing}

We consider the lidar point cloud of the destination scene, $ P_d \in \mathbb{R}^{N \times 4} $, where $N$ represents the number of points and the four channels correspond to the $x, y, z$ coordinates and intensity values.
The lidar points are projected onto a range view $ R_d \in \mathbb{R}^{32 \times 1096 \times 2} $ using the transformation described below.
This transformation is loss-less, except for points near the end of the lidar sweep that overlap with the beginning due to motion compensation.

\paragraph{Point cloud to range view transformation}

We consider the point cloud for a single sweep of the destination scene, \( P_d \in \mathbb{R}^{N \times 4} \), where \( N \) represents the number of points, and the four channels correspond to the \( x, y, z \) coordinates and intensity values. The lidar points are projected onto a range view \( R_d \in \mathbb{R}^{32 \times 1096 \times 2} \) using the transformation described below.

For each point in \( P_d \), the depth (Euclidean distance from the sensor) is calculated as:
\[
d_i = \sqrt{x_i^2 + y_i^2 + z_i^2}.
\]
Points with depths outside the predefined range \( [1.4, 54] \) are filtered out. The yaw and pitch angles are then computed as:
\[
\text{yaw}_i = -\arctan2(y_i, x_i), \quad \text{pitch}_i = \arcsin\left(\frac{z_i}{d_i}\right).
\]

The beam pitch angles \( \{\theta_k\}_{k=1}^{H} \) are chosen as \( \theta_k = 0.0232 \cdot x_k \), where \( x_k \in \{-23, -22, \ldots, 8\} \), to best match the binning of the nuScenes~\cite{caesar2020nuscenes} lidar sensor's vertical beams and its field of view. Each point is assigned to the closest vertical beam based on its pitch angle, determining its $y_i$ vertical coordinate, an integer in the range $ [0, 31] $. 

The yaw angle is mapped to the horizontal coordinate \( x \) of the range view grid as:
\[
x_i = \left\lfloor \frac{\text{yaw}_i}{\pi} \cdot \frac{W}{2} + \frac{W}{2} \right\rfloor,
\]

The final range view representation \( R_d \) of the destination scene encodes depth and intensity for each point projected onto the \( H \times W \) grid, where \( H = 32 \) denotes the number of vertical beams, and \( W = 1096 \) represents the horizontal resolution. Unassigned pixels in the range view are set to a default value. Each point is mapped to a specific pixel coordinate in the range view. 

Note that the transformation is not injective, as some points overlap at the start and end of the lidar sweep due to motion compensation; however, this overlap has minimal impact. We additionally store the original pitch and yaw values for each point assigned to a range view pixel in matrices \( R_d^{\text{yaw}} \in \mathbb{R}^{H \times W} \) and \( R_d^{\text{pitch}} \in \mathbb{R}^{H \times W} \), respectively. These matrices enhance the inverse transformation from range view to point cloud by preserving the unrasterized angular information.

\paragraph{Range view to point cloud Transformation}

To reconstruct the point cloud from the range view, we leverage the stored unrasterized pitch and yaw matrices, \( R_d^{\text{pitch}} \in \mathbb{R}^{H \times W} \) and \( R_d^{\text{yaw}} \in \mathbb{R}^{H \times W} \), which preserve the original angular information for each pixel.

The depth values \( R_d^{\text{depth}} \in \mathbb{R}^{H \times W} \) are flattened to the vector \( \mathbf{d} \in \mathbb{R}^N \), where \( N = H \times W \). Similarly, the pitch and yaw matrices are flattened to the vectors \( \boldsymbol{\theta} \in \mathbb{R}^N \) and \( \boldsymbol{\phi} \in \mathbb{R}^N \), representing the pitch and yaw angles for each pixel in the range view. Using these angular and depth values, the point cloud \( P_d \in \mathbb{R}^{N \times 3} \) is reconstructed as:
\begin{align*}
\mathbf{p}_x &= \mathbf{d} \cdot \cos(\boldsymbol{\phi}) \cdot \cos(\boldsymbol{\theta}) \\
\mathbf{p}_y &= -\mathbf{d} \cdot \sin(\boldsymbol{\phi}) \cdot \cos(\boldsymbol{\theta}) \\
\mathbf{p}_z &= \mathbf{d} \cdot \sin(\boldsymbol{\theta}),
\end{align*}
where \( \mathbf{p}_x, \mathbf{p}_y, \mathbf{p}_z \in \mathbb{R}^N \) are the vectors of reconstructed \( x \), \( y \), and \( z \) coordinates, respectively. The reconstructed point cloud \( P_d \) is then given by stacking these coordinate vectors as \( P_d = [\mathbf{p}_x, \mathbf{p}_y, \mathbf{p}_z] \).

By leveraging the stored pitch and yaw matrices, the process accurately restores the point cloud while avoiding misalignments introduced by motion compensation. This ensures that the reconstructed point cloud aligns perfectly with the original input, except for the overlapping points we previously mentioned, which do not get regenerated.

\paragraph{Range view to range image processing}
We project the bounding box $ \text{box}_d $ onto $ R_d $ using the coordinate-to-range transformation, resulting in $ \text{box}_d^{\text{(R)}} \in \mathbb{R}^{8 \times 3} $, while preserving the depth of each bounding box point.
To enhance the region of interest, we employ a zoom-in strategy analogous to that used in the image processing, by cropping the range view width-wise around $ \text{box}_d^{\text{(R)}} $, resulting in a $ {32 \times W^{\text{(R)}} \times 2} $ object-centric range view, and resizing it to obtain the range image $ \mathbf{x}^{\text{(R)}} \in \mathbb{R}^{D \times D \times 2} $. We apply the same viewport transformation to the bounding box $\text{box}_d^{\text{(R)}}$.
The edit region is defined by a mask $ \mathbf{m}^{\text{(R)}} \in \{0, 1\}^{D \times D} $, which is created by inpainting the bounding box $\text{box}_d$ onto an initially all-zero matrix, where the inpainted region has values of 1. The complement of this mask is $\mathbf{\bar{m}}^{\text{(R)}} = \left(\mathbf{J} - \mathbf{m}^{\text{(R)}}\right)$.

\begin{figure*}
    \centering
    \footnotesize
    \includegraphics[width=0.98\linewidth]{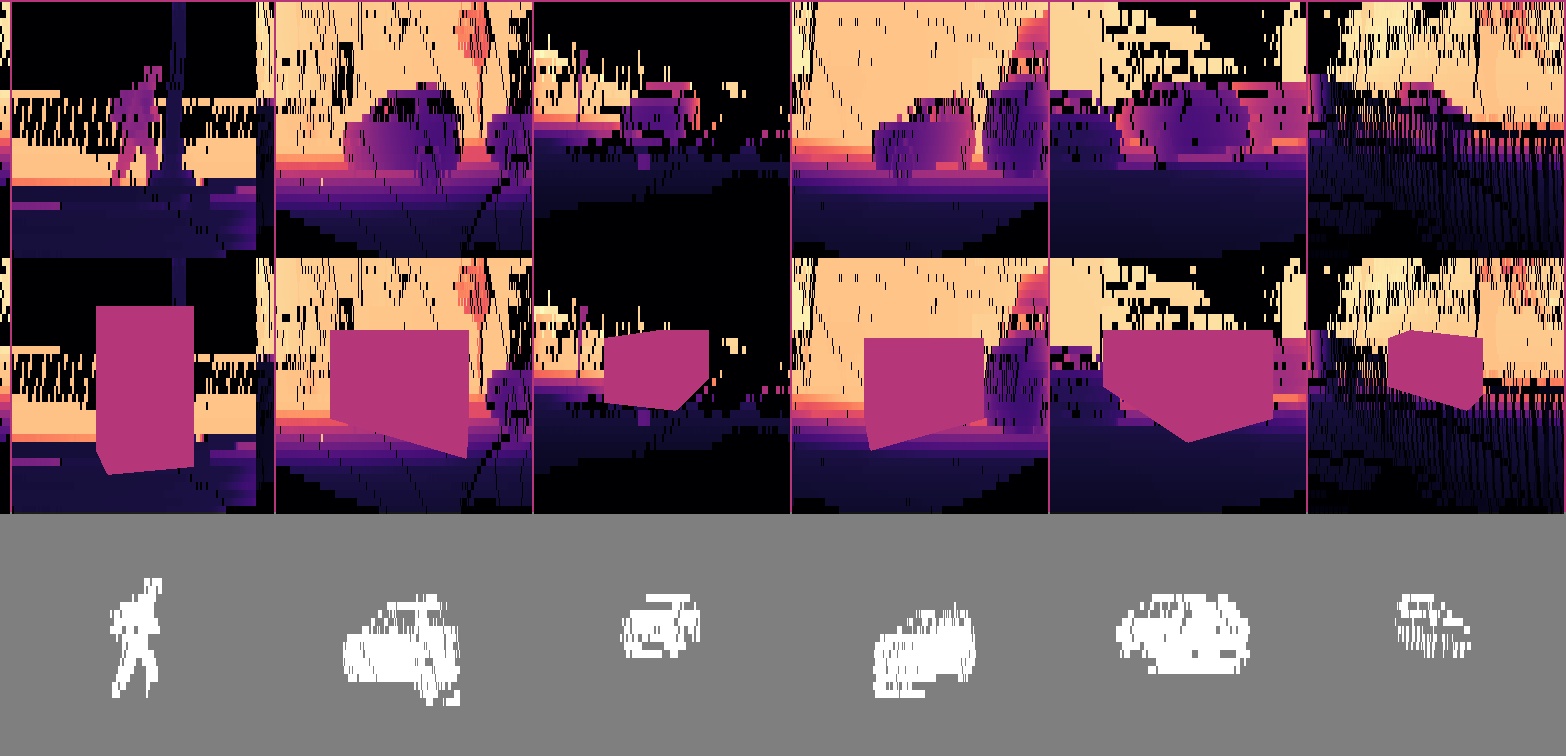}
    \rotatebox{90}{\quad~~ \textbf{Object pixels} \qquad \qquad \quad \textbf{Edit mask} \qquad \qquad~ \textbf{Range image}}
    \caption{From top to bottom: (i) object-centric range depth image, (ii) range depth context with an edit mask, generated by projecting the object bounding box onto the range view, and (iii) object mask highlighting pixels corresponding to points within the 3D bounding box.}
    \label{fig:suppl:range_masks}
\end{figure*}

\paragraph{Range image reconstruction metrics}
An important step towards achieving realistic lidar inpainting is ensuring the autoencoder can reconstruct the input point cloud with high fidelity. Since the point cloud to range view transformation is loss-less, we can focus our attention on evaluating the quality of reconstructed range views.
We restrict our evaluation to the region within the edit mask $ \mathbf{m}^{\text{(R)}} $ and the object points from the target range view, selected using the 3D bounding box, see \cref{fig:suppl:range_masks} for examples.
For each input range view $ \mathbf{X}^{\text{(R)}} $ 
and its reconstruction, $ \mathcal{D}^{\text{(R)}}(\mathcal{E}^{\text{(R)}}(\mathbf{X}^{\text{(R)}})) $, we compute the median depth error and the mean squared error (MSE) of the intensity values, restricted on the object points and the edit mask.

\begin{table*}
    \centering
    \scriptsize
    \begin{tabular}{lcccccccccccc}
        & \multicolumn{6}{c}{\textbf{Reinsertion}} & \multicolumn{6}{c}{\textbf{Replacement}}\\
        \cmidrule(r){2-7} \cmidrule(r){8-13}
        & \multicolumn{3}{c}{\textbf{same ref}} & \multicolumn{3}{c}{\textbf{tracked ref}} & \multicolumn{3}{c}{\textbf{in-domain ref}} & \multicolumn{3}{c}{\textbf{cross-domain ref}} \\
        \cmidrule(r){2-4} \cmidrule(r){5-7} \cmidrule(r){8-10} \cmidrule(r){11-13}
        \textbf{Method} & FID\textdownarrow & LPIPS\textdownarrow & CLIP-I\textuparrow & FID\textdownarrow & LPIPS\textdownarrow & CLIP-I\textuparrow & FID\textdownarrow & LPIPS\textdownarrow & CLIP-I\textuparrow & FID\textdownarrow & LPIPS\textdownarrow & CLIP-I\textuparrow \\
        \midrule
        copy\&paste & \multicolumn{6}{c}{n/a} & 13.50 & 0.196 & n/a & 17.08 & 0.213 & n/a \\
        PbE~\cite{yang2023paint} & 7.34 & 0.131 & 84.50 & 7.58 & 0.135 & 83.31 & 9.62 & 0.148 & 77.44 & 10.54 & 0.150 & \textbf{77.06} \\
        MObI (ours) & \textbf{6.50} & \textbf{0.114} & \textbf{84.94} & \textbf{6.70} & \textbf{0.115} & \textbf{83.50} & \textbf{8.95} & \textbf{0.127} & \textbf{77.50} & \textbf{9.05} & \textbf{0.130} & 76.00 \\
        \bottomrule
        \\
    \end{tabular}
    \caption{Comparison with image inpainting methods at $D=512$ resolution in terms of camera realism.}
    \label{tab:supp:camera-realism-comparison}
\end{table*}

\paragraph{Range image encoding}
We adapt the pre-trained image VAE~\cite{kingma2013auto} of StableDiffusion~\cite{rombach2022high} to the lidar modality through a series of training-free adaptations and a fine-tuning step, ablated in \cref{tab:lidar reconstruction}.

As a naive solution to encode the lidar modality, we take the preprocessed range view $\mathbf{x}^{\text{(R)}} \in \mathbb{R}^{D \times D \times 2}$, duplicate the depth channel, and pass the resulting 3-channel representation through the image VAE~\cite{kingma2013auto}. After discarding one depth channel and resizing back to $32 \times W^{\text{(R)}} \times 2$ using nearest neighbour interpolation, we compute reconstruction errors using the metrics described in \cref{sec:suppl:method:lidar processing}. This naive approach results in unsatisfactory reconstruction errors.

To address this, we propose three cumulative adaptations that improve depth and intensity reconstruction for object points and the extended edit mask. First, we leverage the higher resolution of $\mathbf{x}^{\text{(R)}}$ by applying average pooling when downsizing, which serves as an error correction mechanism.

Next, we observe that the reconstruction error of range pixel values is proportional to the interval size of their distribution. Since intensity values follow an exponential distribution, we normalize intensity $i \in [0, 255]$ using the cumulative distribution function (CDF) of the exponential distribution, choosing $\lambda = 4$ experimentally:

\[
i' = 2e^{-\lambda\frac{i}{255}} - 1 \in [-1, 1]
\]

To enhance object-level depth reconstruction, we apply depth normalization based on the minimum and maximum depth of $\text{box}_d^{\text{(R)}}$, scaling the bounding box by 0.1, which extends the interval the object depth values are distributed on and, in turn, improves object reconstruction error:

\[
d' = 
\begin{cases} 
    -\alpha + 2\alpha \cdot \frac{d - \text{min}_d}{\text{max}_d - \text{min}_d} & \text{if } \text{min}_d \leq d \leq \text{max}_d \\
    -1 + (-(\alpha - 1)) \cdot \frac{d + 1}{\text{min}_d + 1} & \text{if } -1 \leq d < \text{min}_d \\
    \alpha + (1 - \alpha) \cdot \frac{d - \text{max}_d}{1 - \text{max}_d} & \text{if } \text{max}_d < d \leq 1 
\end{cases}
\]

where $d$ is the depth value, $\alpha$ controls range scaling, and $\text{min}_d$, $\text{max}_d$ define normalization boundaries within $[-1, 1]$. Depth values are originally between $[1.4, 54]$, but are linearly normalized to $[-1, 1]$.

Thirdly, we replace the input and output convolution of the pre-trained image encoder and decoder, with two residual blocks, respectively. We now have 2 input and output channels. We fine-tune the VAE~\cite{kingma2013auto} with an additional discriminant~\cite{esser2021taming}. The same normalization and resizing strategies are applied, yielding the best reconstruction metrics for $\mathbf{\tilde{x}}^{(R)} = \text{resize}(\mathcal{D}^{\text{(R)}}(\mathcal{E}^{\text{(R)}}(\text{norm}({\mathbf{x}^{\text{(R)}}}))))$.

Finally we encode the range image $ \mathbf{x}^{\text{(R)}} $ to obtain a latent representation $ \mathbf{z}_0^{\text{(R)}} = \mathcal{E}^{\text{(R)}}(\text{norm}({\mathbf{x}^{\text{(R)}}})) $.
Similarly, we encode the lidar environment context $ \mathbf{x}^{\text{(R)}} \odot \mathbf{\bar{m}}^{\text{(R)}} $ to obtain a latent conditioning representation $ \mathbf{c}^{\text{(R)}}_{\text{env}} = \mathcal{E}^{\text{(R)}}(\text{norm}(\mathbf{x}^{\text{(R)}} \odot \mathbf{\bar{m}}^{\text{(R)}})) $.

\subsection{Additional training details}
\label{sec:suppl:method:training details}

We start by training the newly added input and output adapters of the range autoencoder while keeping the rest of the image VAE~\cite{kingma2013auto} from Stable Diffusion~\cite{rombach2022high} frozen. This training phase spans 8 epochs (15k steps) with a learning rate of \(4.5 \times 10^{-5}\), selecting the checkpoint with the lowest reconstruction loss.

During fine-tuning of the diffusion model, the autoencoders and all layers from the PbE~\cite{yang2023paint} framework remain frozen. Only the bounding box encoder, bounding box adaptation layer, and cross-modal attention layers are trained over 30 epochs (approximately 90k steps) with a constant learning rate of \(8 \times 10^{-5}\) and a batch size of 2 multimodal samples.

 Training takes approximately 20 hours on 8x 24GB NVIDIA A10G or 2x 80GB NVIDIA A100 GPUs. Inference throughput is about 8 camera+lidar samples per minute on a single A100.

\begin{figure*}[t]
    \centering
    \begin{minipage}{0.48\textwidth}
        \centering
        \includegraphics[width=\textwidth]{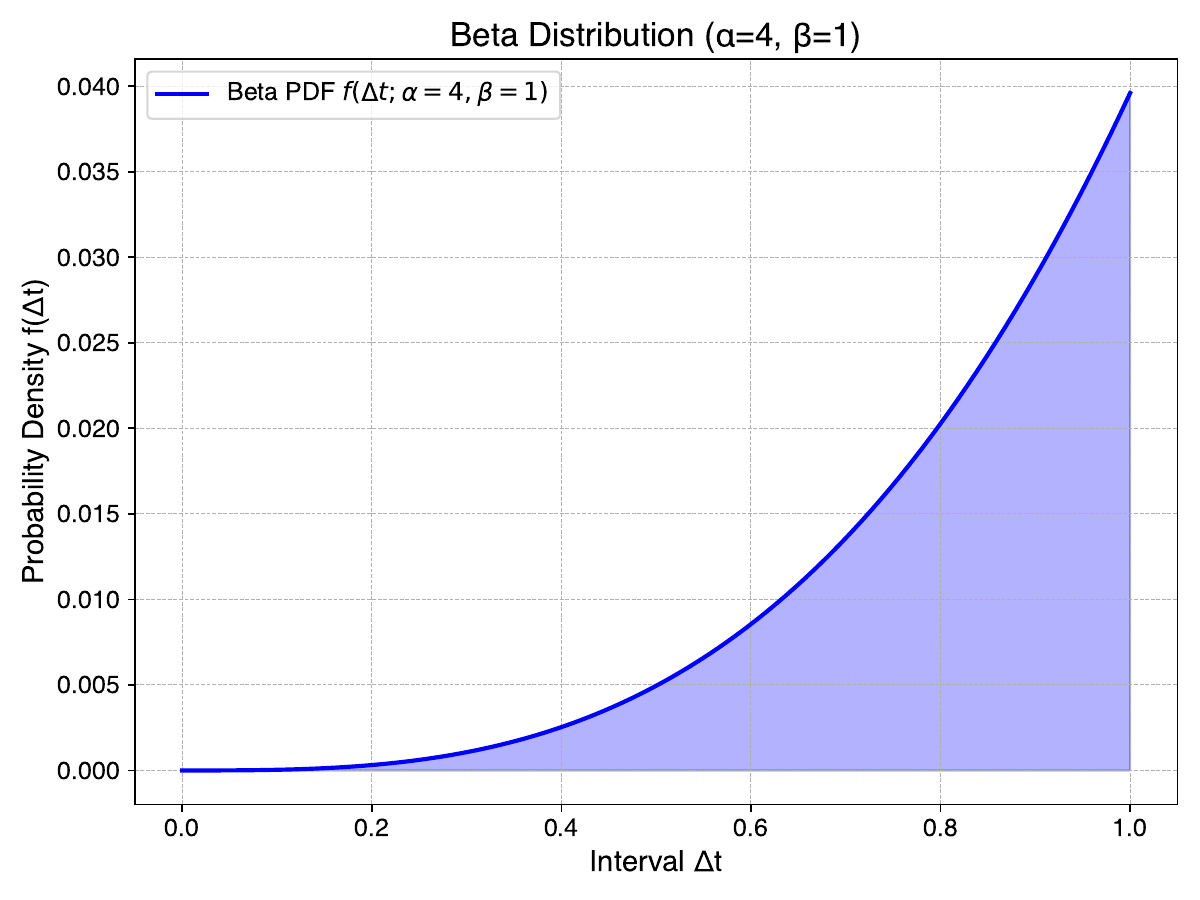}
        \caption{The probability density function of the Beta distribution with parameters \(\alpha=4\) and \(\beta=1\), used to sample reference patches of an object based on the normalized timestamp difference \(\Delta t\) between tracked instances. Patches from further time points are sampled with higher frequency.}
        \label{fig:supply:beta_distribution}
    \end{minipage}
    \hfill
    \begin{minipage}{0.48\textwidth}
        \centering
        \setlength{\tabcolsep}{1pt} 
        \footnotesize
        \begin{tabular}{ccc}
            \textbf{Training input (C)} & \textbf{Empty projected box} & \textbf{Training output (C)} \\
            \includegraphics[width=0.3\columnwidth]{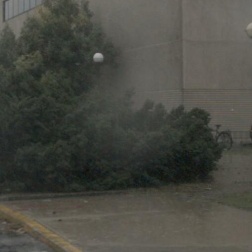} &
            \includegraphics[width=0.3\columnwidth]{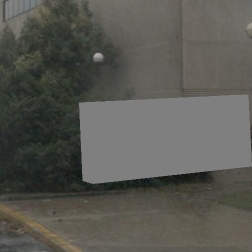} &
            \includegraphics[width=0.3\columnwidth]{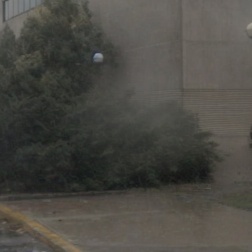} \\
            
            \includegraphics[width=0.3\columnwidth]{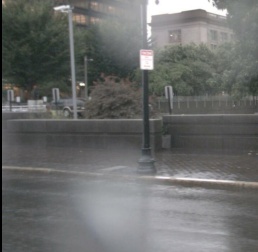} &
            \includegraphics[width=0.3\columnwidth]{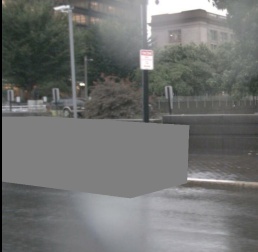} &
            \includegraphics[width=0.3\columnwidth]{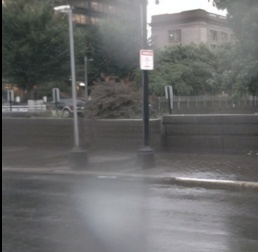} \\
            
            \includegraphics[width=0.3\columnwidth]{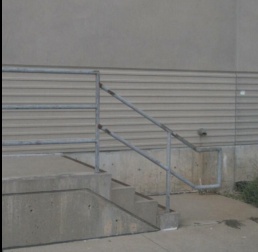} &
            \includegraphics[width=0.3\columnwidth]{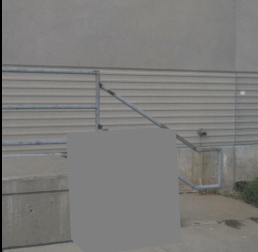} &
            \includegraphics[width=0.3\columnwidth]{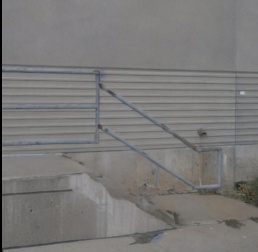} \\
            
            \includegraphics[width=0.3\columnwidth]{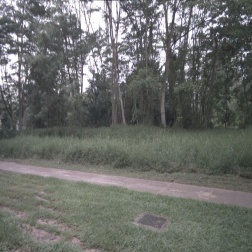} &
            \includegraphics[width=0.3\columnwidth]{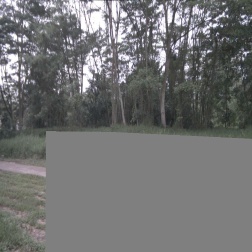} &
            \includegraphics[width=0.3\columnwidth]{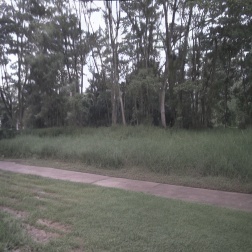} \\
        \end{tabular}
        \caption{Empty boxes are sampled during training for data augmentation, with the reference conditioning set to a black image and the bounding box coordinates set to zero.}
        \label{fig:suppl:erase_training_examples}
    \end{minipage}
\end{figure*}

\paragraph{Sampling Empty Boxes for Augmentation} To enhance augmentation, we sample empty bounding boxes to train the model to reconstruct missing details. A dedicated database of 10,000 such boxes is created. For a given scene, an object from a different scene is selected, ensuring that teleporting the bounding box into the current scene does not result in 3D overlap or a total 2D IoU overlap exceeding 50\% with other objects. During training, 30\% of the samples are drawn from this database. Black images and boxes with zero coordinates are used for these samples, enabling the model to learn how to fill in background details, as shown in \cref{fig:suppl:erase_training_examples}.

\paragraph{Tracked reference sampling} Rather than reinserting objects into the scene using the same reference, we utilize the temporal structure of the nuScenes dataset~\cite{caesar2020nuscenes}. References for the current object are sampled from a different timestamp following the distribution shown in \cref{fig:supply:beta_distribution}.

\begin{figure*}[ht]
    \centering
    \small
    \renewcommand{\arraystretch}{0.3} 
    \begin{tabular}{c@{\hspace{1pt}}c@{\hspace{8pt}}c@{\hspace{1pt}}c@{\hspace{1pt}}c@{\hspace{1pt}}c@{\hspace{1pt}}c@{\hspace{1pt}}c@{\hspace{4pt}}l}
        {\textbf{Ref. image}} &
        {\textbf{Original scene}} &
        \multicolumn{6}{c}{\textbf{Edited scenes}} &
        \\[4pt]
        
        \includegraphics[width=0.12\linewidth]{images/controllability/rot_60_car_red_suv/ref_rgb.jpg} &
        \includegraphics[width=0.12\linewidth]{images/controllability/rot_60_car_red_suv/orig_rgb} &
        \includegraphics[width=0.12\linewidth]{images/controllability/rot_60_car_red_suv/pred_rgb_0} &
        \includegraphics[width=0.12\linewidth]{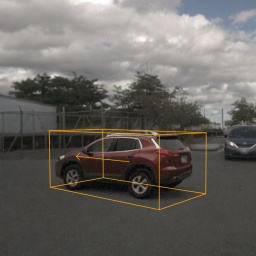} &
        \includegraphics[width=0.12\linewidth]{images/controllability/rot_60_car_red_suv/pred_rgb_2.jpg} &
        \includegraphics[width=0.12\linewidth]{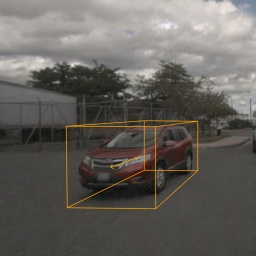} &
        \includegraphics[width=0.12\linewidth]{images/controllability/rot_60_car_red_suv/pred_rgb_4.jpg} &
        \includegraphics[width=0.12\linewidth]{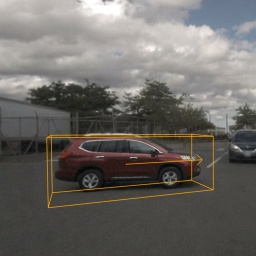} &
        \rotatebox{90}{\quad~ Camera} \\[1pt]
        
        &
        \includegraphics[width=0.12\linewidth]{images/controllability/rot_60_car_red_suv/orig_depth} &
        \includegraphics[width=0.12\linewidth]{images/controllability/rot_60_car_red_suv/pred_depth_0.jpg} &
        \includegraphics[width=0.12\linewidth]{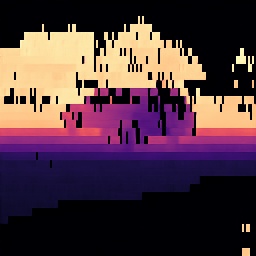} &
        \includegraphics[width=0.12\linewidth]{images/controllability/rot_60_car_red_suv/pred_depth_2.jpg} &
        \includegraphics[width=0.12\linewidth]{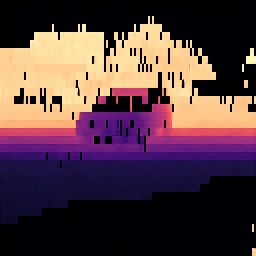} &
        \includegraphics[width=0.12\linewidth]{images/controllability/rot_60_car_red_suv/pred_depth_4.jpg} &
        \includegraphics[width=0.12\linewidth]{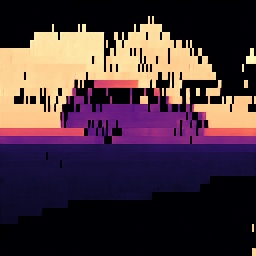} &
        \rotatebox{90}{~~ LiDAR depth} \\[1pt]
        
        &
        \includegraphics[width=0.12\linewidth]{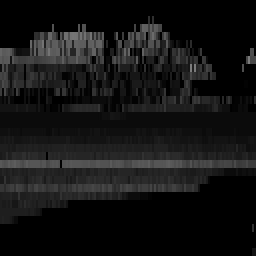} &
        \includegraphics[width=0.12\linewidth]{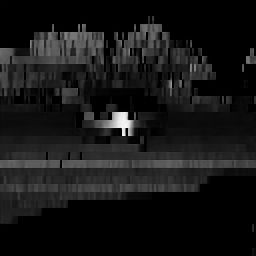} &
        \includegraphics[width=0.12\linewidth]{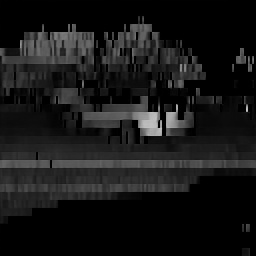} &
        \includegraphics[width=0.12\linewidth]{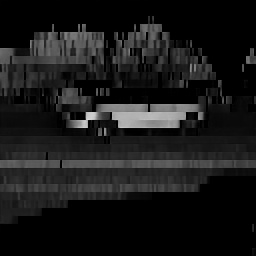} &
        \includegraphics[width=0.12\linewidth]{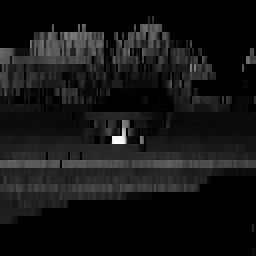} &
        \includegraphics[width=0.12\linewidth]{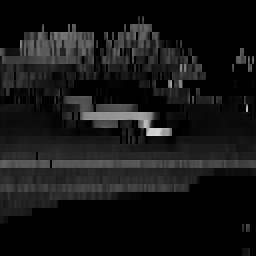} &
        \includegraphics[width=0.12\linewidth]{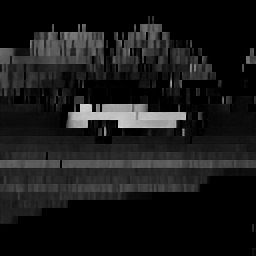} &
        \rotatebox{90}{LiDAR intensity}\vspace{8pt}
    \end{tabular}
    
    \begin{tabular}{c@{\hspace{1pt}}c@{\hspace{8pt}}c@{\hspace{1pt}}c@{\hspace{1pt}}c@{\hspace{1pt}}c@{\hspace{1pt}}c@{\hspace{1pt}}c@{\hspace{4pt}}l}
        
        \includegraphics[width=0.12\linewidth]{images/controllability/move_ped_white_skirt/ref_rgb.jpg} &
        \includegraphics[width=0.12\linewidth]{images/controllability/move_ped_white_skirt/orig_rgb} &
        \includegraphics[width=0.12\linewidth]{images/controllability/move_ped_white_skirt/pred_rgb_0} &
        \includegraphics[width=0.12\linewidth]{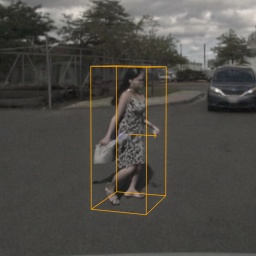} &
        \includegraphics[width=0.12\linewidth]{images/controllability/move_ped_white_skirt/pred_rgb_2.jpg} &
        \includegraphics[width=0.12\linewidth]{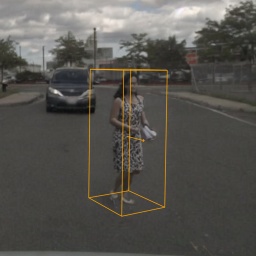} &
        \includegraphics[width=0.12\linewidth]{images/controllability/move_ped_white_skirt/pred_rgb_4.jpg} &
        \includegraphics[width=0.12\linewidth]{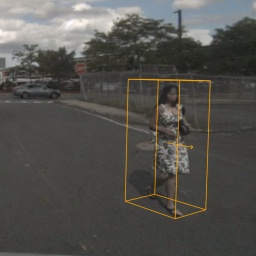} &
        \rotatebox{90}{\quad~ Camera} \\[1pt]
        
        &
        \includegraphics[width=0.12\linewidth]{images/controllability/move_ped_white_skirt/orig_depth} &
        \includegraphics[width=0.12\linewidth]{images/controllability/move_ped_white_skirt/pred_depth_0.jpg} &
        \includegraphics[width=0.12\linewidth]{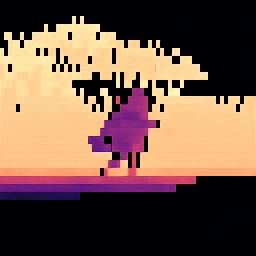} &
        \includegraphics[width=0.12\linewidth]{images/controllability/move_ped_white_skirt/pred_depth_2.jpg} &
        \includegraphics[width=0.12\linewidth]{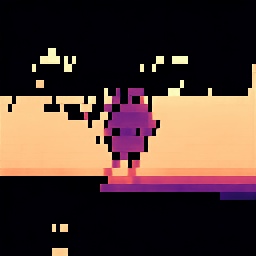} &
        \includegraphics[width=0.12\linewidth]{images/controllability/move_ped_white_skirt/pred_depth_4.jpg} &
        \includegraphics[width=0.12\linewidth]{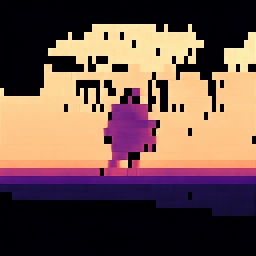} &
        \rotatebox{90}{~~ LiDAR depth} \\[1pt]
        
        &
        \includegraphics[width=0.12\linewidth]{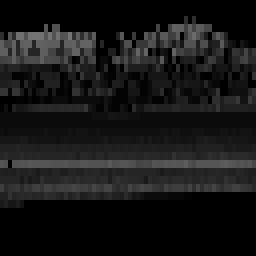} &
        \includegraphics[width=0.12\linewidth]{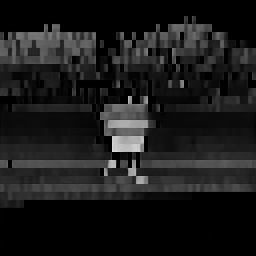} &
        \includegraphics[width=0.12\linewidth]{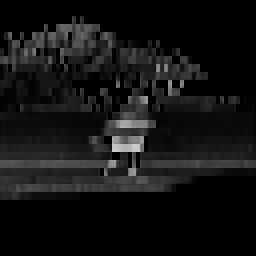} &
        \includegraphics[width=0.12\linewidth]{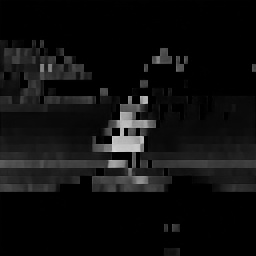} &
        \includegraphics[width=0.12\linewidth]{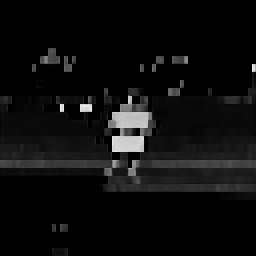} &
        \includegraphics[width=0.12\linewidth]{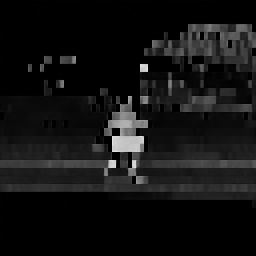} &
        \includegraphics[width=0.12\linewidth]{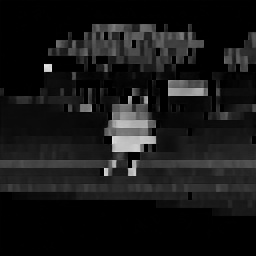} &
        \rotatebox{90}{LiDAR intensity} \vspace{8pt}
    \end{tabular}
    \caption{Additional examples showcasing our method's controllability. From left to right: reference image $ \mathbf{x}_{\text{ref}} $ extracted from a seperate source scene, original destination scene (original RGB image $ \mathbf{x}^{\text{(C)}} $, LiDAR range depth $ \mathbf{x}_0^{\text{(R)}} $ and intensity $ \mathbf{x}_1^{\text{(R)}} $), and edited scenes.}
    \label{fig:suppl:rotation_results}
  \end{figure*}

\begin{figure*}[ht]
    \centering
    \small
    \renewcommand{\arraystretch}{0.3} 
    \begin{tabular}{c@{\hspace{1pt}}c@{\hspace{8pt}}c@{\hspace{1pt}}c@{\hspace{1pt}}c@{\hspace{1pt}}c@{\hspace{1pt}}c@{\hspace{1pt}}c@{\hspace{4pt}}l}
        {\textbf{Ref. image}} &
        {\textbf{Original scene}} &
        \multicolumn{6}{c}{\textbf{Edited scenes}} &
        \\[4pt]
        
        \includegraphics[width=0.12\linewidth]{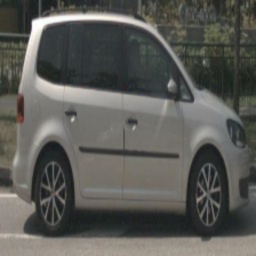} &
        \includegraphics[width=0.12\linewidth]{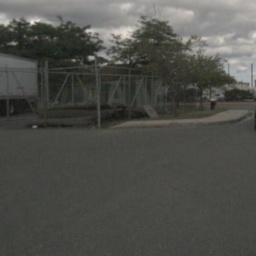} &
        \includegraphics[width=0.12\linewidth]{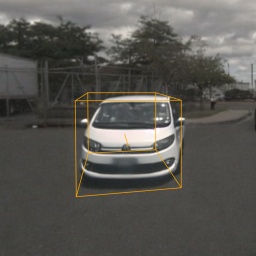} &
        \includegraphics[width=0.12\linewidth]{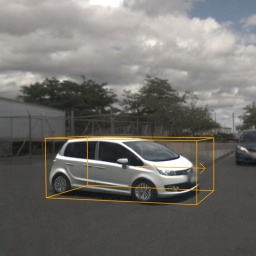} &
        \includegraphics[width=0.12\linewidth]{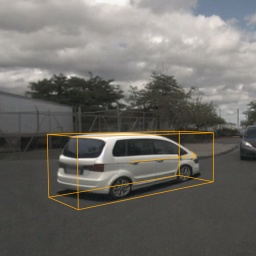} &
        \includegraphics[width=0.12\linewidth]{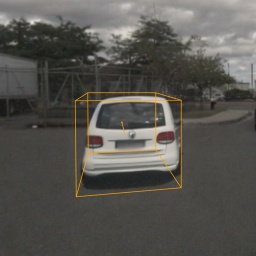} &
        \includegraphics[width=0.12\linewidth]{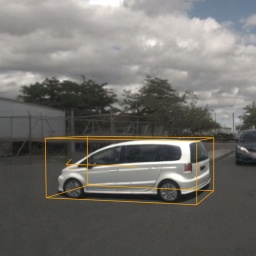} &
        \includegraphics[width=0.12\linewidth]{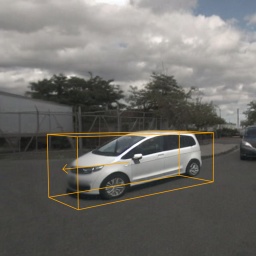} &
        \rotatebox{90}{\quad~ Camera} \\[1pt]
        
        &
        \includegraphics[width=0.12\linewidth]{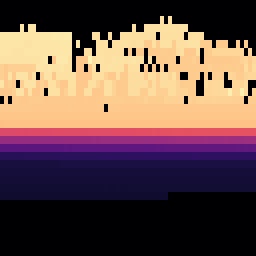} &
        \includegraphics[width=0.12\linewidth]{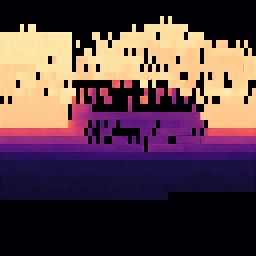} &
        \includegraphics[width=0.12\linewidth]{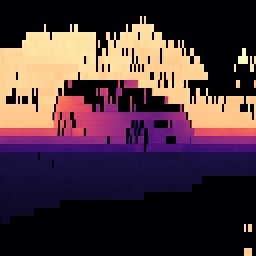} &
        \includegraphics[width=0.12\linewidth]{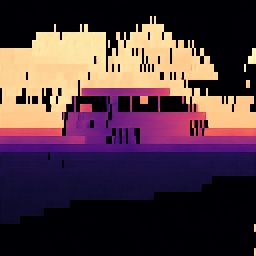} &
        \includegraphics[width=0.12\linewidth]{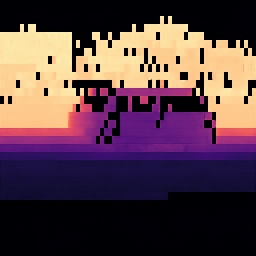} &
        \includegraphics[width=0.12\linewidth]{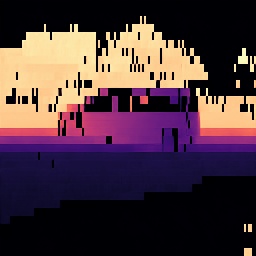} &
        \includegraphics[width=0.12\linewidth]{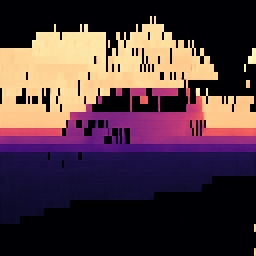} &
        \rotatebox{90}{~~ LiDAR depth} \\[1pt]
        
        &
        \includegraphics[width=0.12\linewidth]{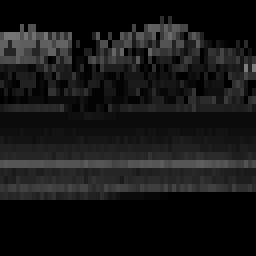} &
        \includegraphics[width=0.12\linewidth]{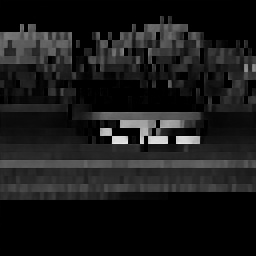} &
        \includegraphics[width=0.12\linewidth]{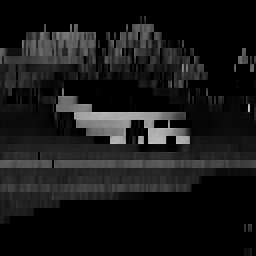} &
        \includegraphics[width=0.12\linewidth]{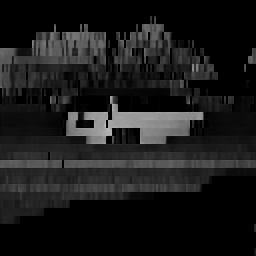} &
        \includegraphics[width=0.12\linewidth]{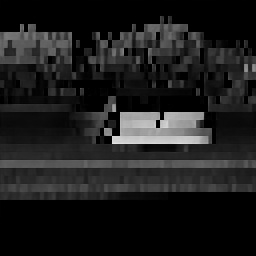} &
        \includegraphics[width=0.12\linewidth]{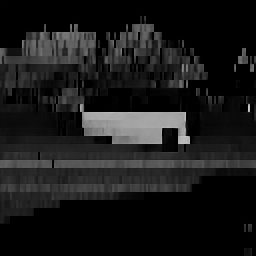} &
        \includegraphics[width=0.12\linewidth]{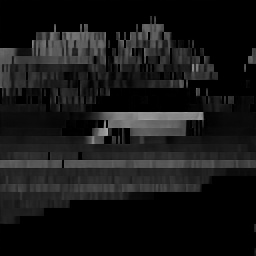} &
        \rotatebox{90}{LiDAR intensity}\vspace{8pt}
    \end{tabular}
    
    \begin{tabular}{c@{\hspace{1pt}}c@{\hspace{8pt}}c@{\hspace{1pt}}c@{\hspace{1pt}}c@{\hspace{1pt}}c@{\hspace{1pt}}c@{\hspace{1pt}}c@{\hspace{4pt}}l}
        
        \includegraphics[width=0.12\linewidth]{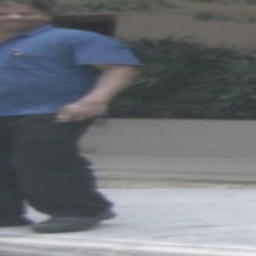} &
        \includegraphics[width=0.12\linewidth]{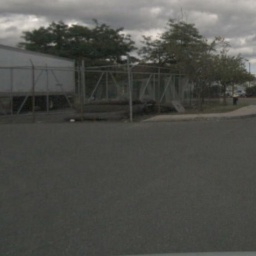} &
        \includegraphics[width=0.12\linewidth]{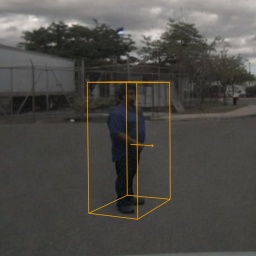} &
        \includegraphics[width=0.12\linewidth]{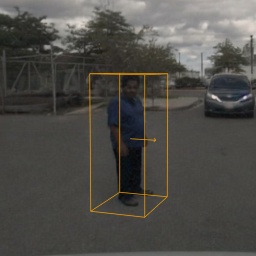} &
        \includegraphics[width=0.12\linewidth]{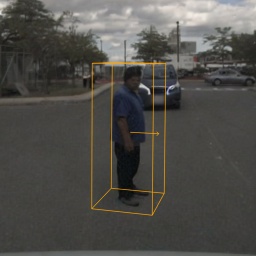} &
        \includegraphics[width=0.12\linewidth]{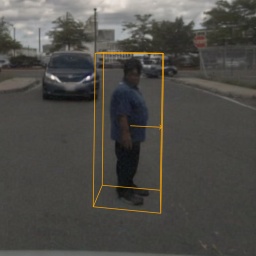} &
        \includegraphics[width=0.12\linewidth]{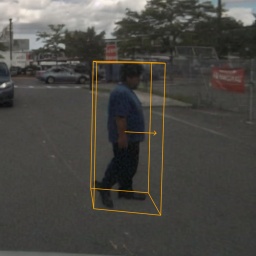} &
        \includegraphics[width=0.12\linewidth]{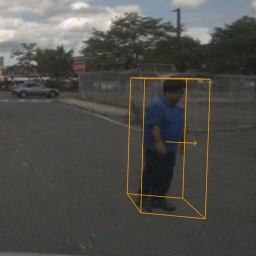} &
        \rotatebox{90}{\quad~ Camera} \\[1pt]
        
        &
        \includegraphics[width=0.12\linewidth]{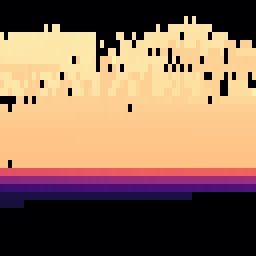} &
        \includegraphics[width=0.12\linewidth]{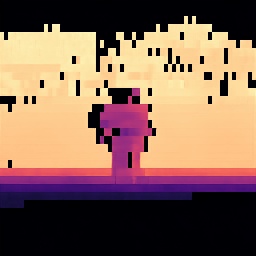} &
        \includegraphics[width=0.12\linewidth]{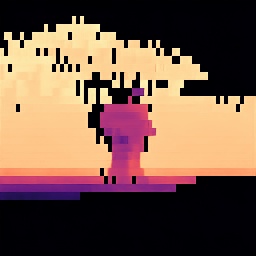} &
        \includegraphics[width=0.12\linewidth]{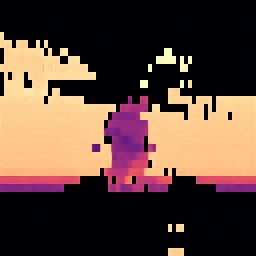} &
        \includegraphics[width=0.12\linewidth]{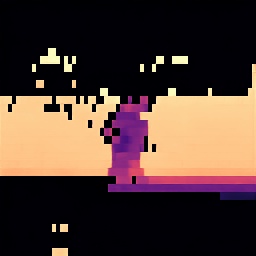} &
        \includegraphics[width=0.12\linewidth]{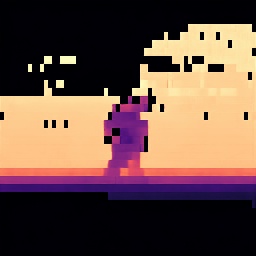} &
        \includegraphics[width=0.12\linewidth]{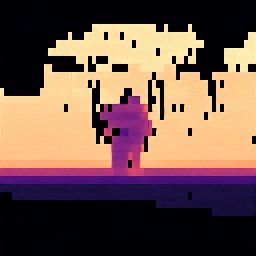} &
        \rotatebox{90}{~~ LiDAR depth} \\[1pt]
        
        &
        \includegraphics[width=0.12\linewidth]{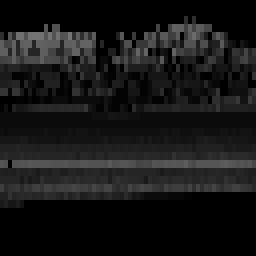} &
        \includegraphics[width=0.12\linewidth]{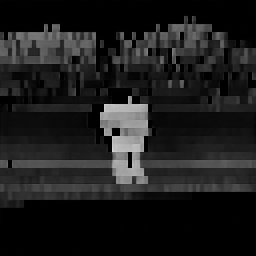} &
        \includegraphics[width=0.12\linewidth]{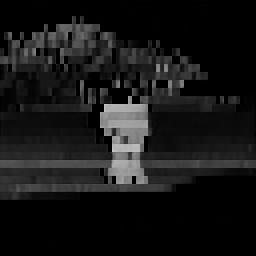} &
        \includegraphics[width=0.12\linewidth]{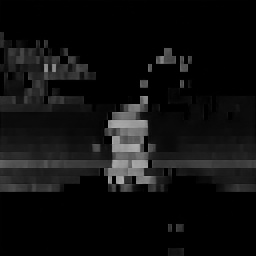} &
        \includegraphics[width=0.12\linewidth]{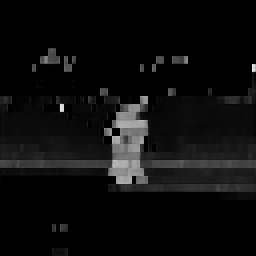} &
        \includegraphics[width=0.12\linewidth]{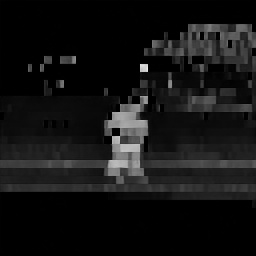} &
        \includegraphics[width=0.12\linewidth]{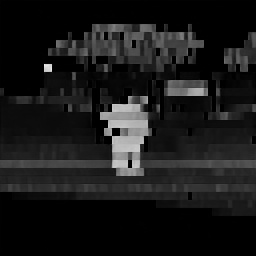} &
        \rotatebox{90}{LiDAR intensity} \vspace{8pt}
    \end{tabular}
    \begin{tabular}{c@{\hspace{1pt}}c@{\hspace{8pt}}c@{\hspace{1pt}}c@{\hspace{1pt}}c@{\hspace{1pt}}c@{\hspace{1pt}}c@{\hspace{1pt}}c@{\hspace{4pt}}l}
        
        \includegraphics[width=0.12\linewidth]{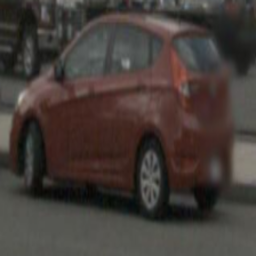} &
        \includegraphics[width=0.12\linewidth]{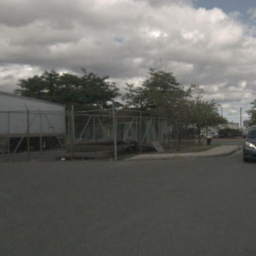} &
        \includegraphics[width=0.12\linewidth]{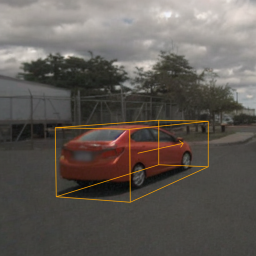} &
        \includegraphics[width=0.12\linewidth]{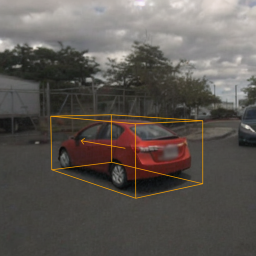} &
        \includegraphics[width=0.12\linewidth]{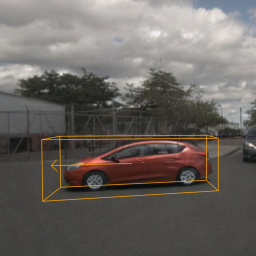} &
        \includegraphics[width=0.12\linewidth]{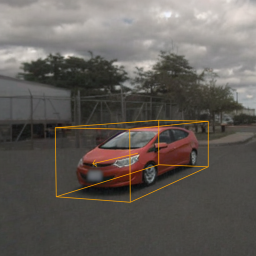} &
        \includegraphics[width=0.12\linewidth]{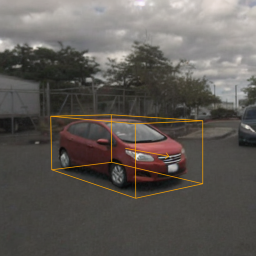} &
        \includegraphics[width=0.12\linewidth]{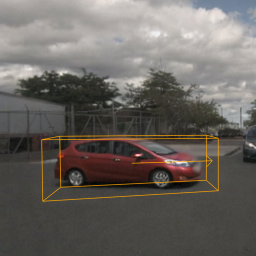} &
        \rotatebox{90}{\quad~ Camera} \\[1pt]
        
        &
        \includegraphics[width=0.12\linewidth]{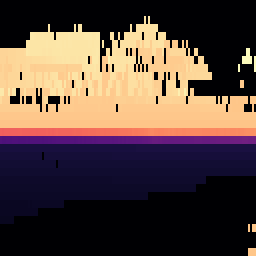} &
        \includegraphics[width=0.12\linewidth]{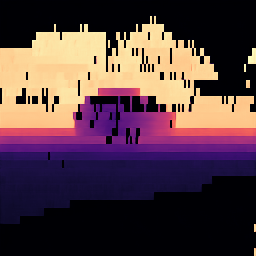} &
        \includegraphics[width=0.12\linewidth]{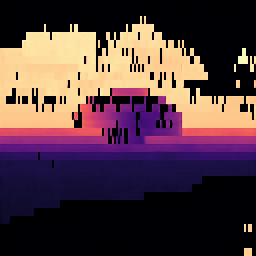} &
        \includegraphics[width=0.12\linewidth]{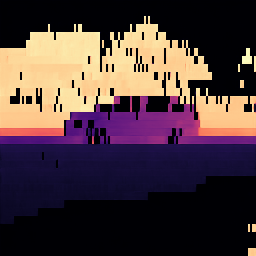} &
        \includegraphics[width=0.12\linewidth]{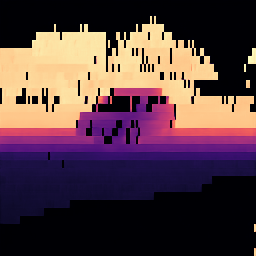} &
        \includegraphics[width=0.12\linewidth]{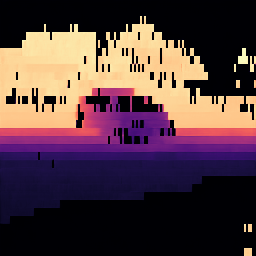} &
        \includegraphics[width=0.12\linewidth]{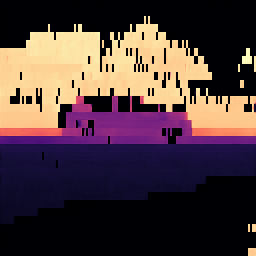} &
        \rotatebox{90}{~~ LiDAR depth} \\[1pt]
        
        &
        \includegraphics[width=0.12\linewidth]{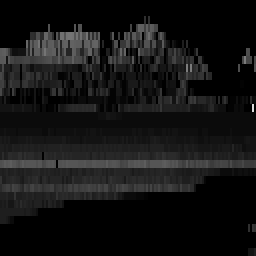} &
        \includegraphics[width=0.12\linewidth]{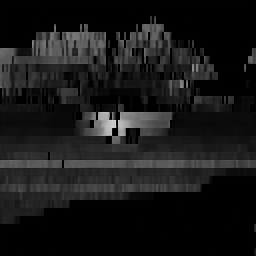} &
        \includegraphics[width=0.12\linewidth]{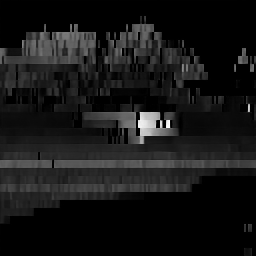} &
        \includegraphics[width=0.12\linewidth]{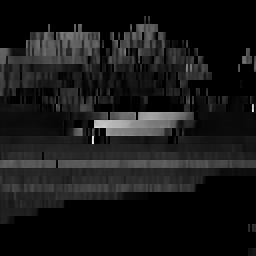} &
        \includegraphics[width=0.12\linewidth]{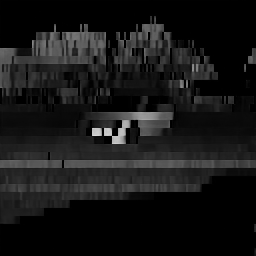} &
        \includegraphics[width=0.12\linewidth]{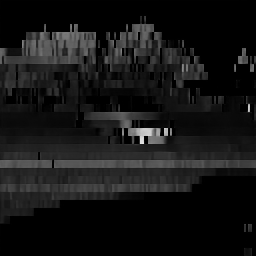} &
        \includegraphics[width=0.12\linewidth]{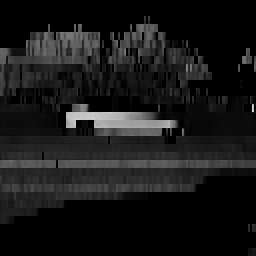} &
        \rotatebox{90}{LiDAR intensity}
    \end{tabular}
    \caption{Additional examples showcasing our method's controllability. From left to right: reference image $ \mathbf{x}_{\text{ref}} $ extracted from a seperate source scene, original destination scene (original RGB image $ \mathbf{x}^{\text{(C)}} $, LiDAR range depth $ \mathbf{x}_0^{\text{(R)}} $ and intensity $ \mathbf{x}_1^{\text{(R)}} $), and edited scenes.}
    \label{fig:suppl:controllability_main_full}
  \end{figure*}

\begin{figure*}[ht]
    \centering
    \begin{minipage}{\textwidth}
    \centering
    \begin{tabular}{ccc}
    \text{Ground Truth} & \text{Vanilla} & \text{Ours} \\
    \includegraphics[width=0.33\linewidth]{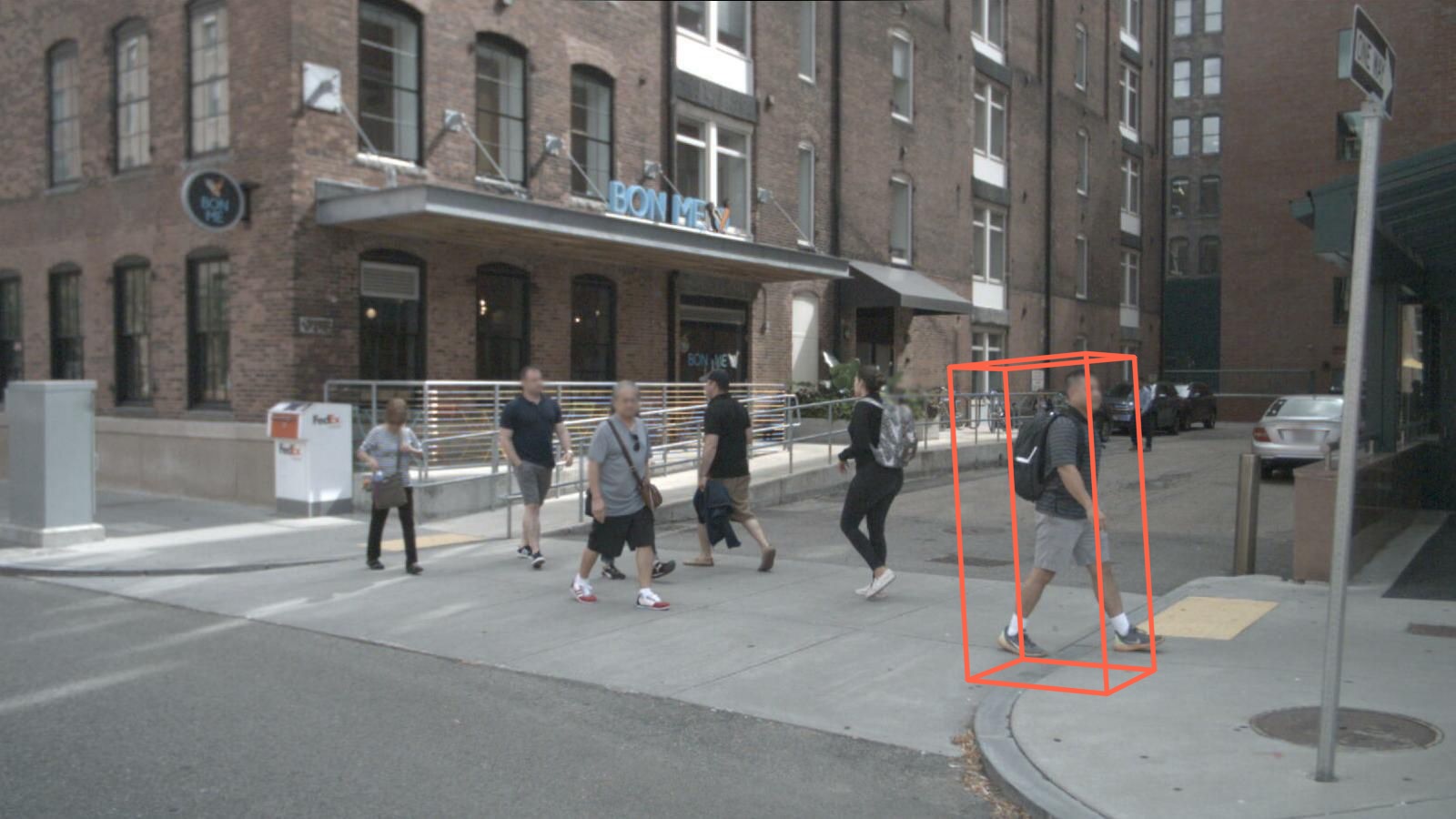} & 
    \includegraphics[width=0.33\linewidth]{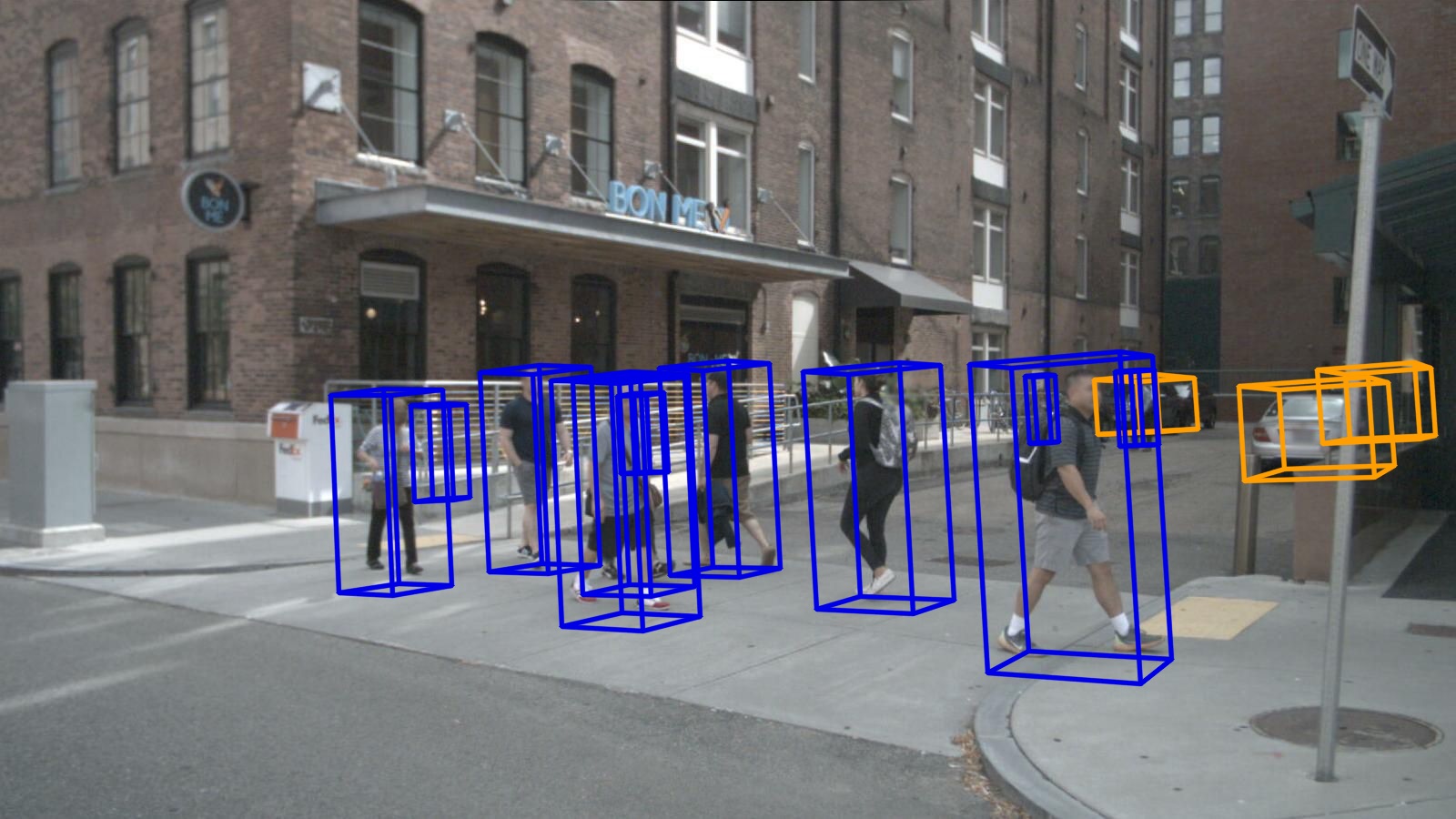} & 
    \includegraphics[width=0.33\linewidth]{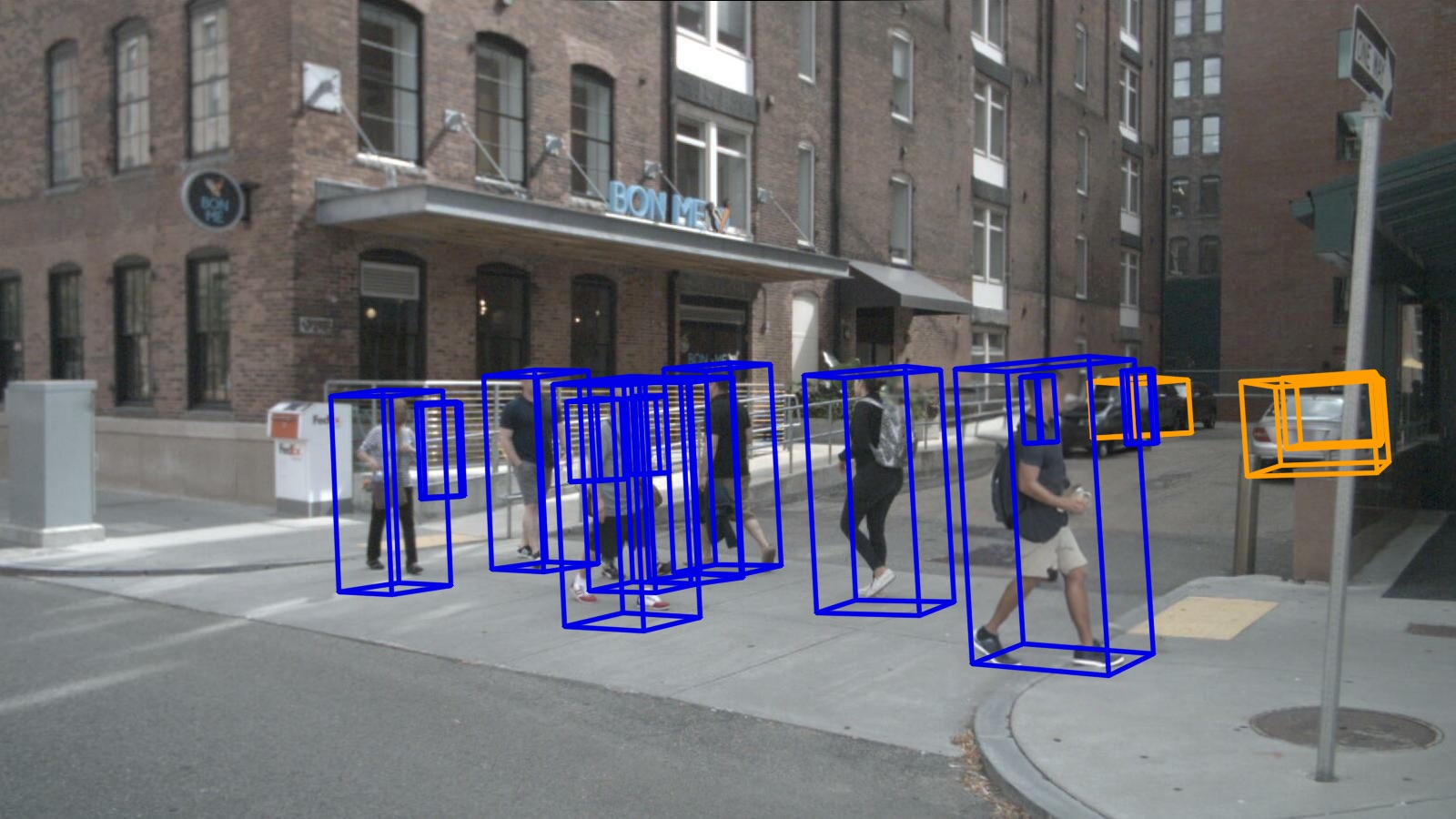} \\
    \includegraphics[width=0.33\linewidth]{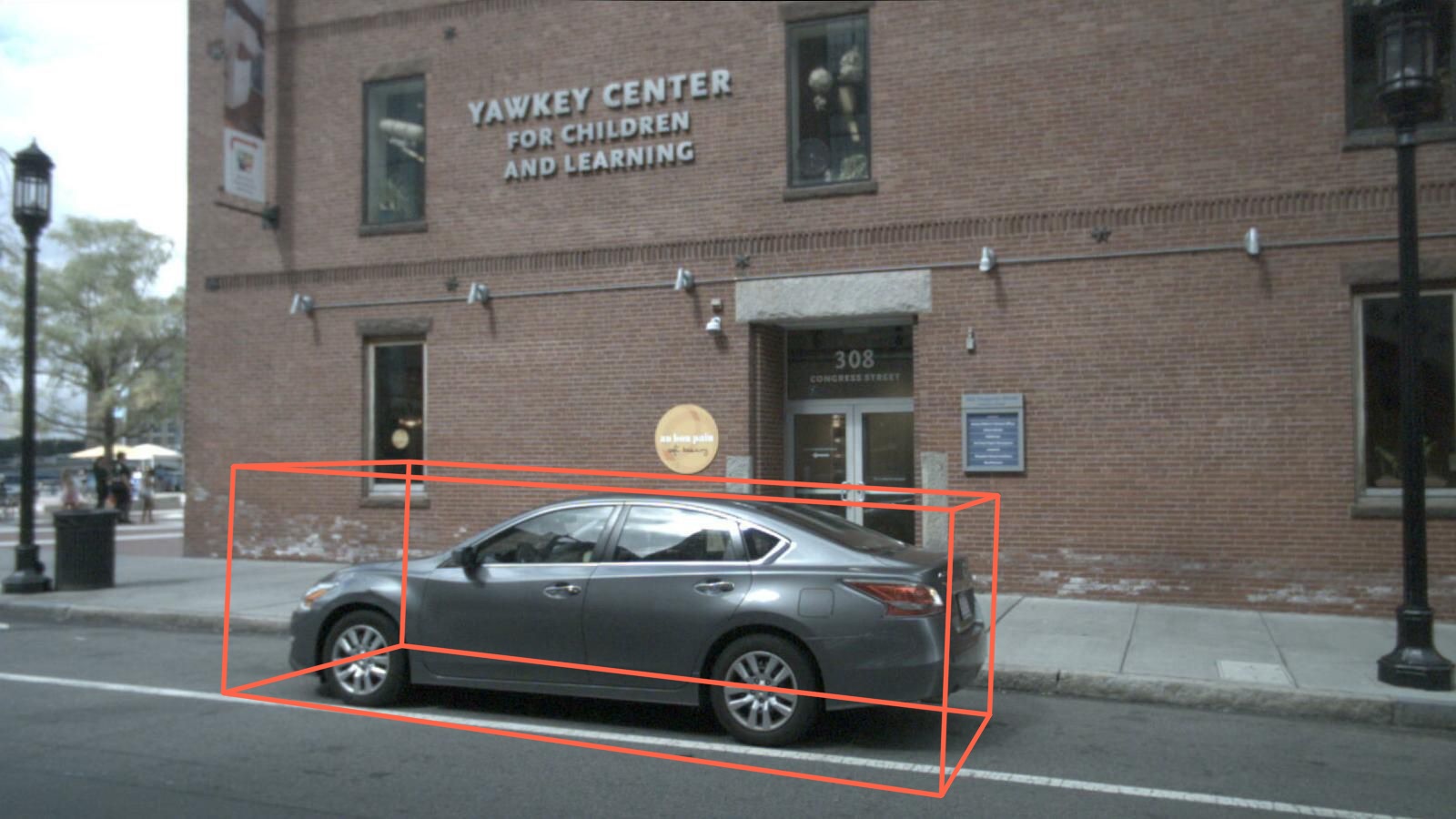} & 
    \includegraphics[width=0.33\textwidth]{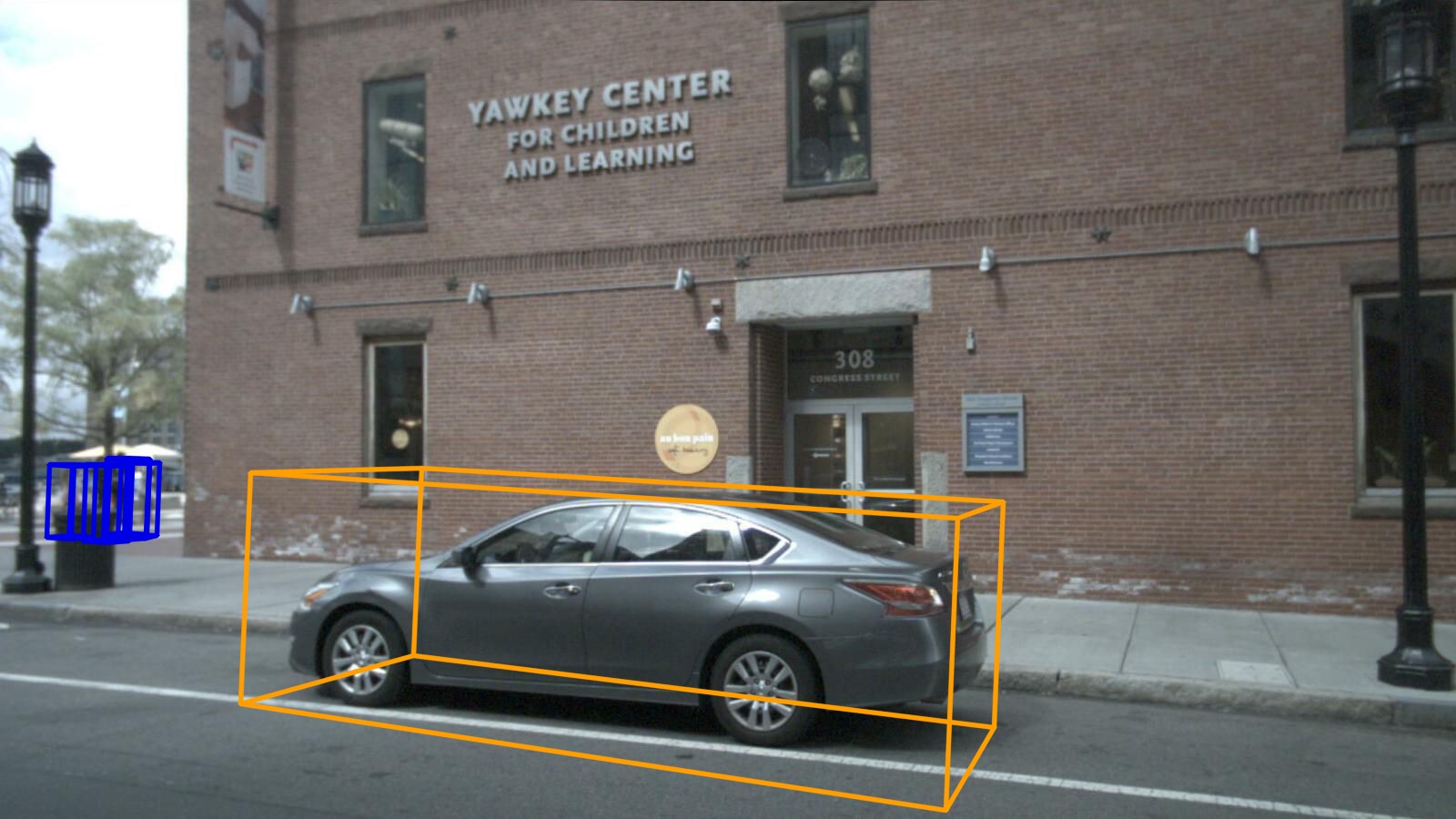} & 
    \includegraphics[width=0.33\linewidth]{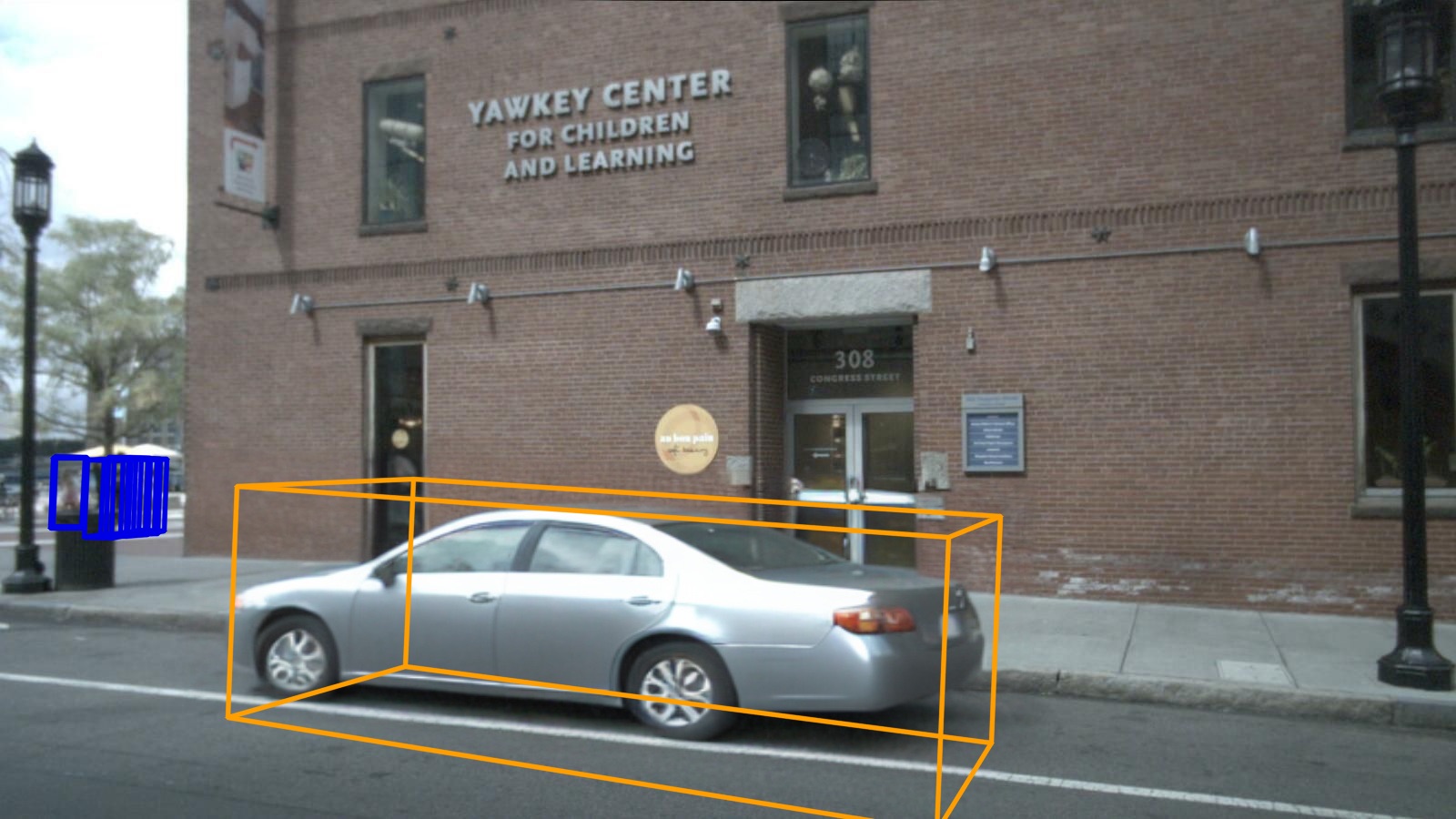} \\
    \includegraphics[width=0.33\linewidth]{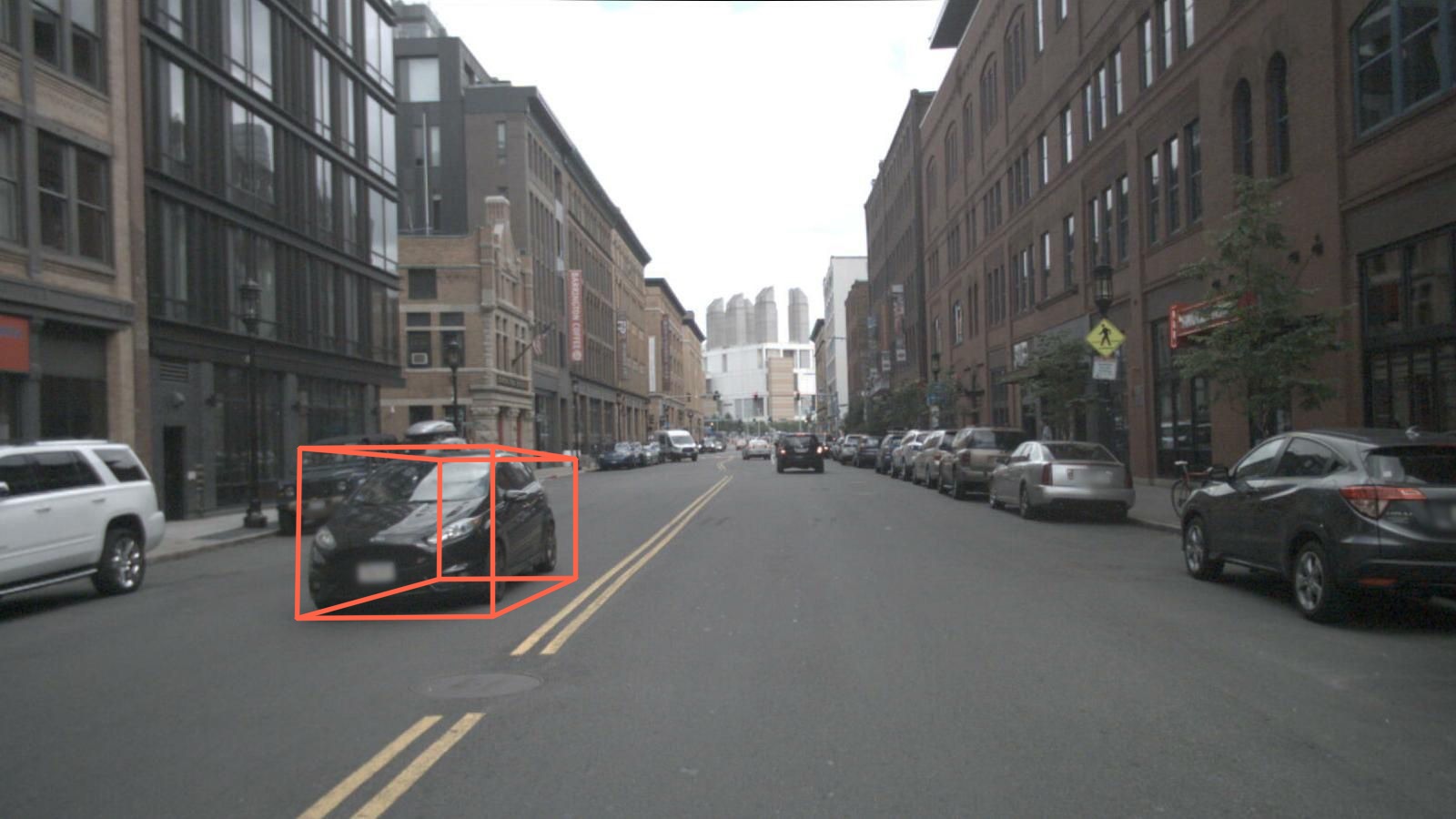} & 
    \includegraphics[width=0.33\linewidth]{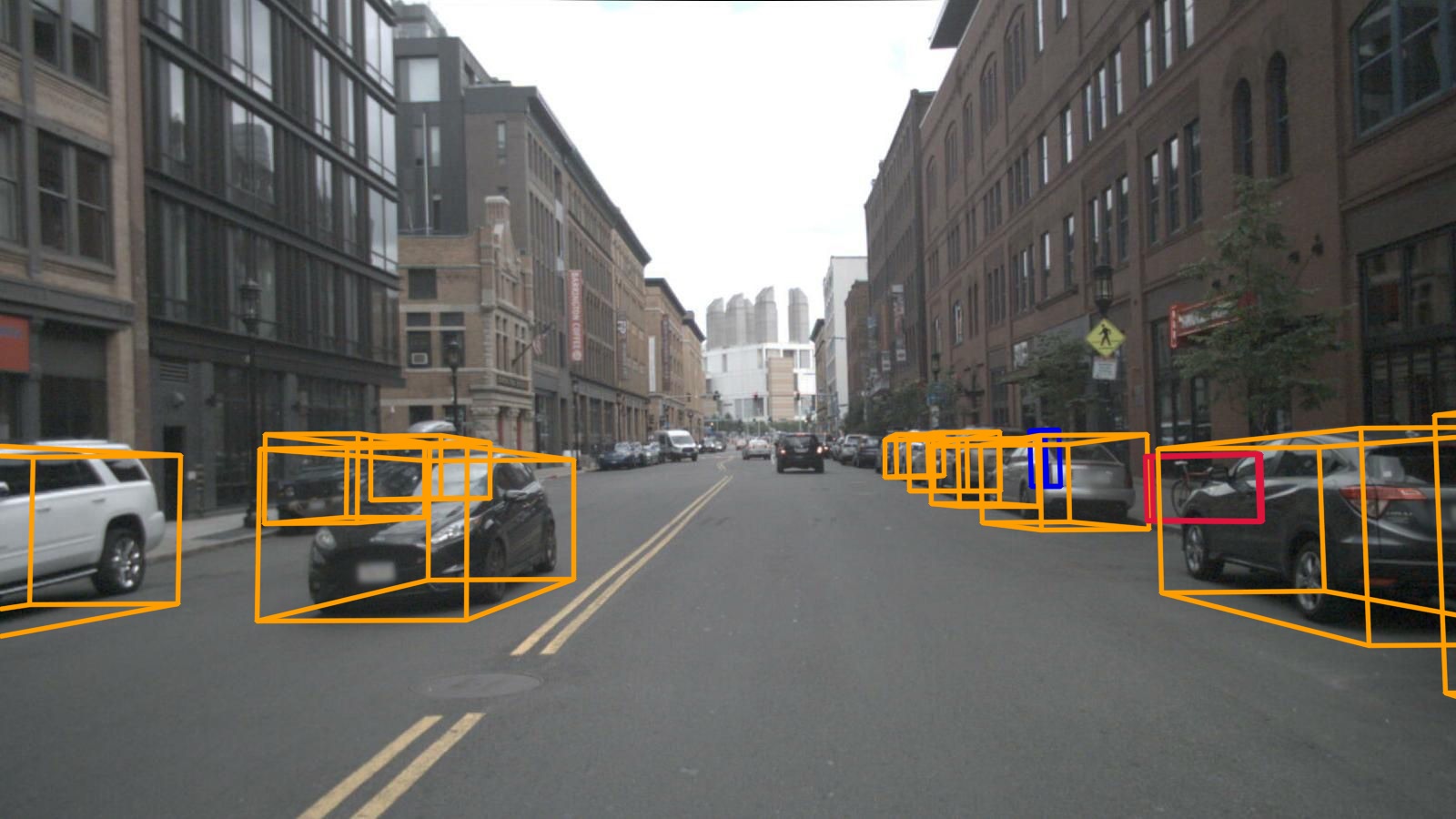} & 
    \includegraphics[width=0.33\linewidth]{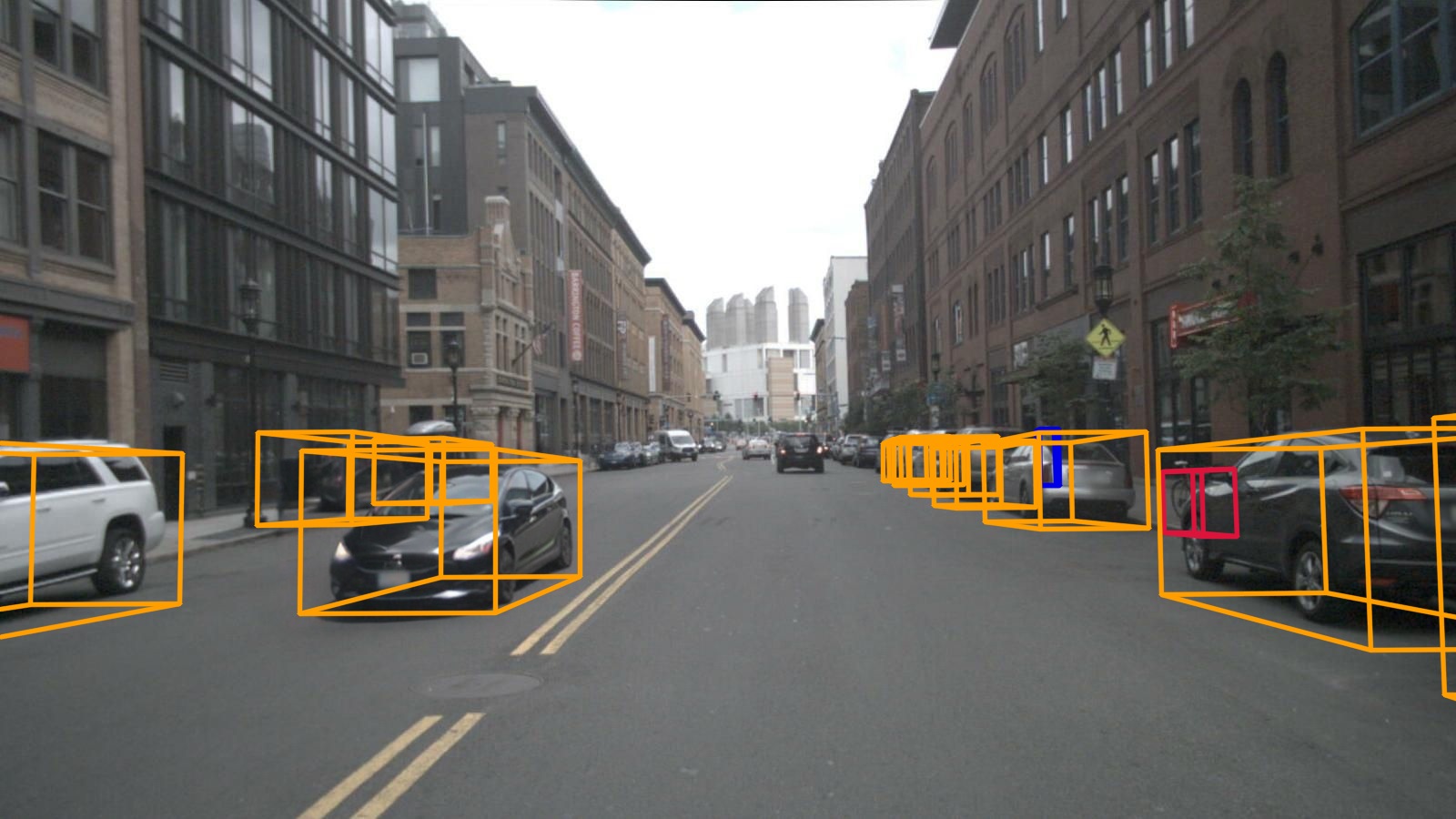} \\
    \end{tabular}
    \end{minipage}
    \caption{Comparison of detection results between the original scene and the same scene with the object shown in red replaced. BEVFusion~\cite{liu2023bevfusion} achieves good detection performance on the object reinserted using our method, while leaving the boxes of the other objects undisturbed.
    Interestingly, even though the aspect of the car behind the reinserted object in the third column is changed slightly, it does not seem to affect detection much.
    We hypothesise that this is due to the fact that while the camera view is sensitive to occlusions, the range view is much less so since we reinsert only the points that are in the box used for conditioning, see \cref{sec:method:spatial compositing}.
    All detections are filtered using a score threshold of 0.08.
    }
    \label{fig:detection-comparison}
\end{figure*}

\begin{figure*}[ht!]
\begin{minipage}{\textwidth}
\centering
\setlength{\tabcolsep}{1pt}
\begin{tabular}{ccccc}
Original & Reference & Edited (C) & Edited (R) - depth & Edited (R) - intensity \\
\includegraphics[width=0.18\textwidth]{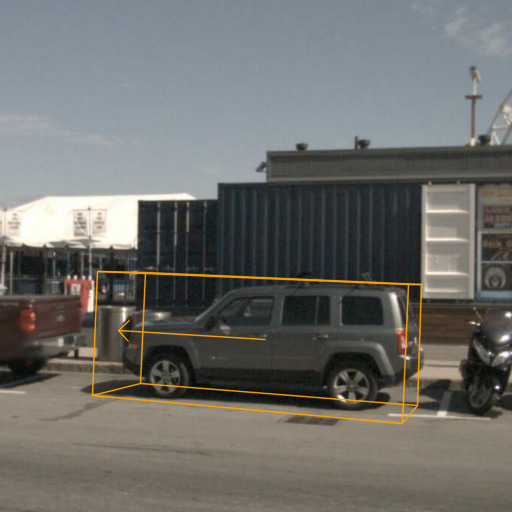} &
\includegraphics[width=0.18\textwidth]{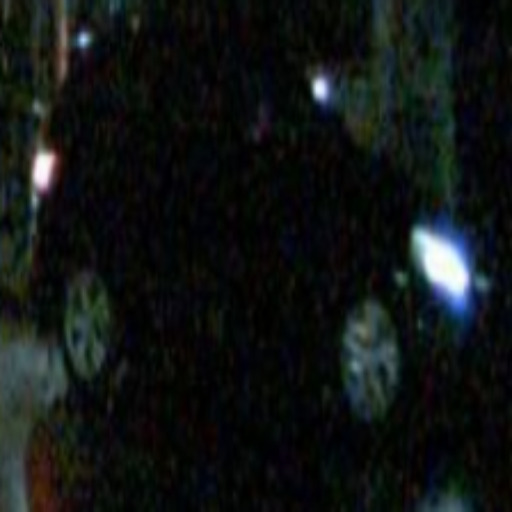} &
\includegraphics[width=0.18\textwidth]{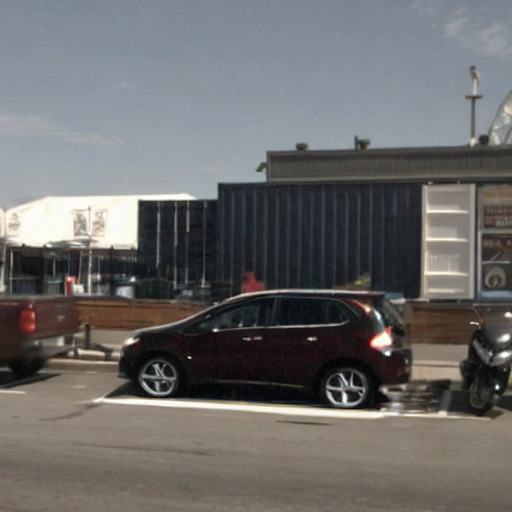} &
\includegraphics[width=0.18\textwidth]{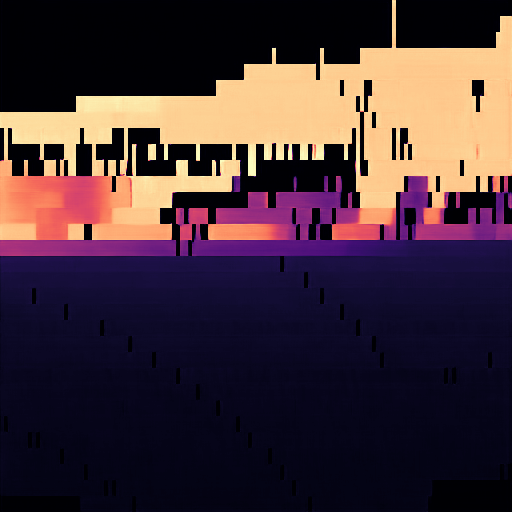} &
\includegraphics[width=0.18\textwidth]{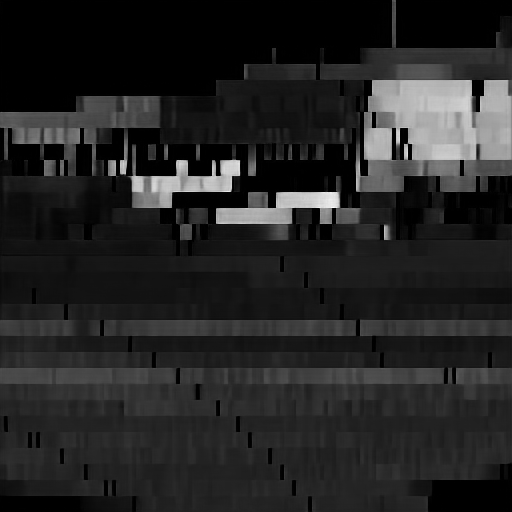} \\
\includegraphics[width=0.18\textwidth]{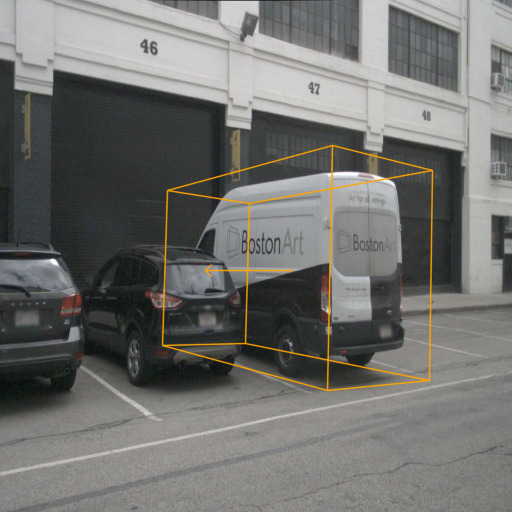} &
\includegraphics[width=0.18\textwidth]{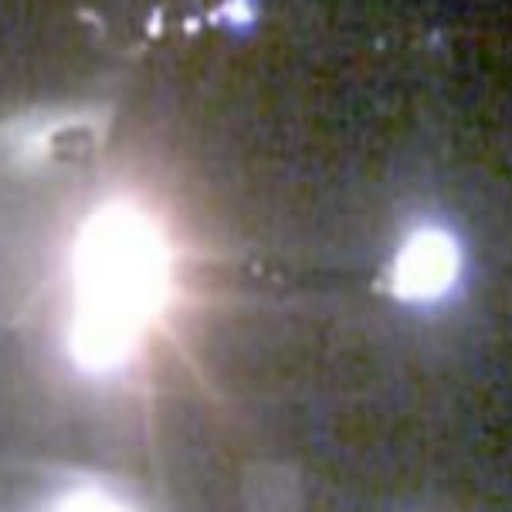} &
\includegraphics[width=0.18\textwidth]{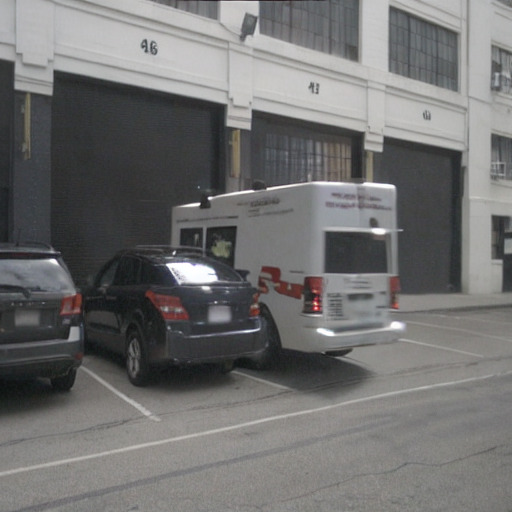} &
\includegraphics[width=0.18\textwidth]{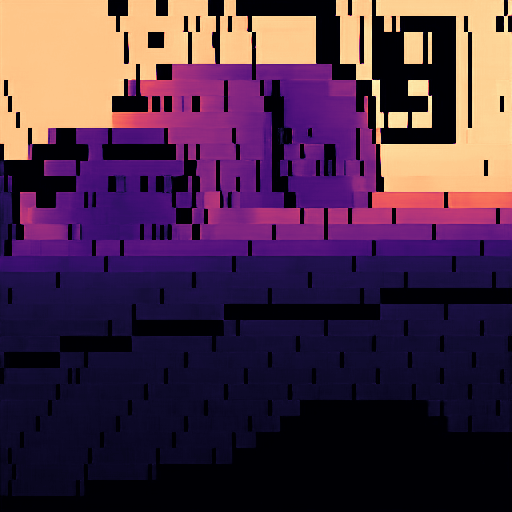} &
\includegraphics[width=0.18\textwidth]{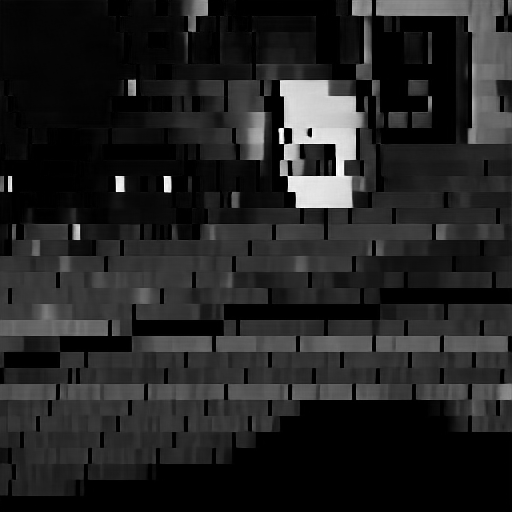} \\
\includegraphics[width=0.18\textwidth]{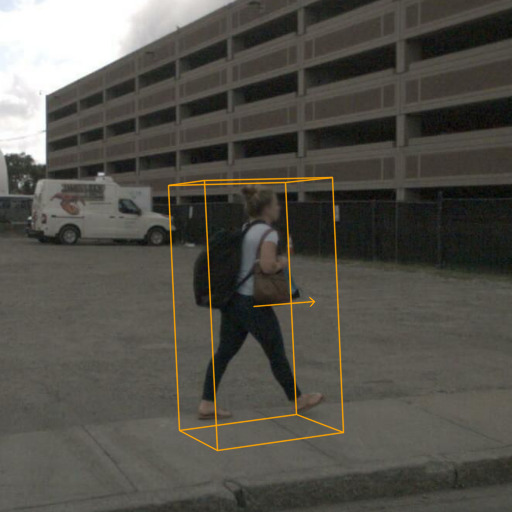} &
\includegraphics[width=0.18\textwidth]{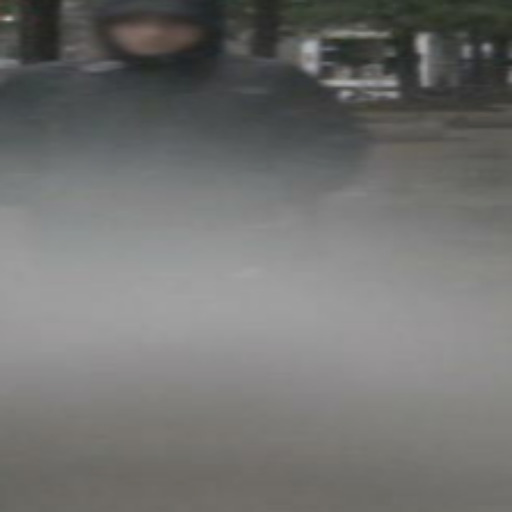} &
\includegraphics[width=0.18\textwidth]{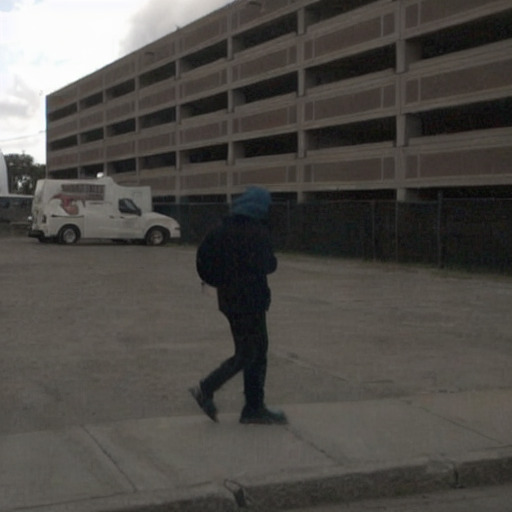} &
\includegraphics[width=0.18\textwidth]{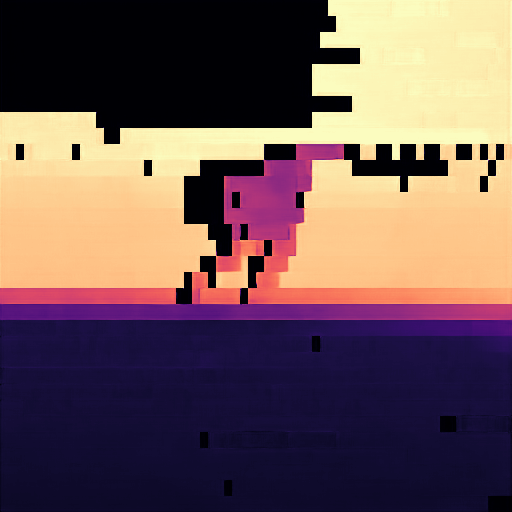} &
\includegraphics[width=0.18\textwidth]{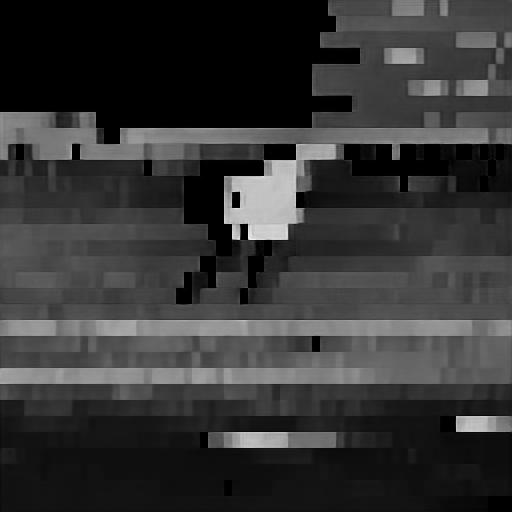} \\
\includegraphics[width=0.18\textwidth]{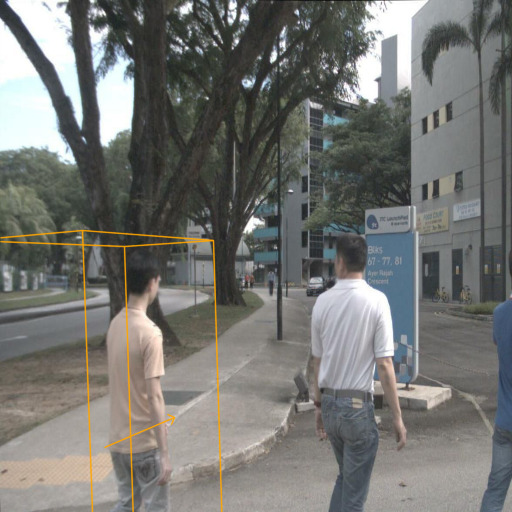} &
\includegraphics[width=0.18\textwidth]{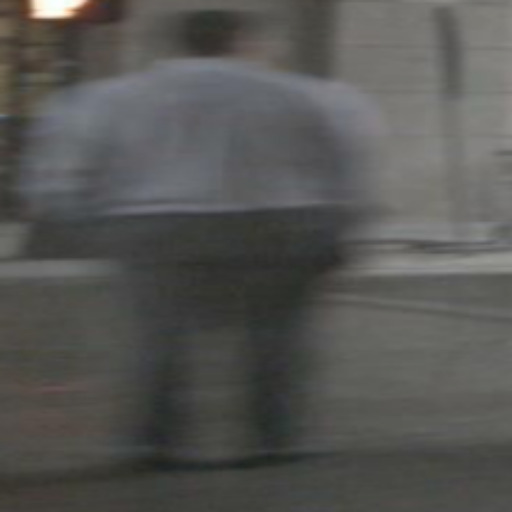} &
\includegraphics[width=0.18\textwidth]{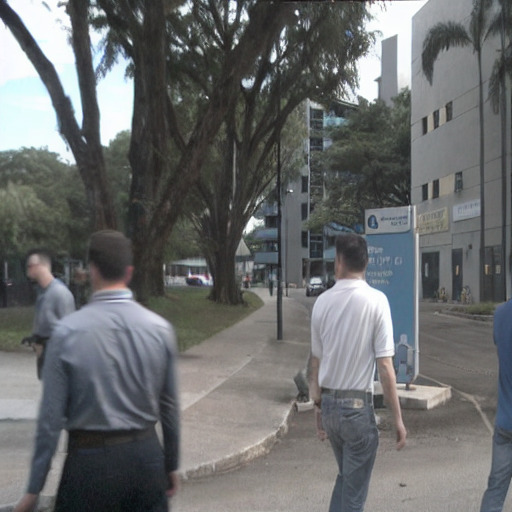} &
\includegraphics[width=0.18\textwidth]{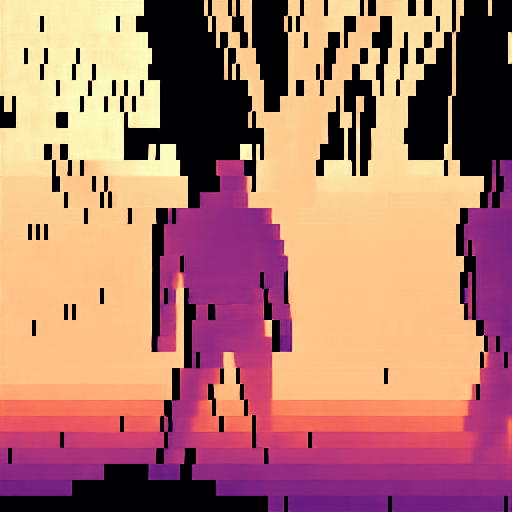} &
\includegraphics[width=0.18\textwidth]{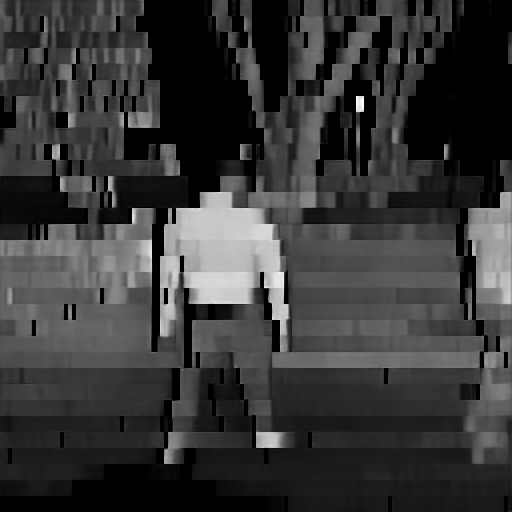} \\
\includegraphics[width=0.18\textwidth]{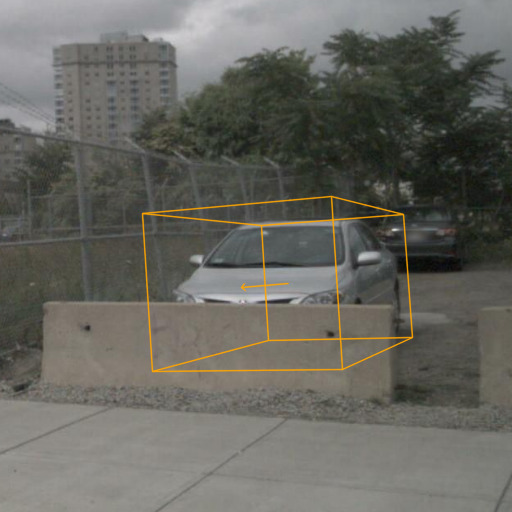} &
\includegraphics[width=0.18\textwidth]{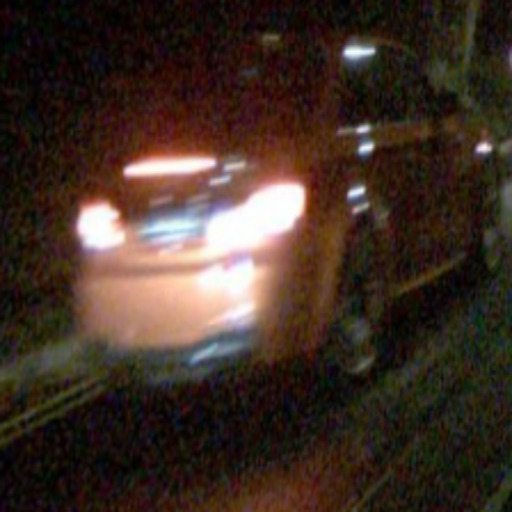} &
\includegraphics[width=0.18\textwidth]{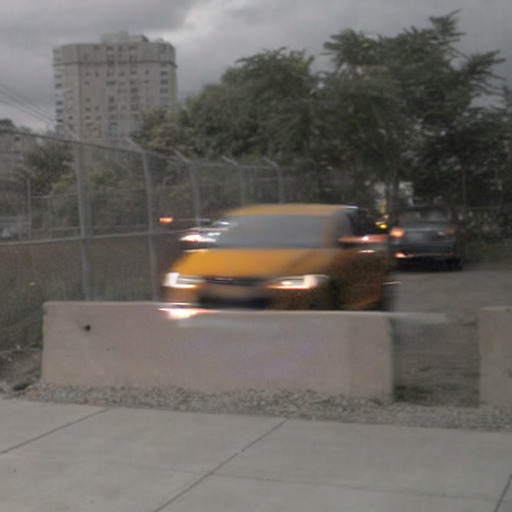} &
\includegraphics[width=0.18\textwidth]{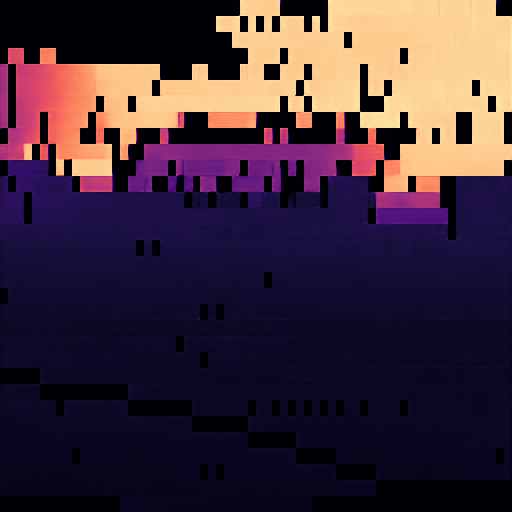} &
\includegraphics[width=0.18\textwidth]{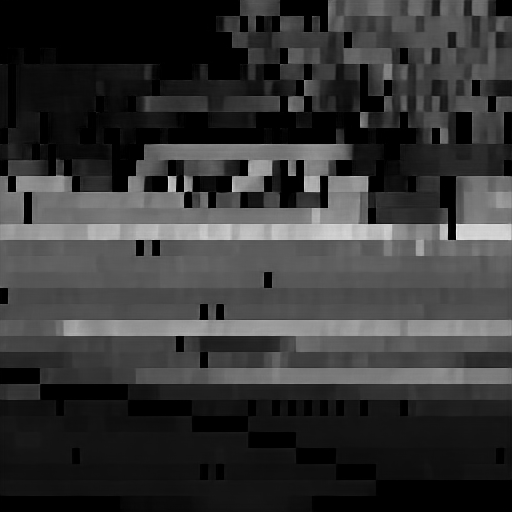} \\
\includegraphics[width=0.18\textwidth]{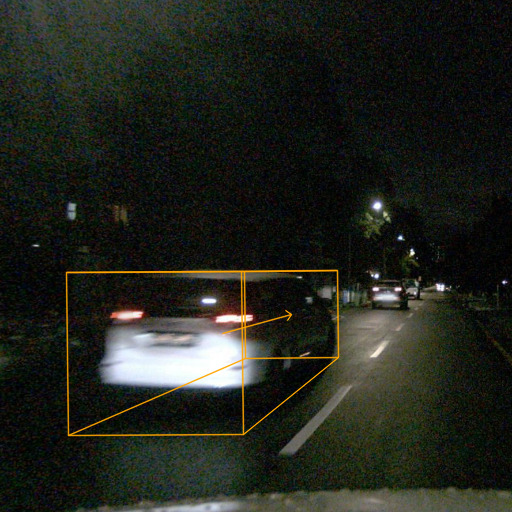} &
\includegraphics[width=0.18\textwidth]{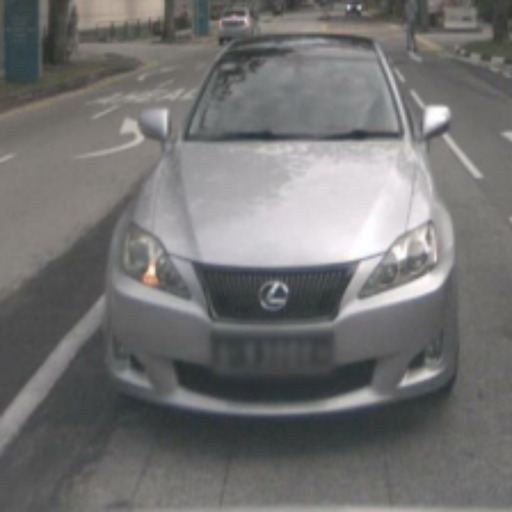} &
\includegraphics[width=0.18\textwidth]{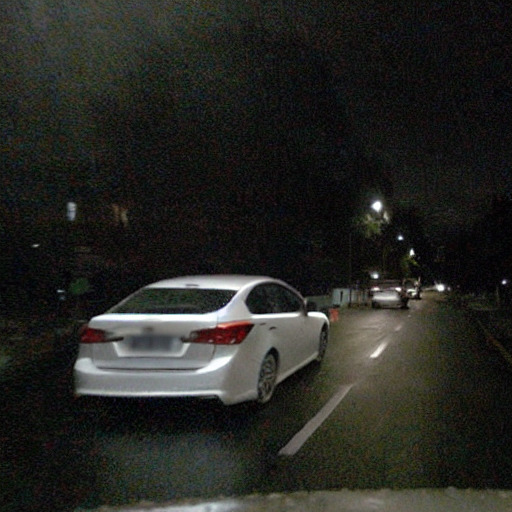} &
\includegraphics[width=0.18\textwidth]{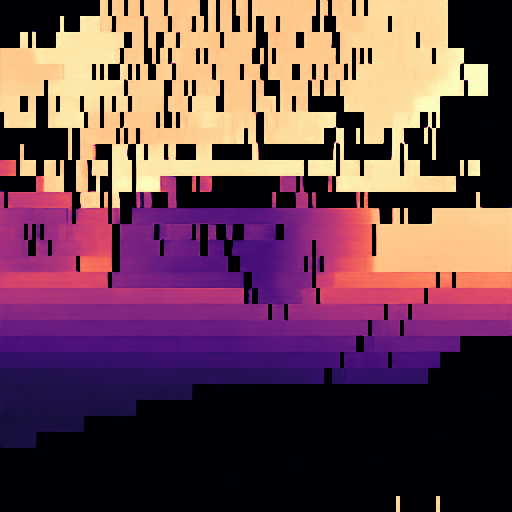} &
\includegraphics[width=0.18\textwidth]{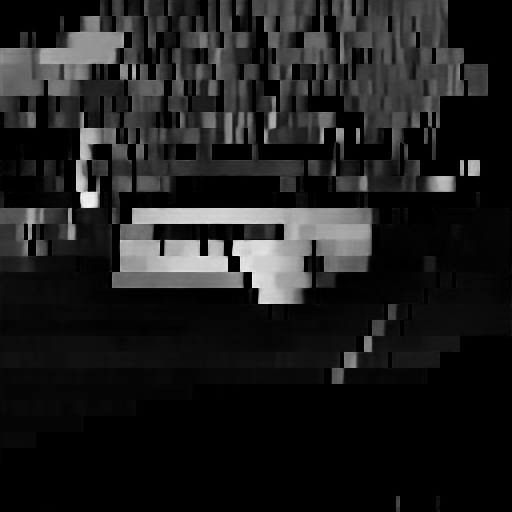} \\
\end{tabular}
\end{minipage}
\caption{Object replacement results using hard references (different weather conditions or time of day, occlusions, etc.). Top three rows: MObI is able to insert these hard references in the target bounding box successfully while preserving the overall scene consistency. Bottom three rows: some examples of failure cases (a new pedestrian is hallucinated, the inserted car shows too much motion blur, the lightning is not coherent with the overall scene).}
\label{fig:inpainting-hard-suppl}
\end{figure*}

\begin{figure*}[t]
\centering
\begin{minipage}{0.48\textwidth}
\centering
\setlength{\tabcolsep}{1pt} 
\renewcommand{\arraystretch}{0.8} 
\begin{tabular}{ccc}
Original & Reference & Insertion (C) \\
\includegraphics[width=0.33\linewidth]{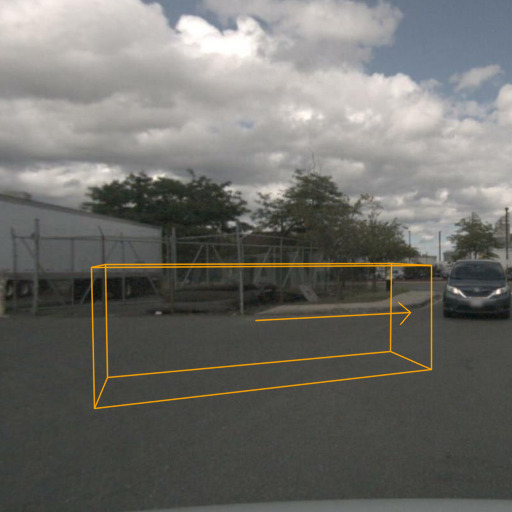} &
\includegraphics[width=0.33\linewidth]{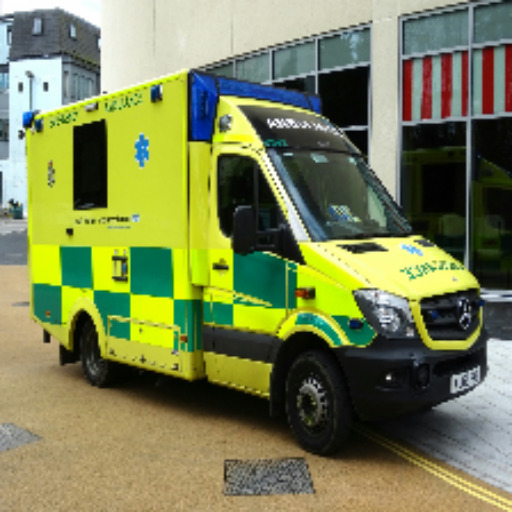} &
\includegraphics[width=0.33\linewidth]{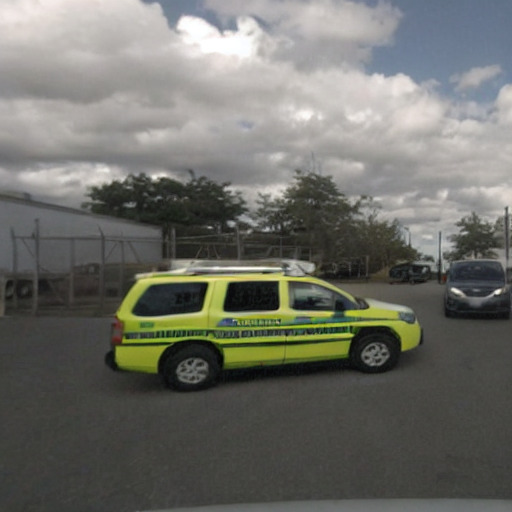} \\
\includegraphics[width=0.33\linewidth]{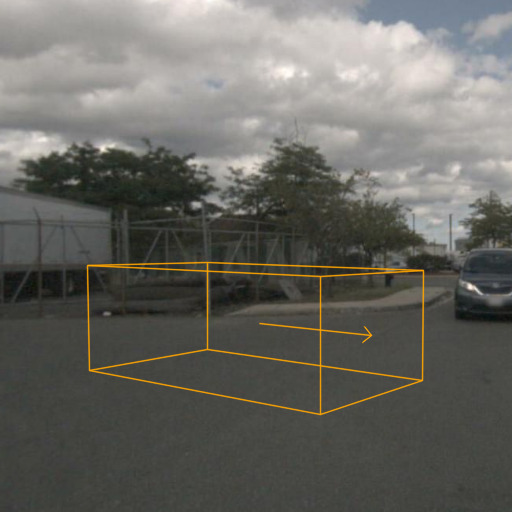} &
\includegraphics[width=0.33\linewidth]{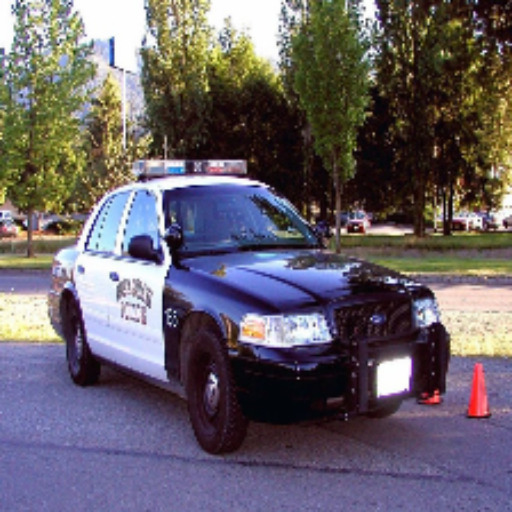} &
\includegraphics[width=0.33\linewidth]{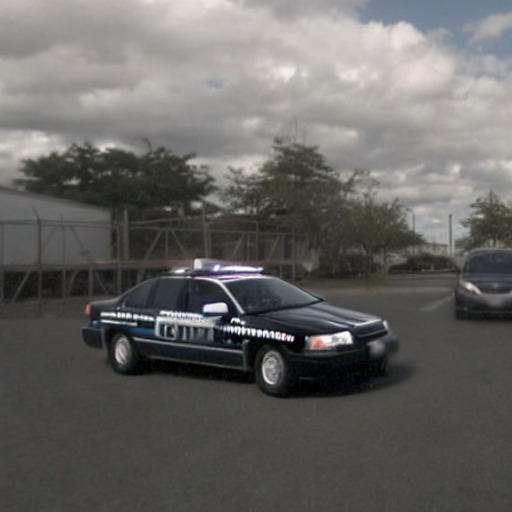} \\
\includegraphics[width=0.33\linewidth]{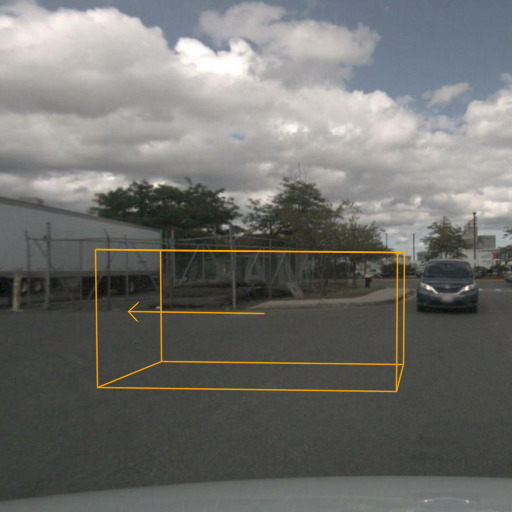} &
\includegraphics[width=0.33\linewidth]{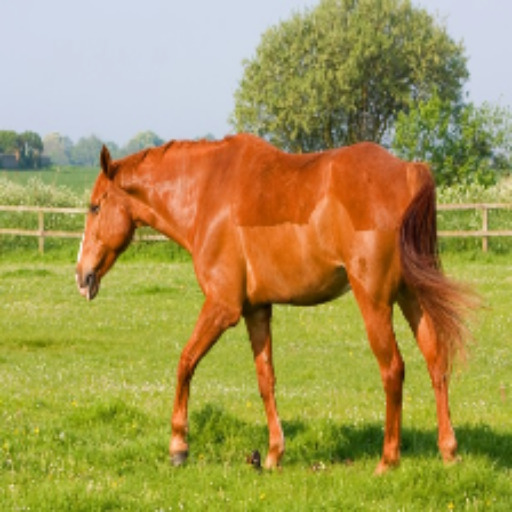} &
\includegraphics[width=0.33\linewidth]{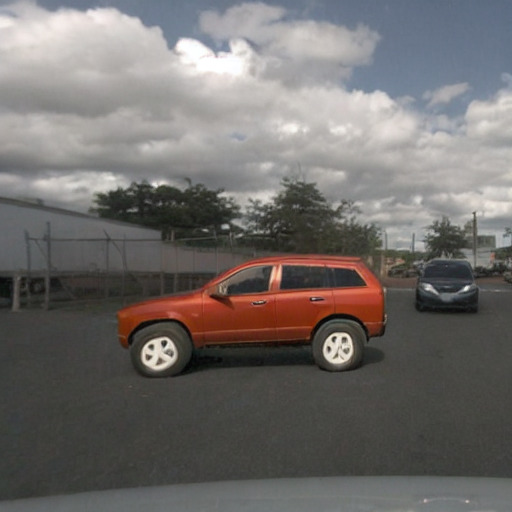} \\
\includegraphics[width=0.33\linewidth]{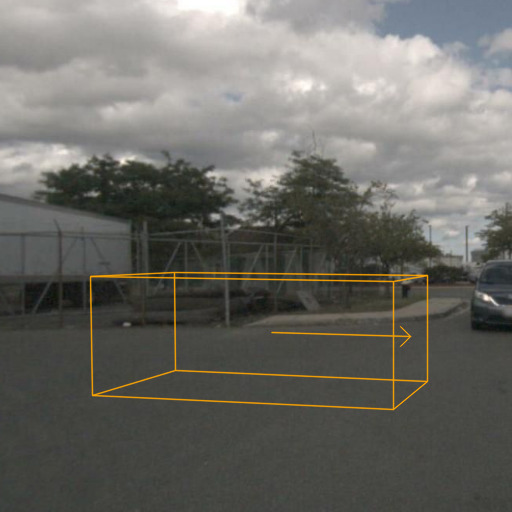} &
\includegraphics[width=0.33\linewidth]{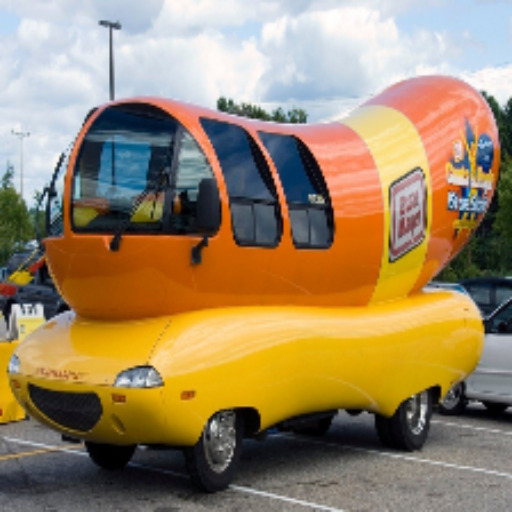} &
\includegraphics[width=0.33\linewidth]{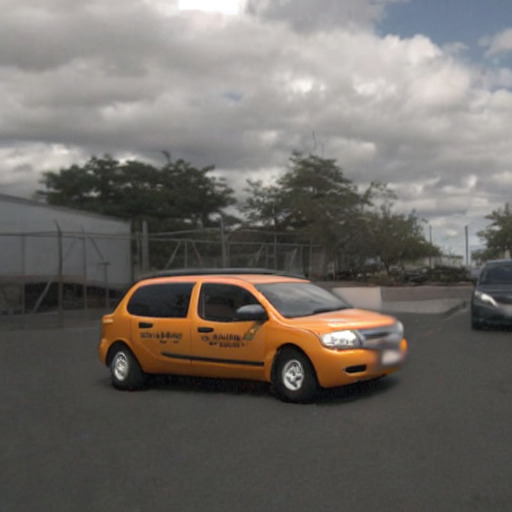}
\end{tabular}
\subcaption{}
\label{fig:suppl:ood_a}
\end{minipage}%
\hfill
\begin{minipage}{0.48\textwidth}
\centering
\setlength{\tabcolsep}{1pt} 
\renewcommand{\arraystretch}{0.8} 
\begin{tabular}{ccc}
Original & Reference & Replacement (C) \\
\includegraphics[width=0.33\linewidth]{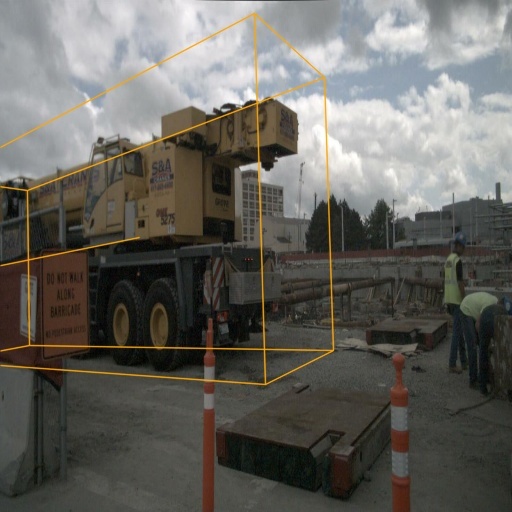} &
\includegraphics[width=0.33\linewidth]{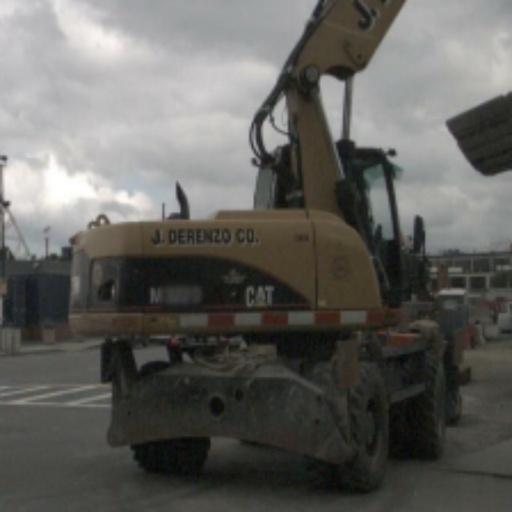} &
\includegraphics[width=0.33\linewidth]{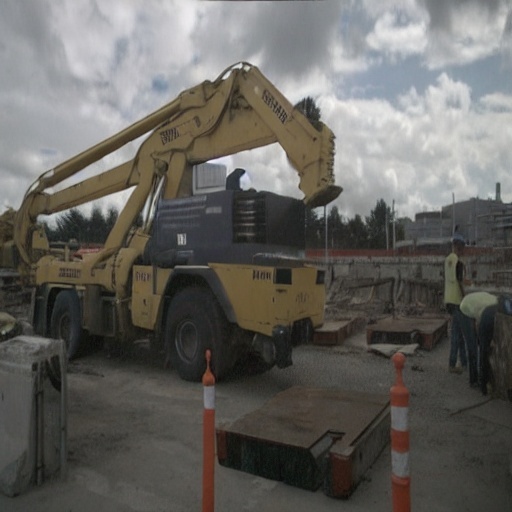} \\
\includegraphics[width=0.33\linewidth]{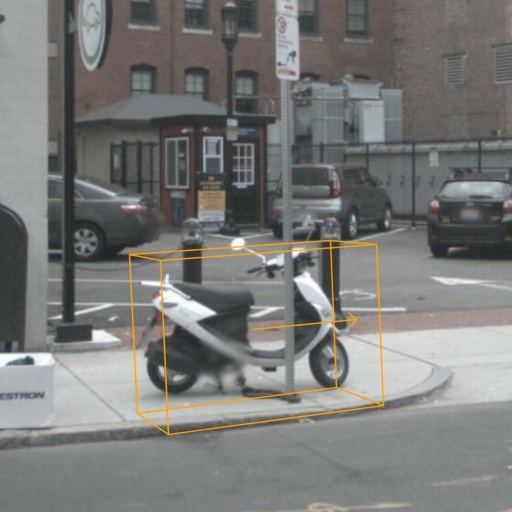} &
\includegraphics[width=0.33\linewidth]{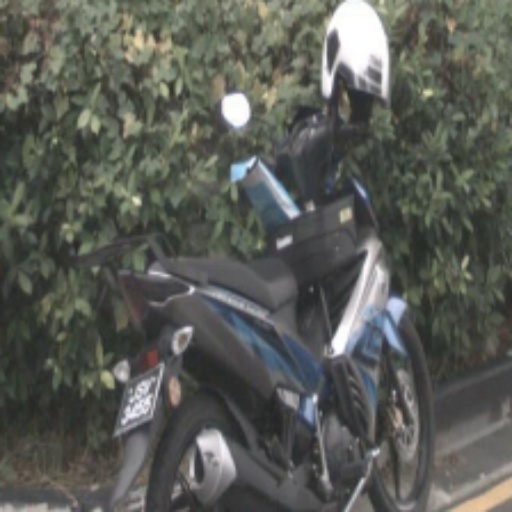} &
\includegraphics[width=0.33\linewidth]{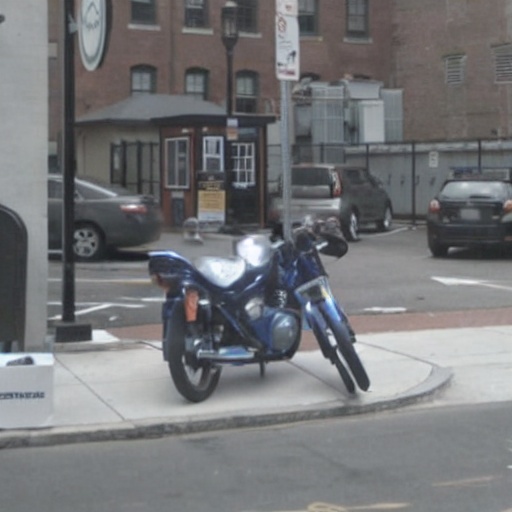} \\
\includegraphics[width=0.33\linewidth]{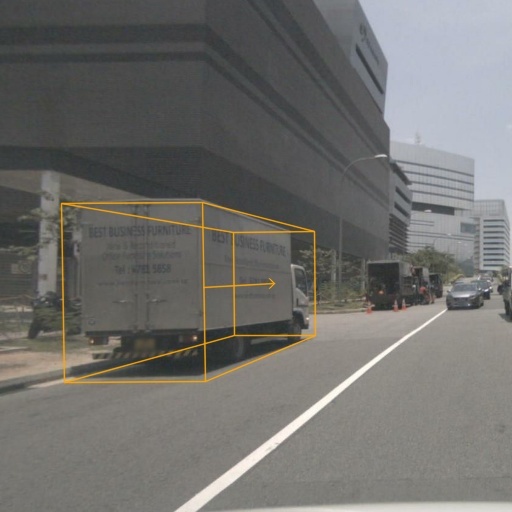} &
\includegraphics[width=0.33\linewidth]{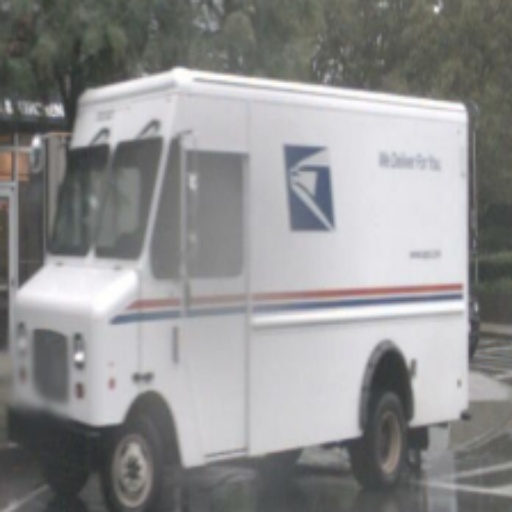} &
\includegraphics[width=0.33\linewidth]{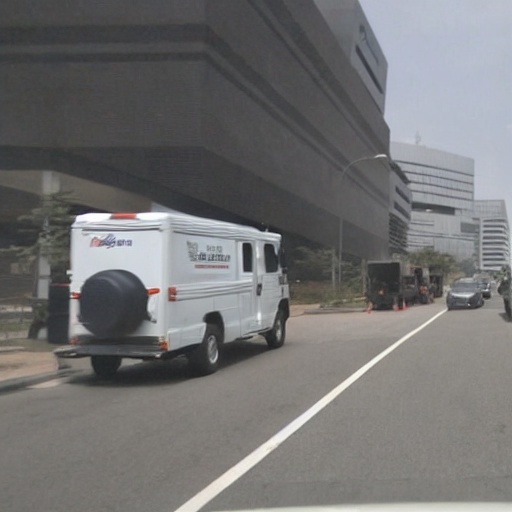} \\
\includegraphics[width=0.33\linewidth]{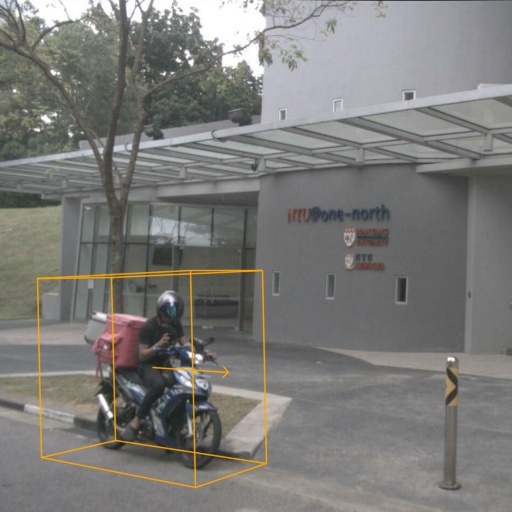} &
\includegraphics[width=0.33\linewidth]{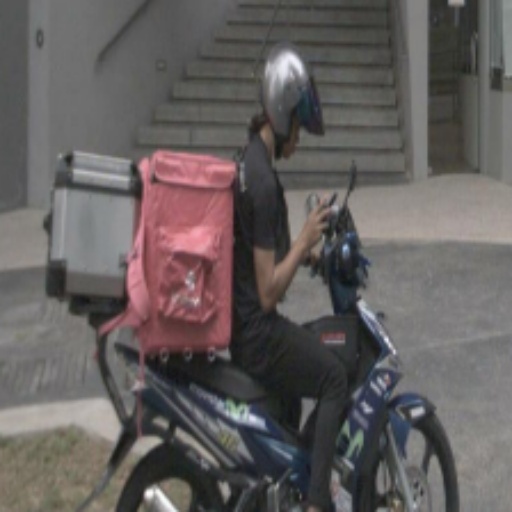} &
\includegraphics[width=0.33\linewidth]{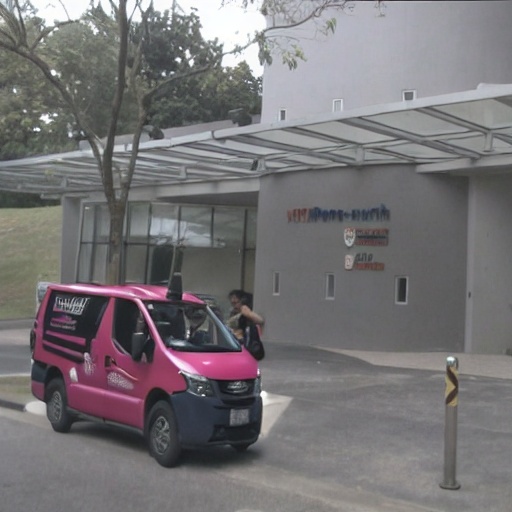}
\end{tabular}
\subcaption{}
\end{minipage}
\caption{Object insertion and replacement with out-of-domain and open-world references for MObI trained only on the pedestrian and car classes of nuScenes. (a) In the first two examples (top left), MObI inserts the correct object successfully but loses fine appearance details. In the last two examples (bottom left), MObI inserts a car instead of the object depicted by the reference. (b) In the first three examples (top right), MObI correctly replaces objects from classes outside of its training set, yet quality degrades. In the last example (bottom right), the model replaces the motorcycle with a small vehicle, reverting to a familiar class. Note that all examples have been correctly inserted in the target bounding box with the correct orientation.}
\label{fig:suppl:ood}
\end{figure*}

\end{document}